\documentclass[fleqn,12pt]{wlscirep}
\usepackage{multirow}
\usepackage[utf8]{inputenc}
\usepackage[T1]{fontenc}
\usepackage{listings}
\usepackage{graphicx}
\usepackage{booktabs}
\usepackage{ragged2e}
\usepackage{makecell}
\usepackage{adjustbox}
\usepackage[normalem]{ulem}

\usepackage{subcaption}
\usepackage{tcolorbox}
\usepackage{setspace}
\usepackage{microtype}
\usepackage{longtable}
\usepackage{placeins}
\tcbuselibrary{breakable}
\usepackage[capitalise]{cleveref}

\newcommand{\soneimg}{86{,}965}
\newcommand{\soneart}{26{,}569}
\newcommand{\stwoins}{26{,}929}
\newcommand{\stwocon}{12}
\newcommand{\sthreerep}{913}

\newcommand{\mnames}{MLLMs}

\usepackage{etoolbox}
\makeatletter
\patchcmd{\@maketitle}
  {\fontsize{20}{25}\selectfont}
  {\fontsize{24}{28}\selectfont}
  {}{}

\patchcmd\@maketitle{%
        {%
        \noindent
        \colorbox{color2}{%
            \parbox{\dimexpr\linewidth-2\fboxsep\relax}{%
            \sffamily\small\textbf\\\theabstract
            }%
        }%
        }%
    }{%
        \par\bigskip
        \theabstract\par
        \bigskip
    }%
    {}{}%
\makeatother

\usepackage{siunitx}
\title{VOLMO: Versatile and Open Large Models for Ophthalmology}

\author[1]{Zhenyue Qin}
\author[1]{Younjoon Chung}
\author[1]{Elijah Lee}
\author[2]{Wanyue Feng}
\author[1]{Xuguang Ai}
\author[1]{Serina Applebaum}
\author[3]{Minjie Zou}
\author[2]{Yang Liu}
\author[4]{Pan Xiao}
\author[1]{Mac Singer}
\author[1]{Amisha Dave}
\author[5]{Aidan Gilson}
\author[6]{Tiarnan D. L. Keenan}
\author[6]{Emily Y. Chew}
\author[7]{Zhiyong Lu}
\author[3]{Yih-Chung Tham}
\author[1]{Ron Adelman}
\author[1]{Luciano V. Del Priore}
\author[1,*]{Qingyu Chen}

\affil[1]{Department of Biomedical Informatics \& Data Science, Yale University}
\affil[2]{Ray and Stephanie Lane Computational Biology Department, Carnegie Mellon University}
\affil[3]{Yong Loo Lin School of Medicine, National University of Singapore}
\affil[4]{Department of Radiology, Washington University in Saint Louis}
\affil[5]{Harvard Medical School, Harvard University}
\affil[6]{National Eye Institute, National Institutes of Health}
\affil[7]{National Library of Medicine, National Institutes of Health}

\affil[*]{qingyu.chen@yale.edu}

\begin{abstract}
Vision impairment affects millions of people globally, and early detection is critical to preventing irreversible vision loss. Current ophthalmological workflows rely heavily on clinicians manually integrating medical images, structured data such as visual acuity, and free-text notes such as clinical history to determine disease severity and formulate assessment and treatment plans. This process is time-consuming and contributes substantially to clinical burden. Recent advances in multimodal large language models (MLLMs) offer new capabilities. However, existing general-domain and medical-domain MLLMs perform poorly in ophthalmology; for example, they often achieve near-random diagnostic accuracy on primary eye diseases. To date, few ophthalmology-specific MLLMs are publicly available.
We present VOLMO (Versatile and Open Large Models for Ophthalmology), a model-agnostic and data-open framework for developing ophthalmology-specific MLLMs. VOLMO consists of three stages:
(1) Ophthalmology knowledge pretraining using \soneimg{} image–text pairs from \soneart{} articles across 82 journals;
(2) Domain task fine-tuning with \stwoins{} annotated instances across 12 eye conditions for disease screening and severity classification; and
(3) Multi-step clinical reasoning using 913 comprehensive patient case reports for assessment, planning, and follow-up care.
Using this framework, we trained a compact 2B-parameter MLLM and compared it against strong baselines, including InternVL-2B, LLaVA-Med-7B, MedGemma-4B, MedGemma-27B, and RETFound (used for task-specific fine-tuned classification tasks). We evaluated these models on three core applications: ophthalmological image-description generation, disease screening and staging classification, and assessment-and-management generation, using standard benchmarks. Beyond benchmark evaluations, we conducted manual assessments by two healthcare professionals on generated image descriptions and performed external validations on three independent patient cohorts covering age-related macular degeneration and diabetic retinopathy.
Across all evaluation settings, VOLMO-2B consistently outperformed baseline models. For instance, in image-description generation, it achieved the highest performance in both automated metrics (e.g., 11\% higher ROUGE-L than MedGemma-27B) and manual evaluations (e.g., average conciseness score of 4.4/5). For disease screening, it reached an average F1 score of 87.4\% across 12 eye conditions. In external validations, it demonstrated robust generalization across independent populations.
\end{abstract}
\begin{document}
\setstretch{1.2}

\flushbottom
\maketitle
\setstretch{1.2}
\thispagestyle{empty}

\section{Introduction}

The increasing prevalence of eye diseases presents a major global public health burden~\cite{varma2016visual,yang2021global,zhang2024global,chong2024diabetic}. Worldwide, vision impairment due to age-related macular degeneration (AMD) is projected to reach approximately 21.34 million cases by 2050~\cite{zhang2024global}. In the United States alone, the number of individuals living with visual impairment or blindness is expected to double by 2050, rising to more than 8 million people~\cite{varma2016visual}. Diabetic retinopathy (DR) also remains a significant challenge, with up to 50\% of patients failing to receive timely examinations or being diagnosed only after irreversible damage has occurred~\cite{jani2017evaluation,chong2024diabetic}. Limited access to eye care and substantial manual burden for both clinicians and patients continue to impede early detection and timely intervention~\cite{ervin2022access,baxter2019time,allison2025analysis}.

Advances in artificial intelligence (AI) offer promising solutions to these challenges and have been applied across a broad spectrum of ophthalmological applications over the past decade, including eye disease screening~\cite{cheung2019artificial,wu2024systematic}, disease severity classification~\cite{peng2019deepseenet,tognetto2022artificial}, progression prediction~\cite{silva2024automated,yang2025artificial}, and risk-factor~\cite{keenan2019deep} or lesion segmentation~\cite{nguyen2025deep,ying2024application}. Beyond eye diseases, AI has also been used to identify systemic conditions such as cardiovascular and neurological diseases directly from retinal images~\cite{tan2024prognostic,diaz2022predicting}. Importantly, AI systems have begun to enter real-world clinical workflows, with validation studies demonstrating improvements in manual diagnostic accuracy and efficiency for eye diseases~\cite{chen2025ai,skevas2024implementing}. Several systems have also undergone clinical trials and have been deployed in practice~\cite{wu2025eyecare,chen2025artificial}.

The major paradigm of AI in ophthalmology has also evolved substantially. Early work focused on convolutional neural networks (CNNs), where models used convolutional and pooling layers to automatically extract imaging features and learn mappings to disease labels~\cite{diaz2019cnns,ting2019artificial}.
These models are typically supervised fine-tuned on manually labeled eye imaging datasets for specific diseases, such as AMD classification~\cite{chen2021classification}. More recently, foundation vision models have been proposed, using self-supervised learning (e.g., randomly masking regions of an image and predicting the masked regions) to leverage extensive unlabeled eye images, followed by supervised fine-tuning for specific tasks~\cite{cai2022uni4eye}. Pioneering foundation vision models in ophthalmology, such as RETFound~\cite{zhou2023foundation}, self-supervised over ~1.6 million retinal
images, have demonstrated that foundation models consistently improved the effectiveness of downstream applications and have been widely adopted~\cite{zhang2024retfound,chen2025independent,hou2026can}.
More recent
efforts scale vision foundation models to vision language models (VLM). Pioneering studies leverage contrastive learning that draws matched image-text pairs closer and pushes unmatched pairs apart in the embedding space. Representative models in ophthalmology, such as EyeCLIP~\cite{shi2025multimodal}
and RetiZero~\cite{wang2025enhancing}, demonstrate that this approach can produce more robust imaging features that support tasks like image retrieval or can be fine-tuned further for disease diagnosis. 

While contrastive learning has been effective for aligning vision and language representations within a shared semantic space, these models are primarily designed for retrieval and classification tasks and lack the core capabilities required to synthesize visual and textual information through generation and reasoning~\cite{kamath2023text,hou2024vision} i.e., capabilities that are essential for clinical applications beyond image--text matching. Vision foundation models and other contrastive-based approaches also depend on task-specific fine-tuning and are difficult to adapt to new tasks~\cite{udandarao2023sus,rajendran2025foundation}, limiting their utility in practical clinical workflows.
In contrast, multimodal large language models (MLLMs) such as GPT-4V~\cite{achiam2023gpt}, InternVL~\cite{chen2024internvl}, Qwen-VL~\cite{wang2024qwen2}, and LLaVA~\cite{liu2023visual} offer substantially broader multimodal reasoning and generative capabilities, and emerging studies show 
their significant potential in medical applications
compared to previous models~\cite{alsaad2024multimodal,sun2024medical,buess2025large,huang2023visual,holland2025specialized}. These models can interpret images and text jointly (e.g., multimodal reasoning across patient data)~\cite{alsaad2024multimodal}, generate clinically structured outputs such as assessments, differential diagnoses, and treatment plans~\cite{sun2024medical}, perform multiple tasks (e.g., screening, diagnosis, and report generation) within a single unified model without task-specific retraining~\cite{buess2025large}, and support zero-shot generalization through instruction-based prompting~\cite{huang2023visual}. Such capabilities are especially crucial in ophthalmology, such as documenting retinal imaging findings, generating comprehensive assessments, formulating treatment plans, and producing follow-up recommendations~\cite{dow2022data,holland2025specialized,swaminathan2024unveiling}, 

Despite their promise, MLLMs demonstrate substantial limitations in ophthalmology. Existing studies consistently show that general-domain MLLMs, and even medical MLLMs, perform poorly in ophthalmological applications. For example, a pilot evaluation of Gemini Pro for identifying 10 ophthalmological risk factors from Optical Coherence Tomography (OCT) scans reported a mean F1 score of just 10.7\%~\cite{antaki2024vision}. 
Systematic assessments across more than 20 MLLMs further reveal that state-of-the-art models struggle with essential components of ophthalmological image understanding (e.g., anatomical structure recognition, lesion identification, and disease severity classification) with the strongest model, GPT-4o, achieving only 54\% accuracy for glaucoma screening as an example~\cite{qin2025lmod,qin2025lmod+}. A recent benchmarking study also showed the best-performing general-domain MLLM achieved only about 55\% accuracy on key ophthalmological classification tasks~\cite{xu2025benchmarking}.
The studies have emphasized that domain-specific training is essential to unlock the full potential of MLLMs in ophthalmology~\cite{antaki2024vision,haghighi2025eye}. To date, however, ophthalmology-specific MLLMs remain undeveloped; to our knowledge, as of when the manuscript has been written, no publicly available MLLM exists for ophthalmology that the community can validate, develop, and deploy.

To address these limitations, we introduce \textbf{VOLMO (Versatile and Open Large Models for Ophthalmology)}, a model-agnostic and data-open framework for developing MLLMs in ophthalmology. VOLMO is designed to enable the development of openly accessible foundation models in ophthalmology and to allow the community to validate, adopt, and further train or adapt these models using their own local data, without proprietary barriers or potential privacy concerns. The framework consists of three stages, each constructed from publicly available data with permissive licenses. Stage~1 ophthalmology knowledge pretraining leverages \soneimg{} image--text pairs curated from 26{,}569 articles across 82 ophthalmology journals to establish broad domain understanding. Stage~2 domain task fine-tuning incorporates \stwoins{} manually annotated instances covering \stwocon{} major eye conditions and signs to support disease screening and severity classification. Stage~3 multi-step clinical reasoning and synthesis uses \sthreerep{} comprehensive patient case reports to teach structured assessments, treatment planning, and follow-up recommendations, to facilitate end-to-end ophthalmological care.

Using this framework, we trained a compact 2B-parameter MLLM (VOLMO-2B) and systematically evaluated it against five strong baselines, including InternVL-2B, LLaVA-Med-7B, MedGemma-4B, MedGemma-27B, and RETFound (for classification-based tasks). The evaluation included head-to-head comparisons for each stage: ophthalmological image-description generation, disease screening and disease staging classification, and assessment-and-management generation. Beyond standard benchmark evaluations, we conducted clinical validations in which two healthcare professionals manually assessed the generated clinically focused image descriptions, and we performed external validations on three independent patient cohorts: (i) AMD and DR screening using the UK Biobank dataset, and (ii) DR severity classification using cohorts from Sydney Innovation and SUSTech.

Overall, VOLMO-2B demonstrated consistently strong performance compared with the baselines across all evaluation settings. For ophthalmological image-description generation, it achieved the highest automatic evaluation scores, including 11\% and 8\% absolute improvements over MedGemma-27B, in ROUGE-L and BERTScore, respectively. VOLMO-2B also received the highest conciseness score (4.43 vs.\ 2.06 for MedGemma-27B) and readability score (4.41 vs.\ 2.38) in manual evaluations conducted by two healthcare professionals. For disease screening, VOLMO-2B reached an average F1 of 87.41\% across \stwocon{} conditions and signs, substantially higher than MedGemma-27B (61.75\%), MedGemma-4B (60.65\%), LLaVA-Med (38.53\%), and InternVL-2B (34.83\%). It also outperformed RETFound, which was fine-tuned separately for each condition, in 8 of the 12 screened diseases. For disease staging, VOLMO-2B achieved the highest performance among the models, including 92.62\% F1 for macular-hole severity and 46.80\% for diabetic retinopathy severity. In assessment-and-management generation, the model produced more clinically aligned outputs, demonstrating 17--44\% higher semantic similarity to clinician-written documentation than the baselines. In external validations, VOLMO-2B achieved the highest AMD screening performance on the UK Biobank cohort (64.6\%~F1, up to 30\% higher than other models) and robust DR screening and severity classification results across the UK Biobank, Sydney Innovation, and SUSTech cohorts. Notably, these results were obtained using a single 2B-parameter model capable of running inference on widely accessible hardware, including consumer-grade GPUs (e.g., RTX~3050--4090) or laptops with 8--16\,GB RAM using quantized formats, which has potential for adoption in resource-limited clinical environments. 
\begin{figure}
    \centering
    \includegraphics[width=1.0\linewidth]{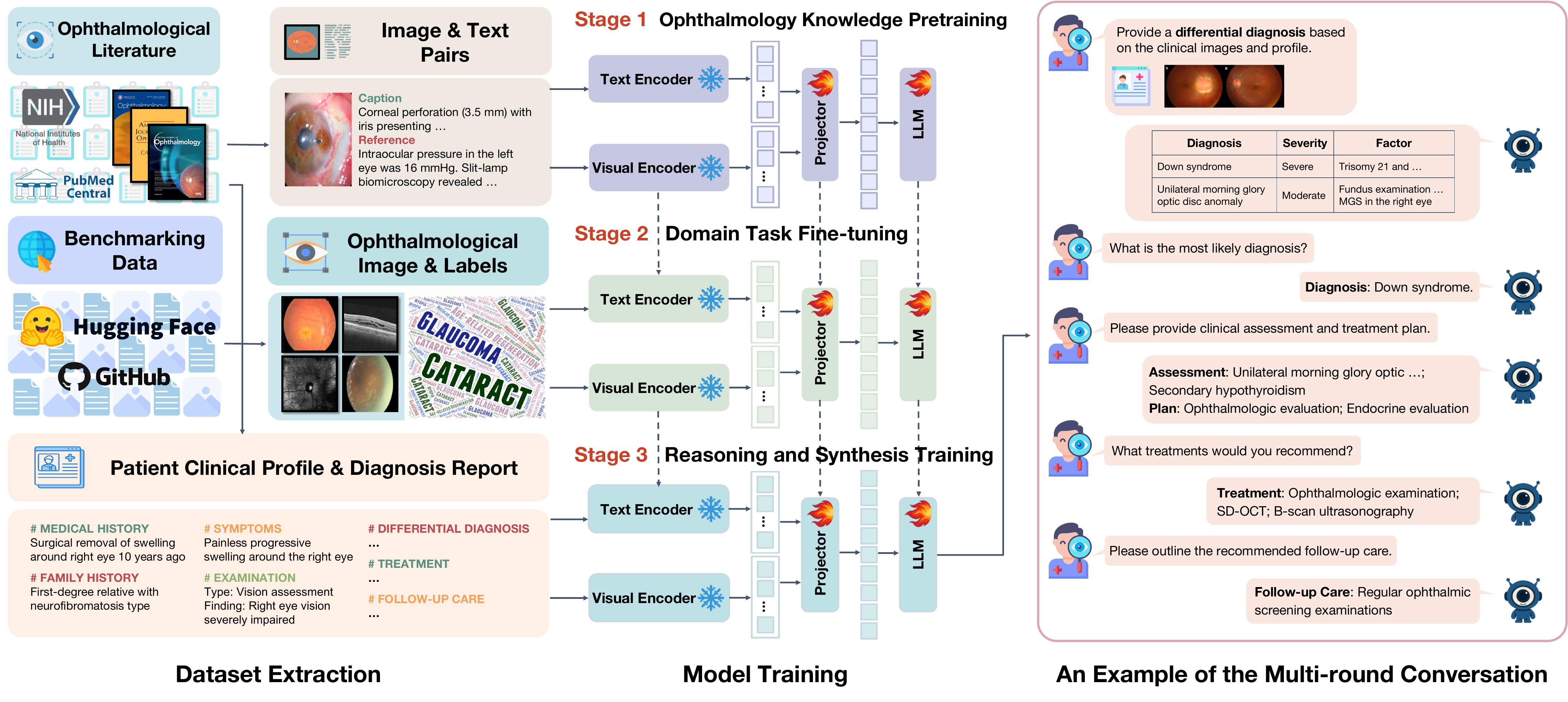}
    \caption{Overview of VOLMO development pipeline and multi-round clinical reasoning. The framework consists of three components: (Left) Dataset Extraction from publicly available sources including ophthalmological literature (PubMed Central), benchmarking datasets (Hugging Face, GitHub, etc.), and patient clinical profiles with diagnosis reports. (Center) Model Training through a three-stage progressive framework: Stage 1 - Ophthalmology knowledge pretraining using \soneimg{} image-text pairs to inject foundational domain knowledge; Stage 2 - Domain task fine-tuning on \stwoins{} disease-labeled instances across \stwocon{} conditions and signs for disease screening and staging; Stage 3 - Reasoning and synthesis training on \sthreerep{} comprehensive case reports to enable clinical assessment generation. Snowflake icons indicate frozen components during training. (Right) Multi-round Conversation Example demonstrating VOLMO's clinical reasoning workflow, where the model sequentially generates differential diagnoses, determines the most likely diagnosis, formulates clinical assessments and treatment plans, recommends specific treatments, and provides follow-up care guidance based on patient clinical profiles and multi-modal ophthalmological imaging.}
    \label{fig:pipeline}
\end{figure}
\begin{figure}
    \centering
    \includegraphics[width=1.0\linewidth]{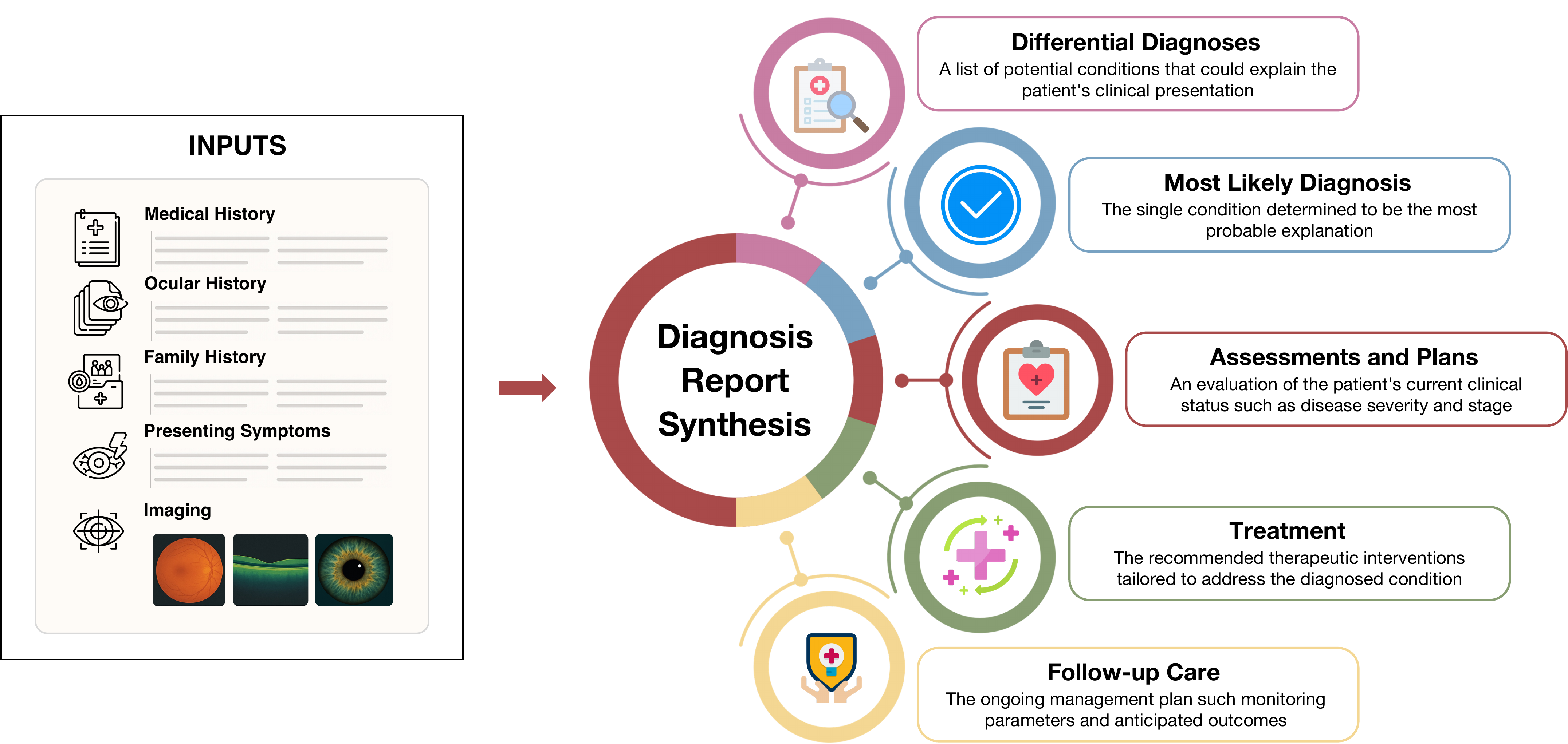}
    \caption{Process of assessment and management generation. Given a patient clinical profile including medical history, ocular history, family history, presenting symptoms, and multi-modal ophthalmology imaging data (left panel), VOLMO generates diagnostic reports through five sequential reasoning stages (right panel): (1) Differential Diagnoses - listing potential conditions that could explain the patient's clinical presentation, (2) Most Likely Diagnosis - the single condition determined to be the most probable explanation with justification, (3) Assessments and Plans - evaluating the patient's current clinical status including disease severity and stage, (4) Treatment - recommending therapeutic interventions tailored to address the diagnosed condition, and (5) Follow-up Care - providing ongoing management plans including monitoring parameters and anticipated outcomes. }
    \label{fig:patient_progress_synthesis}
\end{figure}

\section{Methods}

The overall VOLMO framework is illustrated in Fig.~\ref{fig:pipeline} and comprises three stages: ophthalmology knowledge pretraining, domain task fine-tuning, and multi-step clinical reasoning and synthesis. Each stage uses publicly available datasets with permissive licenses to support the development of openly accessible ophthalmology foundation models that can be used by the community and further adapted with local data. Detailed information on data sources and licensing is provided in~\cref{sec:license_info}. The design and implementation of each stage are described below.

\subsection{Ophthalmology knowledge pretraining}
\label{sec:knowledge_pretrain}

As noted earlier, existing studies show that current MLLMs, both general-domain and medical-domain, lack sufficient ophthalmology-specific knowledge, leading to suboptimal performance on ophthalmological benchmarks and frequent hallucinations~\cite{antaki2024vision,qin2025lmod+}. This stage aims to equip the model with broad foundational knowledge in ophthalmology. However, existing ophthalmology benchmark datasets are not well suited for this purpose: most consist of images without accompanying descriptive narratives, limiting the model's ability to ground visual features in clinical language~\cite{yang2024understanding}, and they often have narrow image modality coverage~\cite{khan2021global}.

To address these gaps, we leveraged large-scale ophthalmology literature to enrich multimodal domain knowledge. We systematically collected \soneart{} full-text PubMed Central (PMC) articles from 82 ophthalmology journals and extracted the associated image--caption pairs. While prior work has directly used captioned images for training~\cite{lin2023pmc,zhang2024biomedclip,lozano2025biomedica}, such captions may contain issues that limit their usefulness. Specifically, they may contain formatting artifacts such as figure references (e.g., \textit{Fig.~3A}), citation markers, or cross-references to other sections. Captions may also contain fragmented sentences or technical abbreviations—e.g., fragments beginning mid-phrase (``A The lesion located in the ciliary body region...'') or unexplained imaging shorthand such as ``T1WI,'' ``T2WI,'' ``T2-tse-fs-cor,'' and ``fs'', resulting in incomplete or ambiguous descriptions. These challenges were also observed in prior work~\cite{chen2024huatuogpt}.

To mitigate these issues, we applied LLM-based augmentation to expand abbreviated content and transform fragmented or highly technical caption text into coherent, natural language descriptions. Details of this augmentation process are provided in \cref{subsec:img_cap_revision}. Importantly, this augmentation was used only for weak supervision; all evaluations were conducted on manually curated datasets. After processing, we obtained a total of \soneimg{} image--text pairs for this pretraining stage.

\begin{table}[ht]
\centering
\caption{Composition of the stage 2 ophthalmological training cohort. The stage 2 dataset comprises 26,929 image–text instances drawn from four independent ophthalmological datasets, covering 12 common eye conditions and signs across diverse patient populations and imaging modalities.}
{
\begin{tabular}{lccc}
\hline
\textbf{Dataset} & \textbf{Images} & \textbf{Modality} & \textbf{Population} \\
\hline
BRSET & 15,201 & CFP & Brazilian \\
OIMHS & 2,933 & OCT & Multi-ethnic \\
FIVES & 240 & CFP & Asian \\
EyePACS & 8,555 & CFP & United States \\
\hline
\end{tabular}
}
\label{tab:train_data_dist}
\end{table}

\subsection{Domain task fine-tuning}
\label{sec:task_tuning}
In this stage, domain task fine-tuning further adapts the model to core ophthalmological applications after acquiring general ophthalmology knowledge. Existing ophthalmology benchmarks were largely designed for earlier architectures such as CNNs or vision foundation models. These datasets typically (1) use single images as input without instructions or prompting, producing only a categorical label as output, and (2) differ in annotation formats and task definitions, requiring separate fine-tuning for individual models. To create a unified training setting for MLLMs, we repurposed 26{,}929 instances from publicly available benchmarks with compatible licenses, covering 12 common ophthalmological conditions and signs: glaucoma, AMD, DR, drusen, hemorrhage, hypertensive retinopathy, increased cup-to-disc ratio, macular edema, myopic fundus, nevus, scar, and vascular occlusion. Details are provided in \cref{tab:train_data_dist}.
Collectively, these datasets span diverse patient populations (Brazilian, Asian, and United States) and imaging modalities (color fundus photography (CFP) and OCT) and support three primary ophthalmological applications: (1) disease screening (binary presence/absence classification), (2) multi-disease classification (predicting multiple disease conditions), and (3) disease staging classification (assigning severity levels).

To train a single unified model across heterogeneous tasks, we developed a standardized instruction–response schema that converts each instance into a structured question–answer pair~\cite{peng2023instruction,keloth2024advancing}. For example, for a DR screening benchmark originally designed as a binary classifier, the transformed instruction is: \textit{``Please tell me whether this image shows diabetic retinopathy. Answer in the format: TRUE or FALSE.''} The model is trained to produce the ground-truth label in natural language. This schema could also extend to new datasets and tasks. Additional details are provided in \cref{subsec:bin_con_screen}.

\subsection{Multi-step clinical reasoning and synthesis}
\label{sec:reasoning}
Although fine-tuning on benchmark datasets is effective for improving task-specific performance, these datasets differ substantially from real ophthalmology workflows~\cite{li2023artificial,chen2025ai}. For instance, eye disease assessment in practice involves far more than interpreting isolated images, as is typical in existing benchmarks. Clinicians integrate multiple sources of information, including patient history, medications, symptoms, and findings from several imaging modalities, to formulate diagnoses, assessments, and management plans~\cite{dow2022data}.

To bridge this gap, this stage enhances the model’s capacity for multi-step reasoning and clinical synthesis by explicitly simulating these workflows. From the ophthalmology journal corpus described above, we identified 913 patient case reports and collected both their free-text narratives and associated multimodal data (e.g., imaging and visual field testing). We developed a structured extraction pipeline to capture nine key clinical elements identified by clinicians: patient medical and ocular history, family history, symptom characterization (onset, duration, progression), diagnostic studies and findings, differential diagnosis with supporting evidence, primary diagnosis with justification, treatment planning with rationale, and follow-up recommendations. These nine elements reflect essential components of ophthalmological clinical evaluation and support the model’s learning of comprehensive diagnostic and management reasoning~\cite{podder2025soap,gagnier2013care}. As in Stage~1, LLMs were used for automatic extraction to provide weak-supervision augmentation during training only, while 40 independent case reports were manually curated for evaluation. Details are provided in \cref{subsec:assess_plan_prompts}. 

To train the model on these complex tasks, the extracted information was further transformed into multi-turn dialogues in which the model is trained to combine structured patient information (e.g., history, medications, symptoms) with multiple examination images across different modalities (e.g., CFP, OCT, visual field testing) and perform sequential clinical reasoning steps (e.g., interpreting findings, formulating assessments, recommending treatments, and outlining long-term care plans), as illustrated in Fig.~\ref{fig:patient_progress_synthesis}. This multi-turn structure simulates patient–clinician interactions and trains the model to perform reasoning in a manner aligned with clinical practice~\cite{li2025beyond,liu2025dialogue}.

\subsection{Model backbones and training details}

As described earlier, the VOLMO framework is model-agnostic and data-open, allowing it to be applied to different MLLM backbones. Prior work has adapted LLaVA-style architectures for medical applications~\cite{li2023llava}. However, LLaVA and similar MLLMs typically rely on fixed, limited input resolutions, commonly 224$\times$224 or 336$\times$336 pixels, which require aggressive downsampling. This constraint can remove fine-grained visual details that are essential for medical imaging. In ophthalmology, subtle findings such as drusen in age-related macular degeneration or early retinal nerve fiber layer thinning in glaucoma may span only a few pixels and therefore require high-fidelity spatial representation~\cite{spaide2010drusen,hood2013glaucoma}. 

To preserve these clinically important details, we adopted the InternVL architecture~\cite{chen2024internvl} as the backbone. InternVL incorporates a dynamic-resolution mechanism that partitions each input image into multiple tiles based on its native resolution, encoding each tile at 448$\times$448 pixels. The model then integrates information across tiles using position-aware attention mechanisms, enabling high-resolution perception while maintaining global contextual understanding.

We trained a 2B-parameter model using the VOLMO framework. Following established practices in the literature~\cite{liu2023visual,chen2024internvl}, the vision encoder was kept frozen during training, while the MLP projector and language-model components were updated (Fig.~\ref{fig:pipeline}). The training was conducted with 4 H100 GPUs.
Detailed training procedures and hyperparameters are provided in~\cref{sec:training_hyperparameters}. All code and trained models are also made publicly available to the community.
For comparison, we additionally performed head-to-head evaluations using InternVL with and without the VOLMO framework, and benchmarked against representative models as described in~\cref{evaluation}.
\section{Evaluations}
\label{evaluation}

We systematically evaluated the VOLMO framework across applications aligned with each stage of the three-stage training pipeline: ophthalmology image-description generation (Stage~1), disease screening and staging classification (Stage~2), and assessment and management generation (Stage~3). In addition to these primary evaluations, we conducted manual expert assessments with two healthcare professionals to evaluate the quality of the generated clinically focused image descriptions, and performed external validations on three independent patient cohorts. Details are presented below.

\begin{figure}
    \centering
    \includegraphics[width=0.85\linewidth]{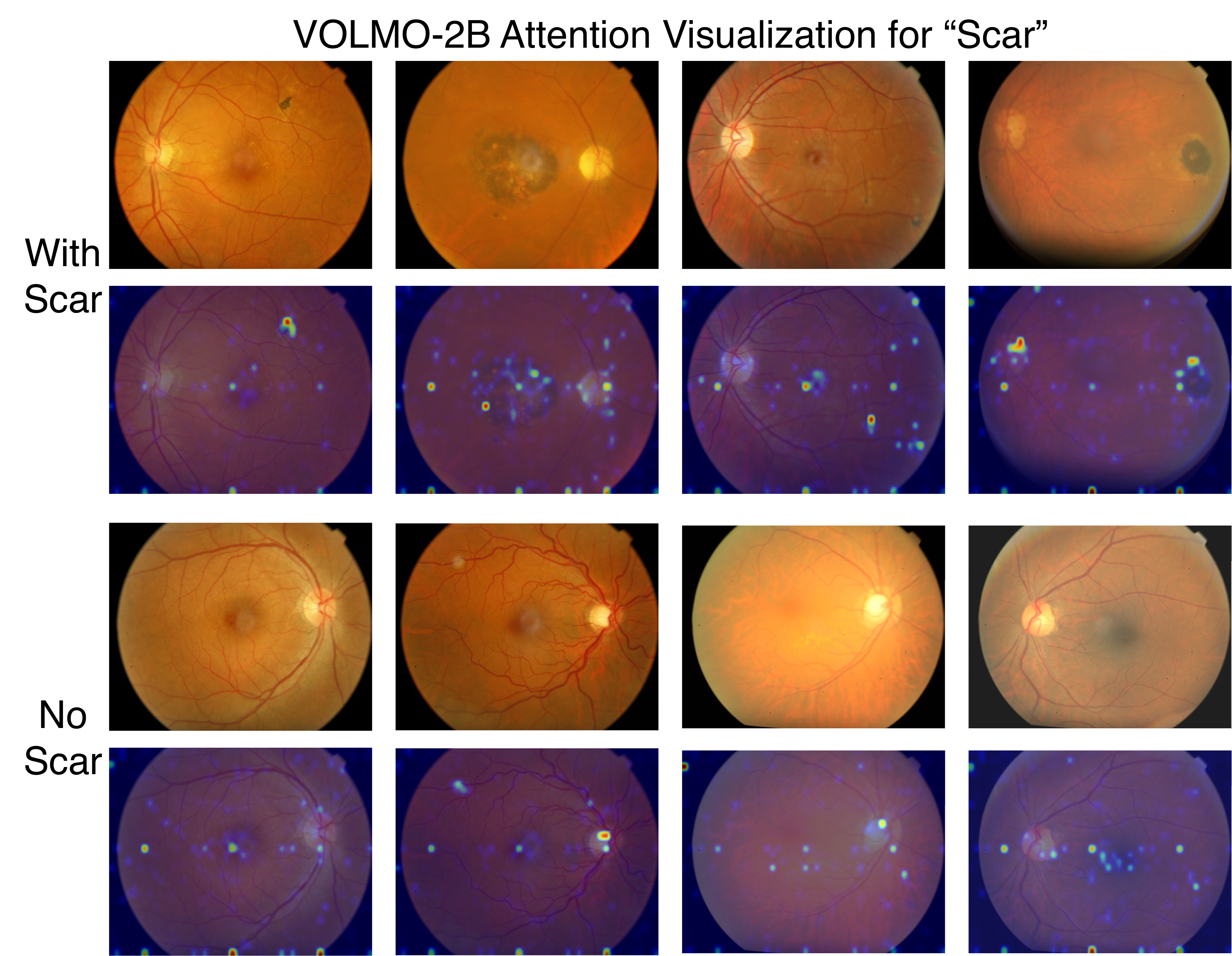}
    \caption{Visual interpretation of VOLMO's attention patterns for the query word \textit{scar}. Top two rows show fundus images with retinal scars (first row) and corresponding attention heatmaps (second row). Bottom two rows display fundus images without scars (third row) and their attention heatmaps (fourth row). Attention intensity is visualized using a color scale ranging from blue (low attention) through green and yellow to red (high attention). }
    \label{fig:heatmap_scar}
\end{figure}
\begin{table*}[tp]
\centering
\caption{Statistical comparison of baseline models against VOLMO-2B for image description tasks. Values show mean $\pm$ standard deviation. P-values from Wilcoxon signed-rank test are shown in parentheses.}
\label{tab:volmo_comparison}
{\scriptsize
\begin{tabular}{l|cccc}
\toprule
\textbf{Model} & \textbf{BLEU-1} & \textbf{ROUGE-L F1} & \textbf{BERTScore F1} & \textbf{SBERT Similarity} \\
\midrule
InternVL-2B & $0.0796 \pm 0.0044$ & $0.1070 \pm 0.0042$ & $0.6464 \pm 0.0032$ & $0.4409 \pm 0.0092$ \\
& ($p < 0.0001$) & ($p < 0.0001$) & ($p < 0.0001$) & ($p < 0.0001$) \\
\midrule
LLaVA-Med & $0.1181 \pm 0.0068$ & $0.1644 \pm 0.0037$ & $0.6867 \pm 0.0028$ & $0.4468 \pm 0.0117$ \\
& ($p < 0.0001$) & ($p < 0.0001$) & ($p < 0.0001$) & ($p < 0.0001$) \\
\midrule
MedGemma-4B & $0.1083 \pm 0.0068$ & $0.1212 \pm 0.0052$ & $0.6428 \pm 0.0034$ & $0.4669 \pm 0.0090$ \\
& ($p < 0.0001$) & ($p < 0.0001$) & ($p < 0.0001$) & ($p < 0.0001$) \\
\midrule
MedGemma-27B & $0.0912 \pm 0.0057$ & $0.1114 \pm 0.0043$ & $0.6397 \pm 0.0028$ & $\mathbf{0.4855 \pm 0.0083}$ \\
& ($p < 0.0001$) & ($p < 0.0001$) & ($p < 0.0001$) & ($p < 0.0001$) \\
\midrule
VOLMO-2B & $\mathbf{0.1741 \pm 0.0080}$ & $\mathbf{0.2170 \pm 0.0074}$ & $\mathbf{0.7140 \pm 0.0042}$ & $0.4727 \pm 0.0113$ \\
\bottomrule
\end{tabular}
}
\end{table*}

\FloatBarrier 
{\scriptsize
\setlength{\tabcolsep}{2.5pt}

\begin{longtable}{c|ccc||ccc||ccc}
\caption{Statistical comparison of baseline models against VOLMO-2B for binary condition and sign classification tasks. Values show mean $\pm$ standard deviation. All values are percentages (0--100) without the percent sign. P-values from Wilcoxon signed-rank test vs.\ VOLMO-2B are shown in parentheses. The best F1 score in each column is highlighted in bold.}%
\label{tab:volmo_classification_main}\\

\toprule
\textbf{Model} &
\textbf{F1} & \textbf{Sensitivity} & \textbf{Specificity} &
\textbf{F1} & \textbf{Sensitivity} & \textbf{Specificity} &
\textbf{F1} & \textbf{Sensitivity} & \textbf{Specificity} \\
\midrule
\endfirsthead

\multicolumn{10}{l}{\small\itshape Table~\thetable\ (continued)}\\
\toprule
\textbf{Model} &
\textbf{F1} & \textbf{Sensitivity} & \textbf{Specificity} &
\textbf{F1} & \textbf{Sensitivity} & \textbf{Specificity} &
\textbf{F1} & \textbf{Sensitivity} & \textbf{Specificity} \\
\midrule
\endhead

\endfoot

\bottomrule
\endlastfoot

& \multicolumn{3}{c||}{\textbf{Glaucoma}} &
  \multicolumn{3}{c||}{\textbf{AMD}} &
  \multicolumn{3}{c}{\textbf{DR}} \\
RETFound & $\mathbf{92.33}$ & $92.33$ & $92.71$
         & $87.91$ & $87.93$ & $87.87$
         & $84.94$ & $84.93$ & $85.15$ \\
& $(p = 0.3172)$ & $(p = 0.2324)$ & $(p = 0.3814)$
& $(p = 0.0027)$ & $(p < 0.0001)$ & $(p = 0.0008)$
& $(p < 0.0001)$ & $(p < 0.0001)$ & $(p < 0.0001)$ \\
\midrule
InternVL-2B & $31.79$ & $48.37$ & $51.63$
            & $35.46$ & $50.90$ & $50.77$
            & $33.11$ & $49.23$ & $50.77$ \\
& $(p < 0.0001)$ & $(p < 0.0001)$ & $(p < 0.0001)$
& $(p < 0.0001)$ & $(p < 0.0001)$ & $(p < 0.0001)$
& $(p < 0.0001)$ & $(p < 0.0001)$ & $(p < 0.0001)$ \\
\midrule
LLaVA-Med & $34.44$ & $50.76$ & $49.24$
          & $33.78$ & $50.10$ & $49.90$
          & $39.24$ & $54.67$ & $45.33$ \\
& $(p < 0.0001)$ & $(p < 0.0001)$ & $(p < 0.0001)$
& $(p < 0.0001)$ & $(p < 0.0001)$ & $(p < 0.0001)$
& $(p < 0.0001)$ & $(p < 0.0001)$ & $(p < 0.0001)$ \\
\midrule
MedGemma-4B & $63.74$ & $67.27$ & $69.22$
            & $34.10$ & $50.38$ & $49.62$
            & $\mathbf{94.25}$ & $94.25$ & $94.25$ \\
& $(p < 0.0001)$ & $(p < 0.0001)$ & $(p < 0.0001)$
& $(p < 0.0001)$ & $(p < 0.0001)$ & $(p < 0.0001)$
& $(p = 0.0013)$ & $(p = 0.0003)$ & $(p = 0.0007)$ \\
\midrule
MedGemma-27B & $32.78$ & $49.26$ & $50.74$
             & $55.92$ & $61.80$ & $61.94$
             & $92.83$ & $92.83$ & $92.78$ \\
& $(p < 0.0001)$ & $(p < 0.0001)$ & $(p < 0.0001)$
& $(p < 0.0001)$ & $(p < 0.0001)$ & $(p < 0.0001)$
& $(p = 0.5135)$ & $(p = 0.2836)$ & $(p = 0.4062)$ \\
\midrule
VOLMO-2B & $91.89$ & $91.90$ & $92.31$
         & $\mathbf{90.34}$ & $90.43$ & $90.26$
         & $92.31$ & $92.30$ & $92.27$ \\
\midrule

& \multicolumn{3}{c||}{\textbf{Drusen}} &
  \multicolumn{3}{c||}{\textbf{Hemorrhage}} &
  \multicolumn{3}{c}{\textbf{Hypertensive Retinopathy}} \\
RETFound & $68.98$ & $69.03$ & $69.22$
         & $\mathbf{79.43}$ & $79.87$ & $79.89$
         & $63.34$ & $64.03$ & $64.46$ \\
& $(p < 0.0001)$ & $(p < 0.0001)$ & $(p < 0.0001)$
& $(p < 0.0001)$ & $(p < 0.0001)$ & $(p < 0.0001)$
& $(p < 0.0001)$ & $(p < 0.0001)$ & $(p < 0.0001)$ \\
\midrule
InternVL-2B & $34.17$ & $49.87$ & $50.39$
            & $33.39$ & $50.00$ & $50.00$
            & $34.30$ & $50.53$ & $49.47$ \\
& $(p < 0.0001)$ & $(p < 0.0001)$ & $(p < 0.0001)$
& $(p < 0.0001)$ & $(p < 0.0001)$ & $(p < 0.0001)$
& $(p < 0.0001)$ & $(p < 0.0001)$ & $(p < 0.0001)$ \\
\midrule
LLaVA-Med & $34.02$ & $50.30$ & $49.70$
          & $33.39$ & $50.00$ & $50.00$
          & $51.56$ & $60.89$ & $57.15$ \\
& $(p < 0.0001)$ & $(p < 0.0001)$ & $(p < 0.0001)$
& $(p < 0.0001)$ & $(p < 0.0001)$ & $(p < 0.0001)$
& $(p < 0.0001)$ & $(p < 0.0001)$ & $(p < 0.0001)$ \\
\midrule
MedGemma-4B & $56.63$ & $60.07$ & $60.08$
            & $76.11$ & $76.47$ & $76.48$
            & $39.68$ & $50.43$ & $49.41$ \\
& $(p < 0.0001)$ & $(p < 0.0001)$ & $(p < 0.0001)$
& $(p = 0.0301)$ & $(p = 0.7668)$ & $(p = 0.5507)$
& $(p < 0.0001)$ & $(p < 0.0001)$ & $(p < 0.0001)$ \\
\midrule
MedGemma-27B & $44.95$ & $53.67$ & $54.09$
             & $76.37$ & $76.47$ & $76.45$
             & $42.27$ & $51.90$ & $50.88$ \\
& $(p < 0.0001)$ & $(p < 0.0001)$ & $(p < 0.0001)$
& $(p = 0.5348)$ & $(p = 0.9151)$ & $(p = 0.7456)$
& $(p < 0.0001)$ & $(p < 0.0001)$ & $(p < 0.0001)$ \\
\midrule
VOLMO-2B & $\mathbf{84.47}$ & $84.53$ & $84.45$
         & $76.37$ & $76.47$ & $76.45$
         & $\mathbf{70.42}$ & $70.43$ & $70.20$ \\
\midrule

& \multicolumn{3}{c||}{\textbf{Increased Cup-Disc}} &
  \multicolumn{3}{c||}{\textbf{Macular Edema}} &
  \multicolumn{3}{c}{\textbf{Myopic Fundus}} \\
RETFound & $72.58$ & $72.77$ & $73.11$
         & $87.30$ & $87.30$ & $87.21$
         & $98.00$ & $98.00$ & $98.05$ \\
& $(p < 0.0001)$ & $(p < 0.0001)$ & $(p < 0.0001)$
& $(p < 0.0001)$ & $(p < 0.0001)$ & $(p < 0.0001)$
& $(p < 0.0001)$ & $(p < 0.0001)$ & $(p < 0.0001)$ \\
\midrule
InternVL-2B & $32.92$ & $49.20$ & $50.80$
            & $45.70$ & $53.97$ & $54.68$
            & $35.85$ & $51.07$ & $50.93$ \\
& $(p < 0.0001)$ & $(p < 0.0001)$ & $(p < 0.0001)$
& $(p < 0.0001)$ & $(p < 0.0001)$ & $(p < 0.0001)$
& $(p < 0.0001)$ & $(p < 0.0001)$ & $(p < 0.0001)$ \\
\midrule
LLaVA-Med & $36.10$ & $50.57$ & $49.04$
          & $34.33$ & $50.43$ & $49.57$
          & $36.13$ & $46.23$ & $31.55$ \\
& $(p < 0.0001)$ & $(p < 0.0001)$ & $(p < 0.0001)$
& $(p < 0.0001)$ & $(p < 0.0001)$ & $(p < 0.0001)$
& $(p < 0.0001)$ & $(p < 0.0001)$ & $(p < 0.0001)$ \\
\midrule
MedGemma-4B & $39.50$ & $52.49$ & $51.95$
            & $91.57$ & $91.57$ & $91.71$
            & $45.89$ & $59.64$ & $38.28$ \\
& $(p < 0.0001)$ & $(p < 0.0001)$ & $(p < 0.0001)$
& $(p < 0.0001)$ & $(p < 0.0001)$ & $(p < 0.0001)$
& $(p < 0.0001)$ & $(p < 0.0001)$ & $(p < 0.0001)$ \\
\midrule
MedGemma-27B & $35.32$ & $49.20$ & $51.67$
             & $92.77$ & $92.77$ & $92.62$
             & $52.71$ & $52.80$ & $53.04$ \\
& $(p < 0.0001)$ & $(p < 0.0001)$ & $(p < 0.0001)$
& $(p < 0.0001)$ & $(p < 0.0001)$ & $(p < 0.0001)$
& $(p < 0.0001)$ & $(p < 0.0001)$ & $(p < 0.0001)$ \\
\midrule
VOLMO-2B & $\mathbf{79.83}$ & $79.83$ & $79.94$
         & $\mathbf{96.00}$ & $96.00$ & $96.00$
         & $\mathbf{98.97}$ & $98.97$ & $98.99$ \\
\midrule

& \multicolumn{3}{c||}{\textbf{Nevus}} &
  \multicolumn{3}{c||}{\textbf{Scar}} &
  \multicolumn{3}{c}{\textbf{Vascular Occlusion}} \\
RETFound & $60.07$ & $60.27$ & $59.70$
         & $80.65$ & $80.83$ & $81.31$
         & $\mathbf{94.85}$ & $94.87$ & $94.92$ \\
& $(p < 0.0001)$ & $(p < 0.0001)$ & $(p < 0.0001)$
& $(p < 0.0001)$ & $(p < 0.0001)$ & $(p < 0.0001)$
& $(p < 0.0001)$ & $(p < 0.0001)$ & $(p < 0.0001)$ \\
\midrule
InternVL-2B & $34.50$ & $50.83$ & $49.17$
            & $33.18$ & $49.50$ & $50.50$
            & $33.63$ & $50.13$ & $49.87$ \\
& $(p < 0.0001)$ & $(p < 0.0001)$ & $(p < 0.0001)$
& $(p < 0.0001)$ & $(p < 0.0001)$ & $(p < 0.0001)$
& $(p < 0.0001)$ & $(p < 0.0001)$ & $(p < 0.0001)$ \\
\midrule
LLaVA-Med & $35.42$ & $49.63$ & $54.73$
          & $34.27$ & $50.50$ & $49.50$
          & $59.67$ & $62.69$ & $62.66$ \\
& $(p < 0.0001)$ & $(p < 0.0001)$ & $(p < 0.0001)$
& $(p < 0.0001)$ & $(p < 0.0001)$ & $(p < 0.0001)$
& $(p < 0.0001)$ & $(p < 0.0001)$ & $(p < 0.0001)$ \\
\midrule
MedGemma-4B & $34.50$ & $50.83$ & $49.17$
            & $74.20$ & $75.06$ & $75.35$
            & $77.65$ & $78.73$ & $78.58$ \\
& $(p < 0.0001)$ & $(p < 0.0001)$ & $(p < 0.0001)$
& $(p < 0.0001)$ & $(p < 0.0001)$ & $(p < 0.0001)$
& $(p < 0.0001)$ & $(p < 0.0001)$ & $(p < 0.0001)$ \\
\midrule
MedGemma-27B & $45.31$ & $55.51$ & $52.36$
             & $82.10$ & $82.17$ & $82.55$
             & $87.72$ & $87.73$ & $87.62$ \\
& $(p < 0.0001)$ & $(p < 0.0001)$ & $(p < 0.0001)$
& $(p < 0.0001)$ & $(p < 0.0001)$ & $(p < 0.0001)$
& $(p = 0.6724)$ & $(p = 0.8575)$ & $(p = 0.6133)$ \\
\midrule
VOLMO-2B & $\mathbf{84.46}$ & $84.47$ & $84.79$
         & $\mathbf{96.07}$ & $96.07$ & $96.06$
         & $87.80$ & $87.87$ & $87.96$ \\

\end{longtable}
}
\FloatBarrier

\begin{table*}[t]
\centering
\caption{External validation performance on UK Biobank cohort for AMD and DR screening. Values show mean F1-score, sensitivity, and specificity ($\pm$ standard deviation) across bootstrap resampling. P-values from Wilcoxon signed-rank tests compare each baseline model against VOLMO-2B. }
\label{tab:ukbiobank_classification}
{\scriptsize
\setlength{\tabcolsep}{3.5pt}
\begin{tabular}{c|ccc||ccc}
\toprule
& \multicolumn{3}{c||}{\textbf{AMD}} & \multicolumn{3}{c}{\textbf{DR}} \\
\textbf{Model} & \textbf{F1} & \textbf{Sensitivity} & \textbf{Specificity} & \textbf{F1} & \textbf{Sensitivity} & \textbf{Specificity} \\
\midrule
RETFound & $46.39 \pm 9.95$ & $55.33 \pm 7.93$ & $56.33 \pm 7.79$ & $53.36 \pm 9.77$ & $56.83 \pm 8.79$ & $67.30 \pm 7.99$ \\
& $(p < 0.0001)$ & $(p < 0.0001)$ & $(p < 0.0001)$ & $(p = 0.0004)$ & $(p = 0.0025)$ & $(p = 0.0130)$ \\
\midrule
InternVL-2B & $41.47 \pm 10.10$ & $53.07 \pm 8.36$ & $52.06 \pm 7.59$ & $24.69 \pm 8.21$ & $41.37 \pm 8.28$ & $58.63 \pm 8.28$ \\
& $(p < 0.0001)$ & $(p < 0.0001)$ & $(p < 0.0001)$ & $(p < 0.0001)$ & $(p < 0.0001)$ & $(p < 0.0001)$ \\
\midrule
LLaVA-Med & $34.41 \pm 8.69$ & $50.63 \pm 7.90$ & $49.37 \pm 7.90$ & $43.77 \pm 9.85$ & $58.71 \pm 8.19$ & $41.29 \pm 8.19$ \\
& $(p < 0.0001)$ & $(p < 0.0001)$ & $(p < 0.0001)$ & $(p = 0.0002)$ & $(p = 0.0032)$ & $(p < 0.0001)$ \\
\midrule
MedGemma-4B & $32.94 \pm 8.86$ & $49.30 \pm 7.98$ & $50.70 \pm 7.98$ & $37.17 \pm 8.23$ & $47.90 \pm 7.33$ & $62.95 \pm 8.73$ \\
& $(p < 0.0001)$ & $(p < 0.0001)$ & $(p < 0.0001)$ & $(p < 0.0001)$ & $(p < 0.0001)$ & $(p = 0.0001)$ \\
\midrule
MedGemma-27B & $33.77 \pm 9.05$ & $49.50 \pm 8.02$ & $50.74 \pm 7.89$ & $40.16 \pm 9.37$ & $50.31 \pm 7.88$ & $62.27 \pm 8.97$ \\
& $(p < 0.0001)$ & $(p < 0.0001)$ & $(p < 0.0001)$ & $(p < 0.0001)$ & $(p < 0.0001)$ & $(p < 0.0001)$ \\
\midrule
VOLMO-2B & $\mathbf{64.58 \pm 9.30}$ & $65.93 \pm 8.69$ & $66.75 \pm 8.50$ & $\mathbf{49.86 \pm 9.06}$ & $54.20 \pm 7.82$ & $65.12 \pm 8.73$ \\
\bottomrule
\end{tabular}
}
\end{table*}

\begin{table*}[t]
\centering
\caption{Disease stage grading performance across multiple datasets. We report F1, Sensitivity, and Specificity for each disease stage. In-domain datasets were used during training, while cross-domain datasets evaluate generalization to unseen data sources. Dashes indicate stages not present in the dataset. Best Overall F1 in each dataset block is bolded.}
\label{tab:disease_stage}

{\scriptsize
\setlength{\tabcolsep}{1.2pt}
\begin{tabular}{l|rrr|rrr|rrr|rrr|rrr|rrr}
\toprule
\textbf{Method}
& \multicolumn{3}{c|}{\textbf{Overall}}
& \multicolumn{3}{c|}{\textbf{Stage 0}}
& \multicolumn{3}{c|}{\textbf{Stage 1}}
& \multicolumn{3}{c|}{\textbf{Stage 2}}
& \multicolumn{3}{c|}{\textbf{Stage 3}}
& \multicolumn{3}{c}{\textbf{Stage 4}} \\
\cmidrule(lr){2-4}\cmidrule(lr){5-7}\cmidrule(lr){8-10}\cmidrule(lr){11-13}\cmidrule(lr){14-16}\cmidrule(lr){17-19}
& \textbf{F1} & \textbf{Sens} & \textbf{Spec}
& \textbf{F1} & \textbf{Sens} & \textbf{Spec}
& \textbf{F1} & \textbf{Sens} & \textbf{Spec}
& \textbf{F1} & \textbf{Sens} & \textbf{Spec}
& \textbf{F1} & \textbf{Sens} & \textbf{Spec}
& \textbf{F1} & \textbf{Sens} & \textbf{Spec} \\
\midrule
\multicolumn{19}{c}{\textbf{Macular Hole}} \\
\midrule
InternVL-2B   &  5.47 &  3.67 & 94.83 &    -- &    -- &    -- &    -- &    -- &    -- & 13.81 &  9.67 & 84.83 &  2.61 &  1.33 & 99.67 &  0.00 &  0.00 & 100.00 \\
LLaVA-Med     & 16.67 & 33.33 & 66.67 &    -- &    -- &    -- &    -- &    -- &    -- &  0.00 &  0.00 & 100.00 &  0.00 &  0.00 & 100.00 & 50.00 & 100.00 &   0.00 \\
MedGemma-27B  & 22.00 & 30.57 & 66.31 &    -- &    -- &    -- &    -- &    -- &    -- & 11.11 &  6.96 & 88.72 & 48.78 & 81.08 & 16.10 &  6.11 &  3.67 &  94.12 \\
MedGemma-4B   & 19.84 & 18.04 & 82.32 &    -- &    -- &    -- &    -- &    -- &    -- & 25.25 & 21.33 & 76.13 & 34.27 & 32.78 & 70.83 &  0.00 &  0.00 & 100.00 \\
RETFound      & 68.12 & 68.22 & 84.11 &    -- &    -- &    -- &    -- &    -- &    -- & 66.25 & 70.00 & 79.33 & 64.14 & 59.33 & 87.17 & 73.98 & 75.33 &  85.83 \\
\midrule 
VOLMO-2B      & \textbf{92.62} & 92.67 & 96.33 &    -- &    -- &    -- &    -- &    -- &    -- & 91.13 & 85.67 & 98.83 & 92.18 & 94.33 & 94.83 & 94.53 & 98.00 &  95.33 \\
\midrule
\multicolumn{19}{c}{\textbf{EyePACS}} \\
\midrule
InternVL-2B   &  0.96 & 20.00 & 80.00 &  0.00 &  0.00 & 100.00 &  0.00 &  0.00 & 100.00 &  0.00 &  0.00 & 100.00 &  4.80 & 100.00 &  0.00 &  0.00 &  0.00 & 100.00 \\
LLaVA-Med     &  0.00 &  0.00 & 100.00 &  0.00 &  0.00 & 100.00 &  0.00 &  0.00 & 100.00 &  0.00 &  0.00 & 100.00 &  0.00 &  0.00 & 100.00 &  0.00 &  0.00 & 100.00 \\
MedGemma-27B  & 39.69 & 45.41 & 90.02 & 92.87 & 98.17 & 61.48 & 12.69 &  8.39 & 98.39 & 36.00 & 23.91 & 98.48 & 14.33 & 19.72 & 96.08 & 42.56 & 76.85 & 95.69 \\
MedGemma-4B   & 45.87 & 50.40 & 91.48 & 93.08 & 95.62 & 71.40 & 22.08 & 18.75 & 96.56 & 43.21 & 31.99 & 97.27 & 21.32 & 37.16 & 94.67 & 49.64 & 68.47 & 97.49 \\
RETFound      & 43.42 & 48.26 & 87.05 & 83.30 & 81.99 & 56.77 & 11.85 & 10.36 & 95.66 & 34.75 & 35.17 & 88.58 & 32.77 & 53.21 & 95.68 & 54.42 & 60.59 & 98.55 \\
\midrule 
VOLMO-2B      & \textbf{46.80} & 57.19 & 89.05 & 78.08 & 68.45 & 79.95 & 18.60 & 44.29 & 77.64 & 44.78 & 39.44 & 93.76 & 40.35 & 62.84 & 96.26 & 52.17 & 70.94 & 97.63 \\
\bottomrule
\end{tabular}
}
\end{table*}

\subsection{Ophthalmology image-description generation}
\label{image-description}
This task evaluates the effectiveness of models in generating meaningful descriptions of ophthalmological images. As mentioned, a key step in clinical workflows involves clinicians manually documenting imaging findings~\cite{sanders2013electronic,tailor2025evaluation}. Image-description generation serves as the primary evaluation for Stage~1 of the VOLMO framework.

\subsubsection*{Test Data} 

We manually curated 40 image–caption pairs from the ophthalmology literature described in Stage~1 (Section~\ref{sec:knowledge_pretrain}). Recall that LLMs were used to revise raw captions for augmentation during training. For evaluation, we manually curated a separate set of instances to ensure accuracy and quality. All selected pairs were sourced from articles distinct from those used for model training.

Manual assessment comprised two dimensions:  
(1) \textit{revision factuality}, evaluating whether the revised caption accurately described the original content without introducing factual errors or hallucinations; and  
(2) \textit{revision quality}, assessing the clinical relevance, completeness, and professional clarity of the caption on a five-point scale (1 = poor, 5 = excellent).  

Detailed evaluation guidelines are provided in \cref{sec:imgdesc_annotation_guidelines}. Only image–caption pairs receiving full marks (score = 5) in both dimensions were included in the final test set.

\subsubsection*{Evaluation Metrics}

We evaluated generated captions using established text-generation metrics~\cite{xie2025medical,shool2025systematic,van2024field}. Following standard practice, we employed both overlap-based metrics BLEU~\cite{papineni2002bleu} and ROUGE~\cite{lin2004rouge}, which assess lexical similarity by measuring how many words or sequences in the generated text match the gold-standard reference, and semantic-based metrics BERTScore~\cite{zhang2019bertscore} and Sentence-BERT (SBERT) embeddings~\cite{reimers2019sentence} which evaluate whether the generated text conveys similar meaning by comparing contextualized or sentence-level embeddings.
For overlap-based metrics, BLEU-1 was used to assess token-level lexical matches between the model-generated caption and the reference caption, whereas ROUGE-L captures sequence-level similarity by measuring the longest common subsequences across entire descriptions.
For semantic-based metrics, BERTScore computes similarity by comparing contextualized token embeddings between the generated and reference captions, providing fine-grained semantic alignment. SBERT, in contrast, embeds entire sentences or passages into a unified semantic space and measures cosine similarity, capturing global semantic coherence beyond word-level~\cite{reimers2019sentence,chen2018sentence}.

While automatic metrics are widely used for scalable evaluations, they have known limitations and may not always reflect human preferences, particularly in medical contexts~\cite{fleming2024medalign,chen2025benchmarking}. We also conducted manual evaluations by two healthcare professionals. Details are summarized in Section~\ref{sec:manual} below.

\subsection*{Baseline Models}

VOLMO-2B was systematically compared against the following representative models for this task:

\textbf{InternVL-2B}~\cite{chen2024internvl} is a general-domain MLLM that achieves state-of-the-art performance across diverse multimodal benchmarks~\cite{chen2024internvl}. InternVL also serves as the backbone architecture for VOLMO-2B, making this a direct head-to-head comparison to quantify the effectiveness of the VOLMO framework.

\textbf{LLaVA-Med-7B}~\cite{li2023llava} is a widely used medical MLLM and one of the earliest models pretrained on large-scale medical image--text corpora (approximately 15 million pairs). It demonstrates strong performance across medical vision--language benchmarks and is commonly adopted as a baseline for multimodal medical applications.

\textbf{MedGemma-4B}~\cite{sellergren2025medgemma} is a recent MLLM built on the Gemma-3 architecture. Its image encoder is pretrained on multiple medical imaging modalities, including chest radiographs and ophthalmological images, and its language component is further trained on medical text such as question–answer pairs. It represents a strong mid-sized baseline for domain-specific multimodal medical tasks.

\textbf{MedGemma-27B}~\cite{sellergren2025medgemma} is the largest model in the MedGemma family, with 27 billion parameters and both text-only and multimodal variants (the multimodal version is used in this study). It incorporates additional pretraining on electronic health record data and is optimized for inference-time reasoning. By comparing both the 4B and 27B variants with VOLMO-2B, we evaluate whether a compact model trained using a domain-specific framework can exceed the performance of substantially larger general or medical MLLMs.

\subsection{Disease screening and staging classification}

We next evaluated two core clinical tasks: \textit{disease screening}, which determines the presence or absence of a condition (binary classification), and \textit{disease staging}, which assesses disease severity across multiple levels. Both tasks are central to ophthalmological diagnosis and management~\cite{mills2006categorizing,abacha2023investigation,ting2021artificial,deng2025current} and constitute the primary evaluation for Stage~2 of the VOLMO framework.

\subsubsection*{Test Data}

The standard test splits from the repurposed benchmark datasets described in Stage~2 (Section~\ref{sec:task_tuning}) were used for evaluation. More details are provided in \cref{tab:train_data_dist} as mentioned.
For disease screening, models were evaluated on 3{,}596 test instances covering 12 ophthalmological conditions and signs.  
For disease staging classification, we used 8{,}871 test instances covering macular hole severity (Stages~2--4) and DR severity (Stages~0--4). Notably, the original dataset on macular hole included only 19 cases labeled as Stages~0 and~1. Due to the extremely limited sample size, these early stages were excluded from statistical reporting, and analyses focused on Stages~2--4.  

\subsubsection*{Evaluation Metrics} 

We evaluated model performance using standard classification metrics commonly adopted in ophthalmology and medical  applications~\cite{lu2018applications,van2024performance,rajpurkar2022ai}
.  
The primary metric was the F1-score, which balances precision and recall.  We also report Sensitivity (true positive rate) and Specificity (true negative rate), as additional measures.

\subsection*{Baseline Models}

We compared VOLMO-2B with the four MLLM baselines described above and additionally included RETFound~\cite{zhou2023foundation} for both disease screening and staging classification tasks. As mentioned, RETFound is a state-of-the-art ophthalmology vision foundation model pretrained on 1.6 million retinal images using self-supervised learning. It has demonstrated strong performance across ophthalmological classification tasks and is widely adopted in downstream studies~\cite{zhang2024retfound,chen2025independent,hou2026can}. 

Unlike MLLMs, RETFound is built on a vision-transformer backbone and is designed for discriminative tasks rather than generative outputs. It therefore requires task-specific fine-tuning. Following the procedures established in the original study~\cite{zhou2023foundation}, we fine-tuned RETFound separately for each disease condition, using customized classification heads, 14 models in total (12 for disease screening and 2 for severity classification).
We evaluated both frozen and unfrozen fine-tuning strategies. In the frozen setting, the pretrained encoder weights were kept fixed and only the classification head was trained; in the unfrozen setting, all model parameters were updated. The frozen approach had higher average performance across the 12 disease-screening conditions and features (macro-F1: 80.87\% vs.\ 78.17\%; see \cref{sec:retfound_frozen} for details). Therefore, we report frozen RETFound results in all main comparisons.

\subsection{Assessment-and-Management generation}

This task evaluates the effectiveness of models in automatically generating clinical assessments and management plans for ophthalmology patients. In clinical practice, this is a complex and high-value task in which clinicians manually synthesize heterogeneous information, including imaging from multiple modalities, examination findings, and patient history, to formulate diagnostic impressions and treatment plans~\cite{dow2022data,holland2025specialized}. This task serves as the primary evaluation for Stage~3 of the VOLMO framework.

\subsubsection*{Test Data}

From the patient case corpus curated in Stage~3 (Section~\ref{sec:reasoning}), a subset of 40 independent patient cases was manually annotated for evaluation. These cases were not used anywhere in Stage~3 training. 
Two attending-level ophthalmologists manually curated:  
(1) \textit{Clinical Assessment and Plan}, providing a detailed diagnostic assessment and management plan for each case, and  
(2) \textit{Treatment Recommendations and (3) Follow-up Care}, specifying appropriate interventions and follow-up strategies.  
Annotation followed structured guidelines (see Appendix~\ref{sec:imgdesc_annotation_guidelines} and~\ref{sec:pmc_annotation_guidelines}). 
Any discrepancies between annotators were resolved through iterative discussion until consensus was reached.

\subsection*{Baseline Models}

The four MLLM baselines described above were used for comparison.

\subsubsection*{Evaluations}

For each case, the corresponding patient profile, including medical and family history, multimodal images, and other relevant clinical findings, was provided as input. Each model generated both the clinical assessment and the management plan.  
Consistent with the ophthalmology image-description task, we evaluated outputs using standard text-generation metrics. 

\subsubsection*{Manual and External Evaluations}
\label{sec:manual}

In addition to the evaluations described above, we further conducted  
(1) manual expert assessments of ophthalmology image-description generation and  
(2) external validation of disease screening and staging classification performance on independent patient cohorts, as detailed below.

\subsubsection{Manual Evaluation of Ophthalmology Image-Description Generation}

Generated ophthalmology image descriptions from Section~\ref{image-description} were evaluated manually across three dimensions:  
(1) Conciseness: precision without unnecessary verbosity;  
(2) Accuracy: factual correctness and absence of hallucinations; and  
(3) Readability: clarity, fluency, and appropriateness of clinical tone.  
Each dimension was rated on a 5-point scale. These criteria are widely used in clinician-centered evaluations of LLM outputs in the medical domain~\cite{singhal2025toward,wan2024outpatient}.  
Detailed evaluation guidelines are provided in \cref{sec:imgdesc_annotation_guidelines}.  
To minimize bias, model identities were anonymized, and the evaluation order was randomized.  
Two ophthalmology residents independently assessed all 40 samples.

\subsubsection{External Evaluation of Disease Screening and Staging Classification}

We additionally performed external validation for both disease screening and disease staging classification.  
Models were directly applied to independent patient cohorts without any retraining or adaptation.

For disease screening, we used the UK Biobank (UKB) cohort, one of the largest population-based studies with multimodal imaging ~\cite{sudlow2015uk,littlejohns2020uk}.  
Following prior UKB ophthalmology studies~\cite{warwick2023uk,chua2019cohort}, we curated a random subset comprising  
241 images of DR, and 400 images of AMD. 

For disease staging classification, we further evaluated DR severity classification on two external patient cohorts: 7{,}982 patients from the Sydney Innovation dataset~\cite{kaggle2019sydneyinnovation} and 1{,}151 patients from the SUSTech dataset~\cite{lin2020sustech}.  
The Sydney Innovation dataset originates from the 2019 Diabetic Retinopathy Innovation Challenge, and the patients were collected across multiple Australian eye clinics to support automated DR detection research.  
The SUSTech dataset is a large-scale Chinese retinal image repository designed for benchmarking DR grading algorithms.  
The stage distributions reflected clinical prevalence in each cohort: Sydney Innovation contained 5{,}754 stage~0, 527 stage~1, 1{,}237 stage~2, 254 stage~3, and 210 stage~4 cases, whereas SUSTech comprised 631 stage~0, 24 stage~1, 310 stage~2, 128 stage~3, and 58 stage~4 cases.

All models were applied in a zero-shot manner, and the performance was evaluated using F1-score, Sensitivity, and Specificity, consistent with the metrics described above.

\subsubsection*{Statistical Analysis and Results Reporting}

For each evaluation, we performed bootstrapping with a sample size of 30 randomly selected instances, repeating the process 100 times, and reported the estimated distributions.  
To assess the statistical significance of performance differences between models, we conducted Wilcoxon signed-rank tests.  
These bootstrapping and statistical testing procedures follow established practices~\cite{chen2021multimodal,lenskjold2024artificial}.

\begin{figure*}[t]
\centering
\begin{minipage}[t]{\linewidth}
\vspace{0pt}              
\captionsetup{type=table}
\captionof{table}{Manual evaluation of ophthalmological image descriptions by two ophthalmology residents. Each dimension (conciseness, accuracy, and readability) was rated on a 5-point scale (1 = poor, 5 = excellent) across 40 independently curated test samples. Model identities were anonymized and evaluation order was randomized to minimize bias. ``Ophth.'' is short for "ophthalmology residents".}
\label{tab:manual_evaluation}
\centering
\small
\begin{adjustbox}{width=0.65\linewidth,center}
\begin{tabular}{l|cc|cc|cc}
\toprule
\textbf{Model} & \multicolumn{2}{c|}{\textbf{Conciseness}} & \multicolumn{2}{c|}{\textbf{Accuracy}} & \multicolumn{2}{c}{\textbf{Readability}} \\
\midrule
 & Ophth. 1& Ophth. 2  & Ophth. 1& Ophth. 2 & Ophth. 1& Ophth. 2\\
\midrule
InternVL-2B    & 2.76 & 2.61 & 2.68 & 1.88& 3.37 &3.41 \\
MedGemma-4B    & 2.78 & 2.46 & 2.73 & 1.95& 2.29 &3.05 \\
MedGemma-27B   & 2.07 & 2.05 & 2.95 & 2.37& 1.85 &2.90 \\
\midrule
VOLMO-2B   & \textbf{4.71} & \textbf{4.15}& \textbf{3.15} & \textbf{2.49} & \textbf{4.70}&\textbf{4.12}\\
\bottomrule
\end{tabular}
\end{adjustbox}
\end{minipage}
\end{figure*}

\begin{figure*}[htbp!]
\centering

\begin{subfigure}[t]{\textwidth}
\centering
\begin{minipage}[t]{0.22\linewidth}
\vspace{0pt}
\includegraphics[width=\linewidth,height=\linewidth,keepaspectratio=false]{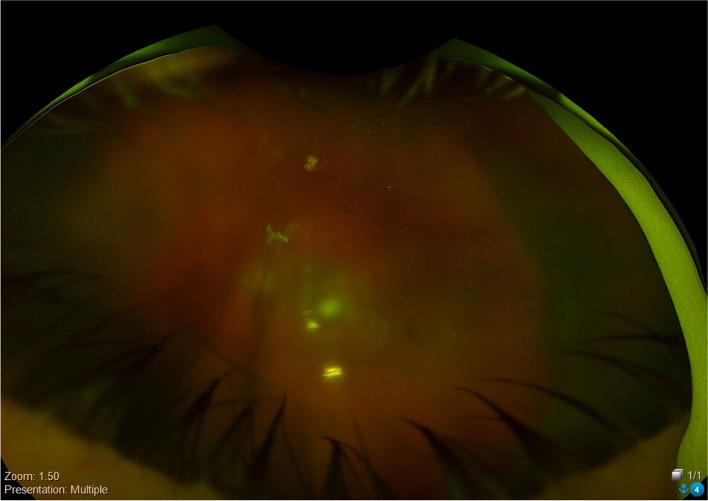}
\end{minipage}%
\hspace{0.05\linewidth}%
\begin{minipage}[t]{0.73\linewidth}
\vspace{0pt}
\footnotesize
\textbf{Gold Standard.} \RaggedRight The Optos image captures the left eye during a recurrence of endophthalmitis.\par
\vspace{0.5em}
\textbf{VOLMO-2B.} \textit{The image illustrates a case of endogenous endophthalitis, characterized by the presence of vitreous exudates and a severe vitreous hemorrhage.}\par
\vspace{0.35em}
\textbf{InternVL-2B.} \textit{This medical image shows a view of the retina, which is the light-sensitive layer at the back of the eye. The retina is responsible for converting light into neural signals that the brain can use to process visual information. The image appears to be taken through a slit lamp biomicroscope, which is used to examine the eye's structures... [truncated from 273 words]}\par
\vspace{0.35em}
\textbf{LLaVA-Med.} \textit{This medical image is a fundus photograph, which is a photograph of the retina and optic nerve of the eye. It is used to visualize and assess the health of the eye, particularly in the context of glaucoma.}\par
\vspace{0.35em}
\textbf{MedGemma-4B.} \textit{The image shows a fundus photograph of the eye. The fundus is the back part of the eye, including the retina, optic disc, and blood vessels. Based on the image, I can see: Retinal Vessels: The retinal blood vessels are visible, with some branching and a general appearance. Macula: The macula, the central part of the retina responsible for sharp, central vision, is present... [truncated from 223 words]}\par
\vspace{0.35em}
\textbf{MedGemma-27B.} \textit{The medical image shows a fundus photograph of the eye. Specifically, it appears to be an image of the posterior segment of the eye, focusing on the retina, optic disc, and macula. Here's a breakdown of what can be observed: Optic Disc: The central, pale circular structure is the optic disc (where the optic nerve enters the eye)... [truncated from 134 words]}\par
\end{minipage}
\caption{}
\label{fig:mri_example}
\end{subfigure}

\vspace{0.8em}

\begin{subfigure}[t]{\textwidth}
\centering
\begin{minipage}[t]{0.22\linewidth}
\vspace{0pt}
\includegraphics[width=\linewidth,height=\linewidth,keepaspectratio=false]{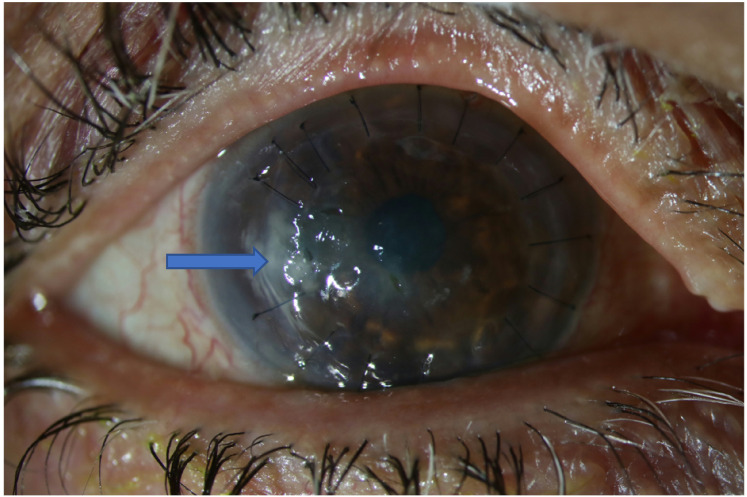}
\end{minipage}%
\hspace{0.05\linewidth}%
\begin{minipage}[t]{0.73\linewidth}
\vspace{0pt}
\footnotesize
\textbf{Gold Standard.} \RaggedRight This case exemplifies the ongoing treatment process with topical voriconazole, highlighting the specific areas of concern with arrows indicating the presence of infiltrates.\par
\vspace{0.5em}
\textbf{VOLMO-2B.} \textit{The photograph illustrates a corneal ulcer, with an arrow indicating the affected area.}\par
\vspace{0.35em}
\textbf{InternVL-2B.} \textit{This medical image shows a close-up view of a human eye. The image is used to examine the eye for signs of conditions such as glaucoma. The blue arrow points to a specific area within the eye, which is likely indicating a region of interest for further examination. In glaucoma, the drainage angle of the eye becomes blocked, leading to increased intraocular pressure... [truncated from 87 words]}\par
\vspace{0.35em}
\textbf{LLaVA-Med.} \textit{The image shows a gross view of a corneal transplant, which is a surgical procedure to replace a damaged or diseased cornea with a healthy one from a donor.}\par
\vspace{0.35em}
\textbf{MedGemma-4B.} \textit{Based on the image, the following is visible: The eye: The primary structure is the eye, with the cornea (the clear front part of the eye) visible. Iridoplasty sutures: There are sutures (stitches) present on the iris (the colored part of the eye). These are likely related to a previous iridoplasty procedure, which is a surgical procedure to reshape the iris... [truncated from 156 words]}\par
\vspace{0.35em}
\textbf{MedGemma-27B.} \textit{The image shows a close-up view of a human eye, specifically focusing on the anterior segment. Key features visible include: Cornea: The clear, dome-shaped outer layer at the front of the eye. Iris: The colored part of the eye, which controls the size of the pupil. Pupil: The black circular opening in the center of the iris, which allows light to enter the eye... [truncated from 159 words]}\par
\end{minipage}
\caption{}
\label{fig:bscan_example}
\end{subfigure}

\caption{Comparison of medical AI model responses on ophthalmological image description tasks.}
\label{fig:medical_ai_comparison}
\end{figure*}

\begin{figure*}[htbp!]\ContinuedFloat
\centering

\begin{subfigure}[t]{\textwidth}
\centering
\begin{minipage}[t]{0.22\linewidth}
\vspace{0pt}
\includegraphics[width=\linewidth,height=\linewidth,keepaspectratio=false]{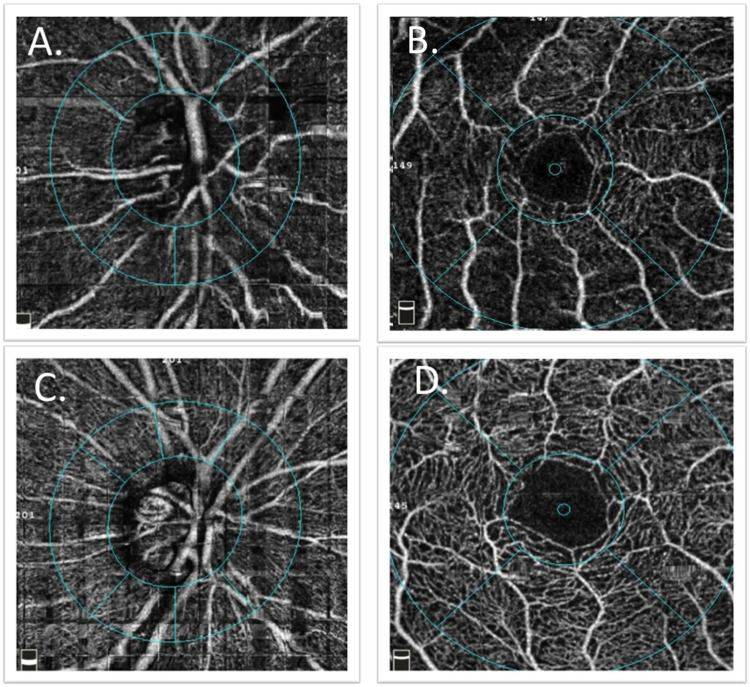}
\end{minipage}%
\hspace{0.05\linewidth}%
\begin{minipage}[t]{0.73\linewidth}
\vspace{0pt}
\footnotesize
\textbf{Gold Standard.} \RaggedRight Figures A and B illustrate the optical coherence tomography angiography (OCTA) images of the optic disc and macular region, respectively, in a patient with Alzheimer's type dementia (ATD). Figures C and D depict the OCTA images of the same regions in a normal participant. The images reveal that vessel density in the optic nerve and macula is reduced in the eye of the ATD patient when compared to the normal eye.\par
\vspace{0.5em}
\textbf{VOLMO-2B.} \textit{The OCTA images illustrate the optic nerve head (ONH) and the peripapillary region in a patient with glaucoma. The first image (A) displays the ONH, while the second image (B) shows the peripapillary region. The third image (C) depicts the ONH, and the fourth image (D) presents the peripapillary region.}\par
\vspace{0.35em}
\textbf{InternVL-2B.} \textit{This medical image shows a cross-sectional view of the human eye, specifically focusing on the retina. The retina is a thin layer of tissue located at the back of the eye, responsible for converting light into neural signals that the brain can interpret as visual images. The image is likely from a fundus examination, which is a procedure used to examine the interior surface of the eye... [truncated from 224 words]}\par
\vspace{0.35em}
\textbf{LLaVA-Med.} \textit{The image is a fundus photograph, which is a photograph of the retina, the back part of the eye. It is used to visualize the structure and health of the retina.}\par
\vspace{0.35em}
\textbf{MedGemma-4B.} \textit{The image shows a fundus photograph of the retina. The central fovea, which is responsible for sharp central vision, is clearly visible in the center of each image. The retinal vessels, including the retinal arteries and veins, are also visible, with their branching patterns. The image appears to be a fundus photograph taken with optical coherence tomography (OCT) or similar imaging technology.}\par
\vspace{0.35em}
\textbf{MedGemma-27B.} \textit{The image shows a series of fundus photographs of the eye, specifically focusing on the macula and optic nerve head. These are typically obtained using a fundus camera. Here's a breakdown of what's visible: Optic Nerve Head (ONH): The circular structure in the center of each image, where the optic nerve enters the eye... [truncated from 147 words]}\par
\end{minipage}
\caption{}
\label{fig:fa_example}
\end{subfigure}

\caption{Comparison of medical AI model responses on ophthalmological image description tasks (continued).}
\end{figure*}

\begin{figure}[htbp!]
    \centering
    
    \begin{tabular}{@{}p{0.3\textwidth}@{\hspace{0.03\textwidth}}p{0.3\textwidth}@{\hspace{0.03\textwidth}}p{0.3\textwidth}@{}}
        \centering\textbf{Assessment \& Plan} & 
        \centering\textbf{Treatments} & 
        \centering\textbf{Follow-up Care} \\
    \end{tabular}
    
    \begin{subfigure}[b]{0.3\textwidth}
        \centering
        \includegraphics[width=\textwidth]{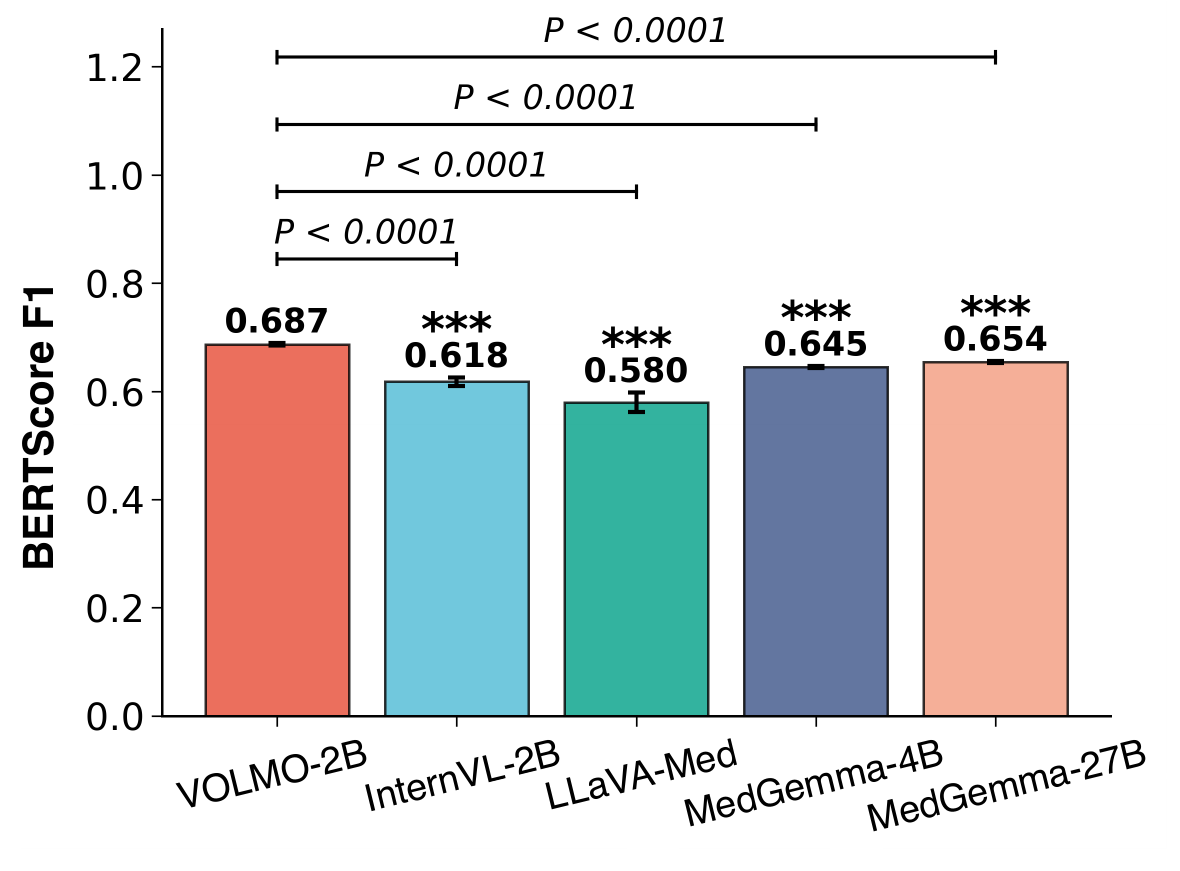}
        \label{fig:dd_f1}
    \end{subfigure}%
    \hspace{0.03\textwidth}%
    \begin{subfigure}[b]{0.3\textwidth}
        \centering
        \includegraphics[width=\textwidth]{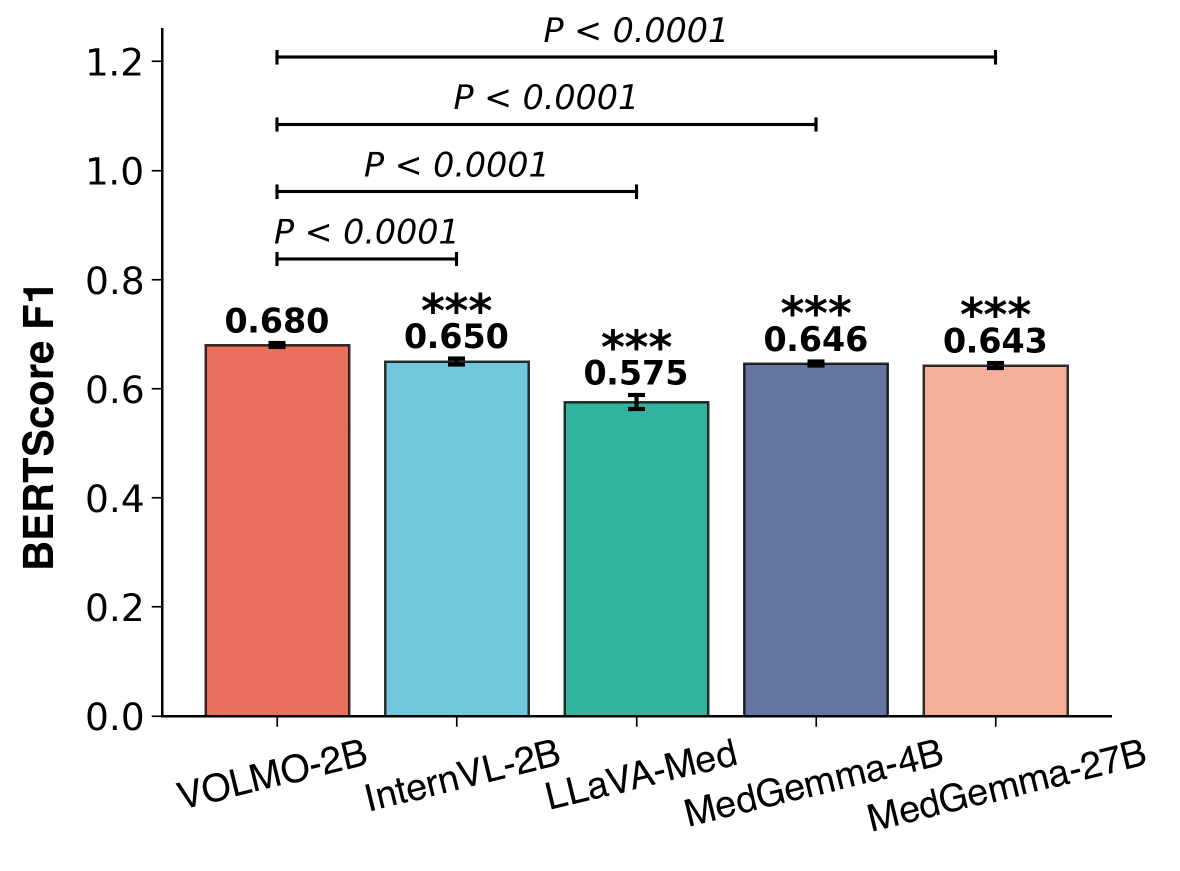}
        \label{fig:ap_bs_f1}
    \end{subfigure}%
    \hspace{0.03\textwidth}%
    \begin{subfigure}[b]{0.3\textwidth}
        \centering
        \includegraphics[width=\textwidth]{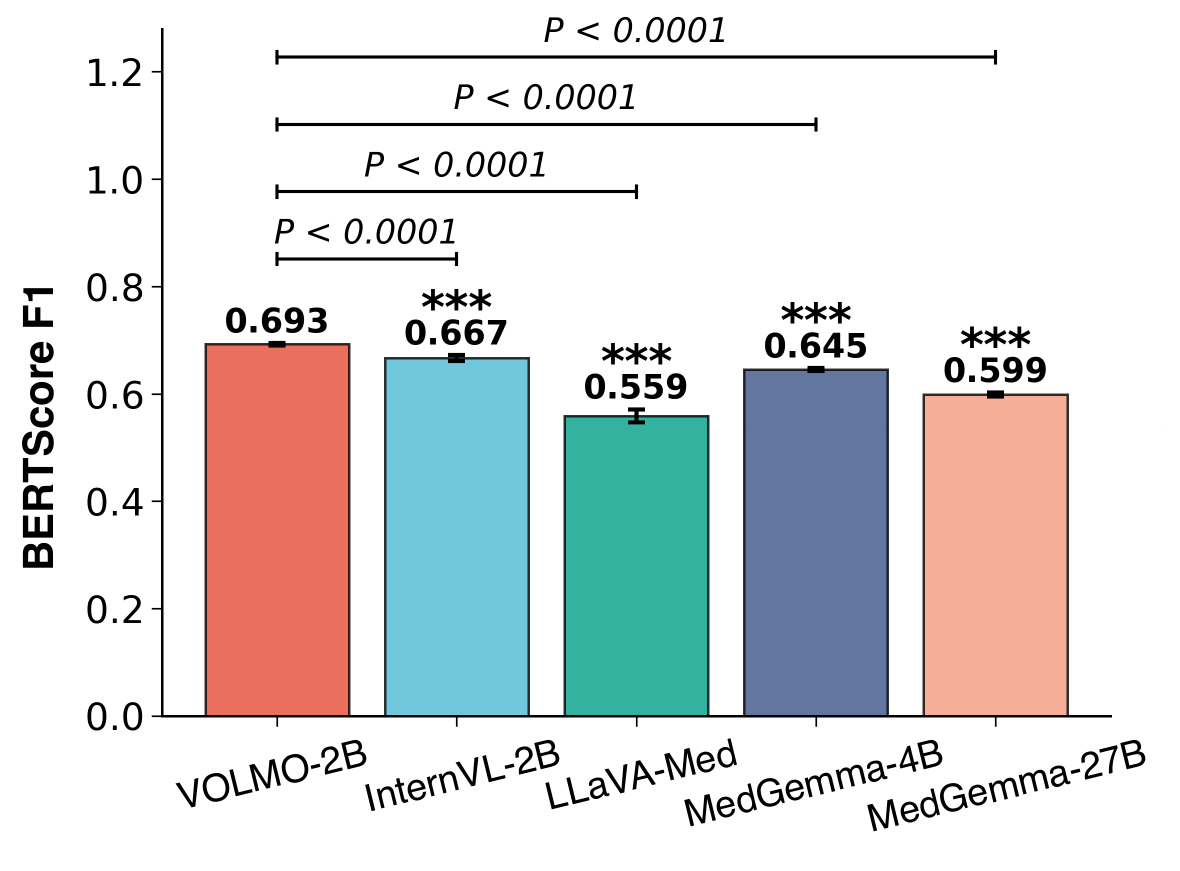}
        \label{fig:tf_bs_f1}
    \end{subfigure}
    
    \vspace{-0.8em}
    
    \begin{subfigure}[b]{0.3\textwidth}
        \centering
        \includegraphics[width=\textwidth]{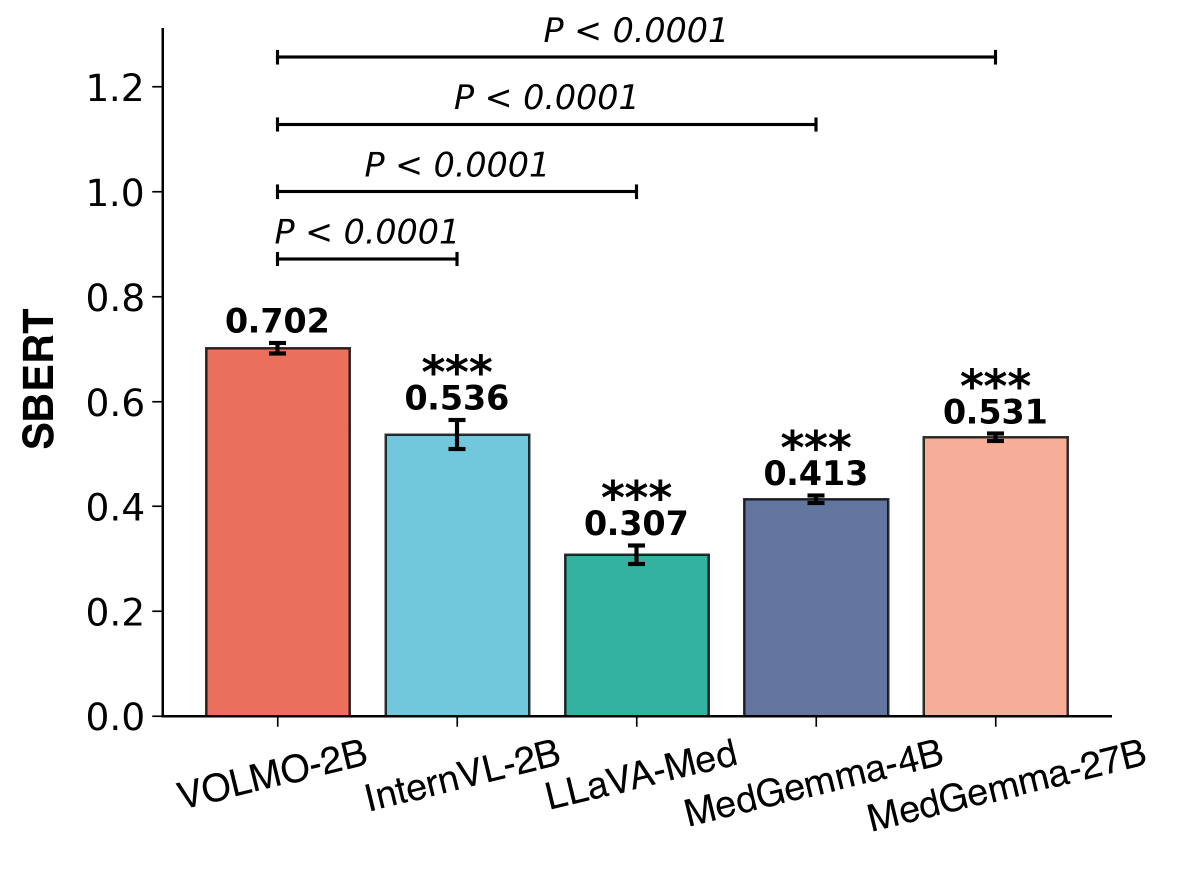}
        \label{fig:dd_bs_f1}
    \end{subfigure}%
    \hspace{0.03\textwidth}%
    \begin{subfigure}[b]{0.3\textwidth}
        \centering
        \includegraphics[width=\textwidth]{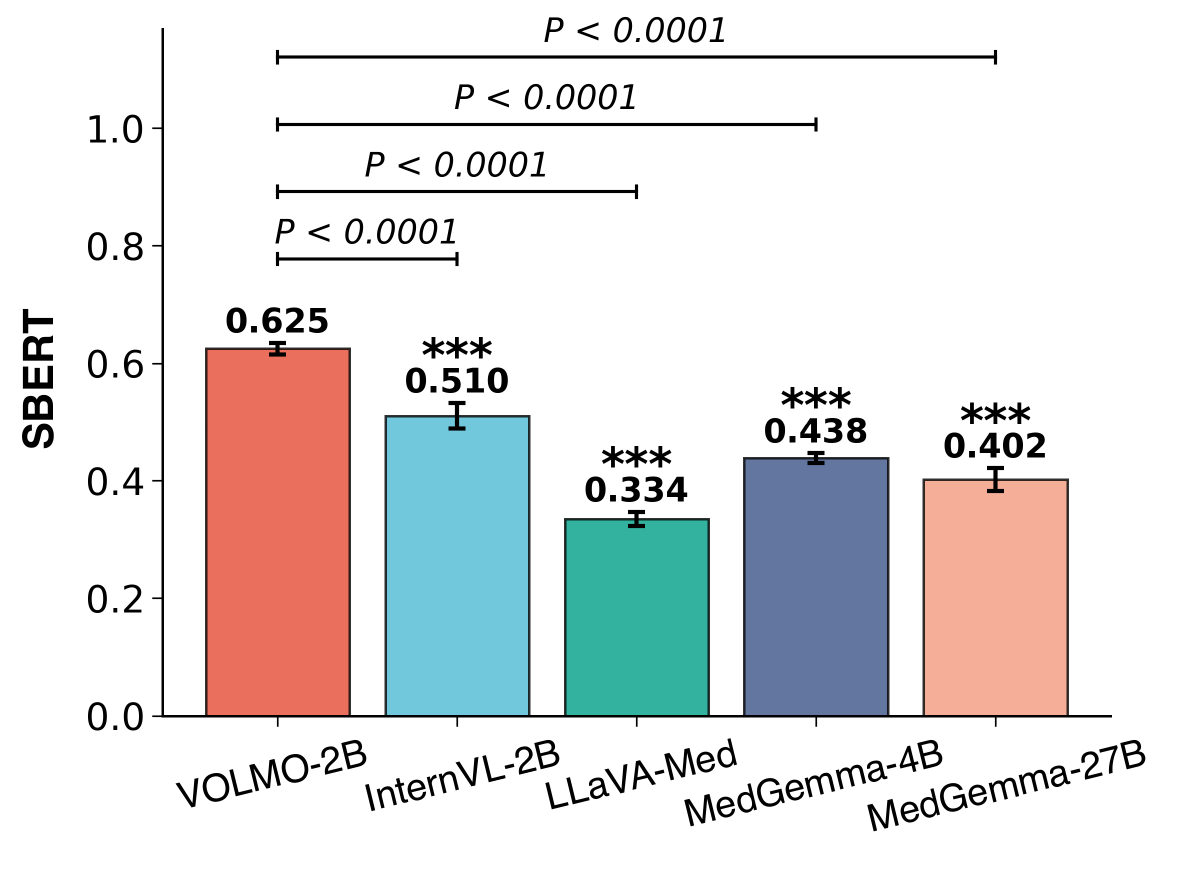}
        \label{fig:ap_sb}
    \end{subfigure}%
    \hspace{0.03\textwidth}%
    \begin{subfigure}[b]{0.3\textwidth}
        \centering
        \includegraphics[width=\textwidth]{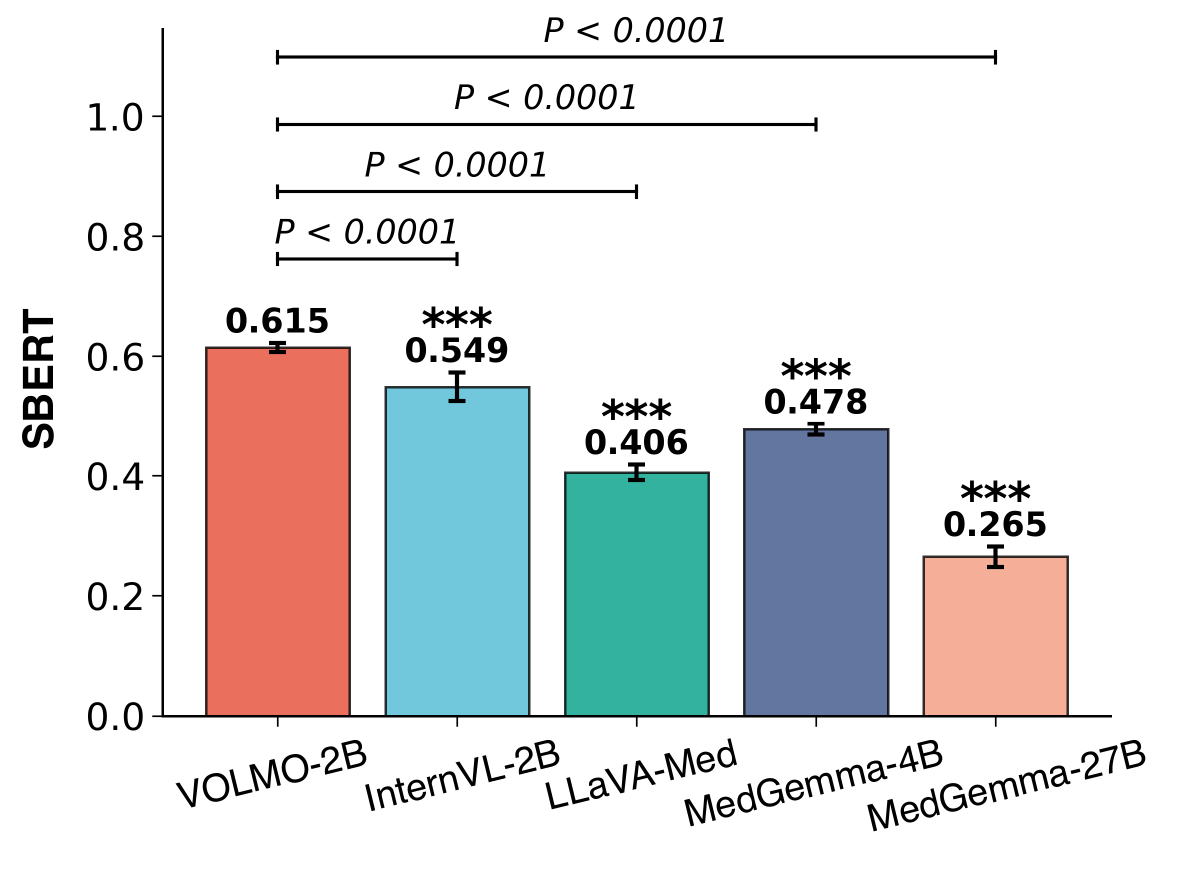}
        \label{fig:tf_sb}
    \end{subfigure}
    
    \caption{
    Performance comparison across different clinical stages. Each column shows Assessment \& Plan, Treatments and Follow-up Care  performance with BERTScore-F1 and SBERT scores. Each subplot compares VOLMO-2B against InternVL-2B, LLaVA-Med, MedGemma-4B and MedGemma-27B. Error bars represent standard deviations, and statistical significance markers (***) indicate p-values compared to VOLMO-2B baseline.
    }
    \label{fig:medical_model_comparison}
\end{figure}

\begin{figure}[htbp]
\centering
\begin{minipage}[t]{0.48\linewidth}
    \centering
    \includegraphics[width=\linewidth]{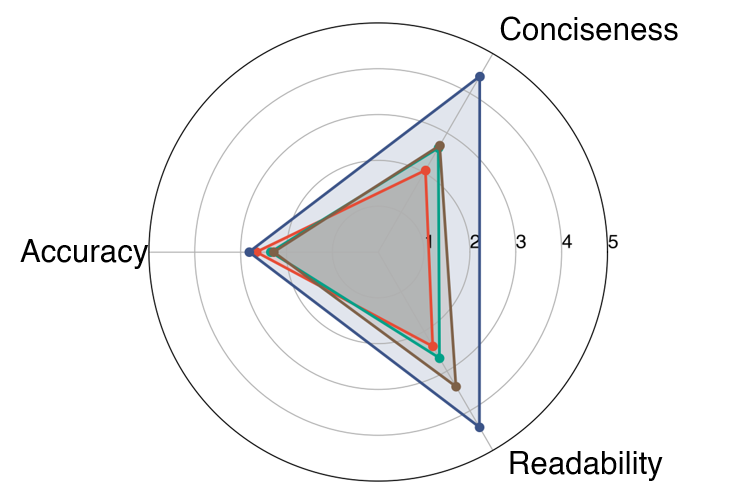}
\end{minipage}
\begin{minipage}[t]{0.14\linewidth}
    \centering
    \includegraphics[width=\linewidth]{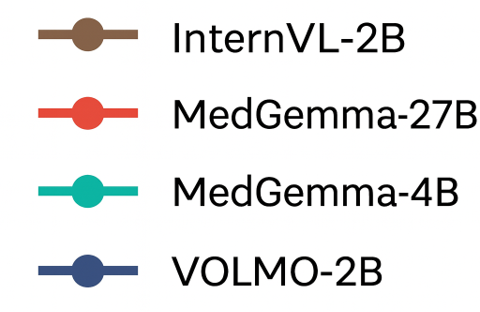}
\end{minipage}
    \caption{Average manual evaluation scores across three quality dimensions. Two ophthalmology residents independently rated generated ophthalmological image descriptions on a 5-point scale (1 = poor, 5 = excellent) across 40 test samples for conciseness, accuracy, and readability. Model identities were anonymized and evaluation order randomized to minimize bias. }
    \label{fig:manual_eval_avg}
\end{figure}

\begin{figure*}[p!]
\centering

\begin{subfigure}[t]{\textwidth}
\centering
\begin{minipage}[t]{0.22\linewidth}
\vspace{0pt}
\includegraphics[width=\linewidth,height=\linewidth,keepaspectratio=false]{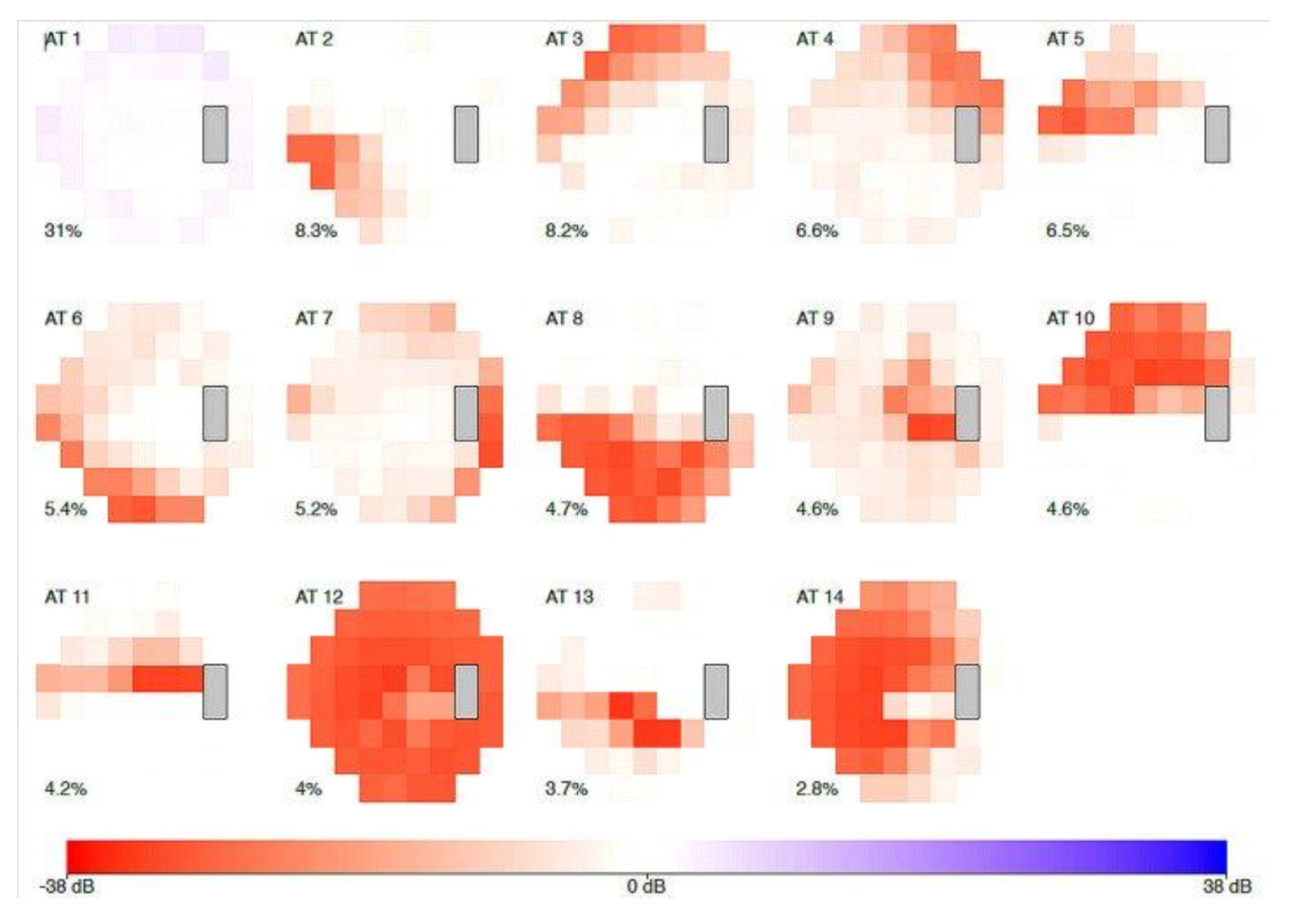}
\end{minipage}%
\hspace{0.05\linewidth}%
\begin{minipage}[t]{0.73\linewidth}
\vspace{0pt}
\footnotesize
\textbf{Gold Standard.} \RaggedRight Fourteen archetypal visual field loss patterns were derived from the visual fields of 1957 primary open-angle glaucoma cases, encompassing 2581 affected eyes. Each archetype is denoted by an integer at the top left, indicating its number, and a percentage at the bottom left, representing the average decomposition weight for that pattern. The algorithm discerned 14 archetypes in total: four indicative of advanced loss patterns, nine of early loss, and one signifying no visual field loss. Although African-American patients constituted 1.3 percent of the study population, they exhibited nearly twice the risk of early visual field loss archetypes and six times the risk for advanced field loss archetypes in comparison to white patients. \par
\vspace{0.5em}
\textbf{VOLMO-2B.} \textit{The visual representation illustrates the distribution of visual field mean deviations across ten distinct regions of the visual field, as determined by microperimetry.}\par
\vspace{0.35em}
\textbf{InternVL-2B.} \textit{This medical image appears to be a spectrogram of an electrocardiogram (ECG or EKG), which is a graphical representation of the electrical activity of the heart over time. The ECG is a crucial tool in diagnosing various heart conditions. Here's a detailed breakdown of what is shown in the image: 1. Spectrogram: The image is a spectrogram, which is a visual representation of the spectrum of frequencies present in a signal. In this case, it shows the heart's electrical activity over time. 2.Grid Layout: The grid is divided into segments, each representing a specific time interval during which the heart's electrical activity is recorded. The segments are labeled from AT 1 to AT 14, which correspond to different time intervals on the ECG... [truncated from 441 words]}\par
\vspace{0.35em}
\textbf{MedGemma-4B.} \textit{Based on the image, it appears to be a sonogram of the brain. Here's a breakdown of what I can infer: Multiple views: The image shows multiple slices (AT 1, AT 2, AT 3, etc.) of the brain, likely obtained from different angles. Color-coded representation: The color intensity (red to blue) likely represents the amplitude of the ultrasound signal. Brighter colors indicate stronger signals, while darker colors indicate weaker signals. Brain structures: The image shows the typical brain structures, including the cerebral hemispheres (the large, rounded areas), the ventricles (the fluid-filled spaces within the brain), and the skull. Measurements: The percentage values (e.g., 8.3\%, 6.6\%) likely represent measurements of the ventricles or other structures... [truncated from 204 words]}\par
\vspace{0.35em}
\textbf{MedGemma-27B.} \textit{The image shows a series of 16 axial CT scans of the chest, labeled AT 1 through AT 16. Each slice is presented in a "lung window" setting, which emphasizes the lung parenchyma and airways. The slices are arranged sequentially from the apex (AT 1) to the base (AT 16) of the chest. Key features visible in the image include:1.  Anatomical Structures: The ribs, vertebrae, mediastinum (containing the heart and great vessels), and the lung fields are clearly visible. 2.  Lung Parenchyma: The density and texture of the lung tissue can be assessed. 3.  Airways: The bronchi and bronchioles are vible as air-filled structures within the lungs. 4.  Pleural Spaces: The space between the lung and the chest wall is visible.5.  Contrast:... [truncated from 324 words]}\par
\end{minipage}
\caption{VOLMO-2B outperforms on conciseness, accuracy, and readability. }
\label{fig:better_example1}
\end{subfigure}

\caption{Representative examples of ophthalmological image descriptions generated by different models, manually evaluated by ophthalmologists. }
\label{fig:manual_eval_example}
\end{figure*}

\begin{figure*}[p!]\ContinuedFloat
\centering

\begin{subfigure}[t]{\textwidth}
\centering
\begin{minipage}[t]{0.22\linewidth}
\vspace{0pt}
\includegraphics[width=\linewidth,height=\linewidth,keepaspectratio=false]{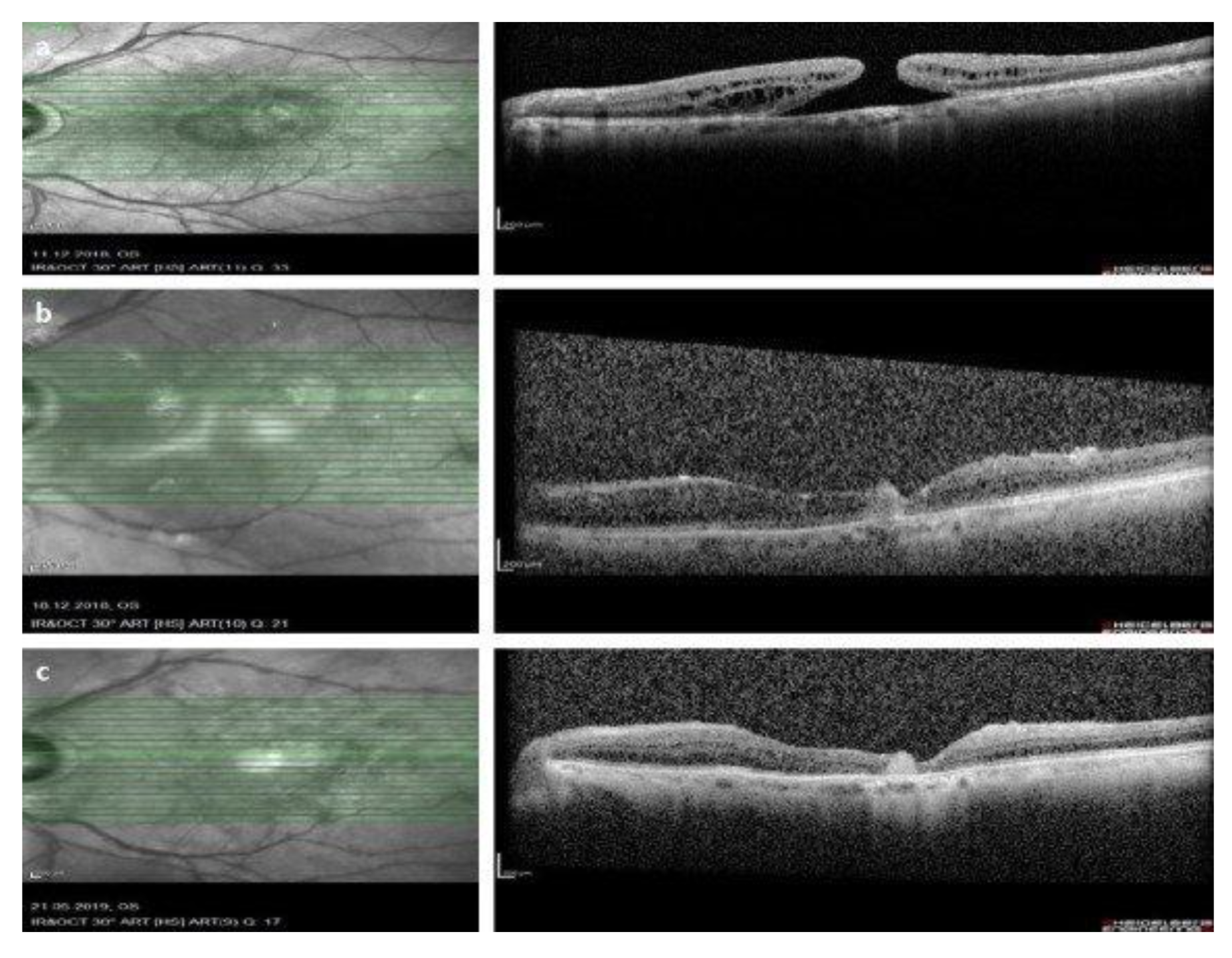}
\end{minipage}%
\hspace{0.05\linewidth}%
\begin{minipage}[t]{0.73\linewidth}
\vspace{0pt}
\footnotesize
\textbf{Gold Standard.} \RaggedRight The preoperative optical coherence tomography (OCT) image revealed a full-thickness macular hole. Twenty days post-surgery, an OCT scan indicated the closure of the hole with the amniotic membrane (AM) properly positioned. At the six-month postoperative mark, the OCT scan demonstrated that the hole remained closed, and the neurosensory retina had overextended the amniotic membrane.\par
\vspace{0.5em}
\textbf{VOLMO-2B.} \textit{The optical coherence tomography (OCT) images illustrate the progression of a patient with a macular hole (MH) and a lamellar macular hole (LMH) at different stages: preoperative (a), 1 month postoperative (b), 3 months postoperative (c), and 6 months postoperative (d).}\par
\vspace{0.35em}
\textbf{InternVL-2B.} \textit{This medical image shows a cross-sectional view of the human retina, which is a critical component of the eye. The retina is responsible for converting light into neural signals that the brain can interpret as visual images. The image is divided into three sections labeled a, b, and c, each depicting different aspects or stages of retinal imaging. 1.Section a: This part of the image appears to show a detailed view of the retinal layers, possibly highlighting the ganglion cell layer, inner plexiform layer, inner nuclear layer, and outer plexiform layer. The green lines could represent some form of imaging technique or overlay used to enhance visibility of specific structures or abnormalities.2.Section b ... [truncated from 436 words]}\par
\vspace{0.35em}
\textbf{MedGemma-4B.} \textit{Based on the provided image, it appears to be a fundus photograph of the retina. The image shows the optic disc, retinal vessels (arteries and veins), and the macula. The green lines are likely a grid overlay used for alignment and measurement. Here's a breakdown of what I can see: Optic Disc: The central point of the image, where the optic nerve exits the eye. Retinal Vessels: The arteries and veins are visible as branching structures. Macula: The central part of the retina responsible for sharp, central vision. It appears as a slightly darker area in the center of the image. Retinal Details: The image shows the fine details of the retinal layers. Without more information about the patient's history and clinical context, it's difficult to provide a definitive diagnosis. However, based on the image alone, it appears to be a normal fundus photograph. If you have more information about the patient, such as their age, medical history, and any symptoms they are experiencing, I can provide a more specific assessment.}\par
\vspace{0.35em}
\textbf{MedGemma-27B.} \textit{The provided image is an Optical Coherence Tomography (OCT) scan of the eye, specifically focusing on the macula and posterior segment. Here's a breakdown of what's visible: OCT Scan: The image is a cross-sectional view of the retina and underlying structures, generated using OCT technology. This allows visualization of the different layers of the retina, choroid, and vitreous. Macula: The central part of the retina, responsible for sharp, detailed central vision, is clearly depicted. The fovea, the very center of the macula, is also visible. Retinal Layers: The distinct layers of the retina (e.g., nerve fiber layer, ganglion cell layer, inner plexiform layer, inner nuclear layer, outer plexiform layer, outer nuclear layer, photoreceptor layer) are identifiable. Choroid: The vascular layer beneath the retina is visible, appearing as a highly reflective structure. Vitreous: The gel-like substance filling the eye is seen above the retina. Scan Parameters: The image includes technical details like scan date, time, and parameters (e.g., scan length, AOU FOV)... [truncated from 299 words]}\par
\end{minipage}
\caption{VOLMO-2B outperforms on conciseness, accuracy, and readability. }
\label{fig:better_example2}
\end{subfigure}

\caption{Representative examples of ophthalmological image descriptions generated by different models, manually evaluated by ophthalmologists (continued).}
\end{figure*}

\begin{figure*}[p!]\ContinuedFloat
\centering

\begin{subfigure}[t]{\textwidth}
\centering
\begin{minipage}[t]{0.22\linewidth}
\vspace{0pt}
\includegraphics[width=\linewidth,height=\linewidth,keepaspectratio=false]{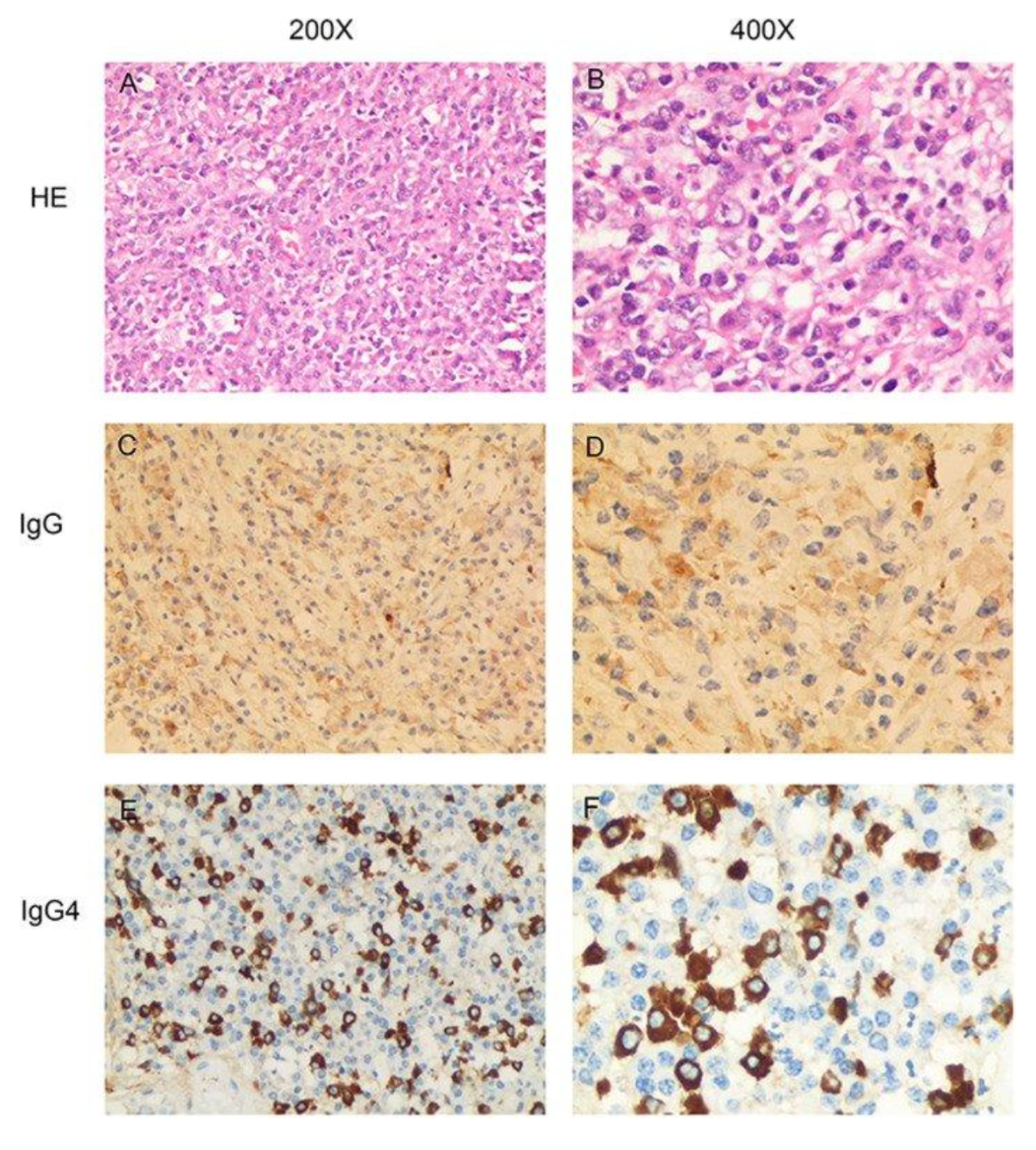}
\end{minipage}%
\hspace{0.05\linewidth}%
\begin{minipage}[t]{0.73\linewidth}
\vspace{0pt}
\footnotesize
\textbf{Gold Standard.} \RaggedRight The histological and immunohistochemical analysis of the left eye revealed notable findings. Hematoxylin-eosin staining indicated the presence of infiltrating plasmacytes, lymphocytes, neutrophils, and histiocytes, alongside the proliferation of myofibroblasts. Immunohistochemical staining demonstrated that the plasmacytes were positive for IgG and IgG4, with an IgG4+/IgG+ plasma cell ratio of approximately 40\%.\par
\vspace{0.5em}
\textbf{VOLMO-2B.} \textit{The histopathological analysis of the conjunctival biopsy reveals a diffuse infiltration of lymphocytes, plasma cells, and histiocytes within the conjunctival epithelium. The neoplastic cells exhibit a high nuclear-to-cytoplasmic ratio, with some cells displaying a high nuclear-to-cytoplasmic ratio, while others display a low nuclear-to-cytoplasmic ratio. The neoplastic cells are characterized by a high nuclear-to-cytoplasmic ratio, and the plasma cells are identified by a high nuclear-to-cytoplasmic ratio. The neoplastic cells are also identified by a high nuclear-to-cytoplasmic ratio, and the plasma cells are identified by a high nuclear-to-cytoplasmic ratio.}\par
\vspace{0.35em}
\textbf{InternVL-2B.} \textit{This medical image shows a histological slide of a tissue sample stained with Hematoxylin and Eosin (H\&E). Hematoxylin stains cell nuclei blue, while eosin stains the cytoplasm and extracellular matrix pink. The slide is viewed at different magnifications: 200X, 400X, and 1000X. In the 200X magnification (A), the tissue appears to have a dense cellular arrangement with numerous cells, some of which have prominent nuclei. The eosinophilic (pink) background suggests the presence of cytoplasmic components and extracellular matrix. In the 400X magnification (B), the cellular density is slightly reduced, and the nuclei appear more distinct... [truncated from 276 words]}\par
\vspace{0.35em}
\textbf{MedGemma-4B.} \textit{Based on the image, it appears to be a histopathology slide showing a tissue sample, likely from a biopsy. The image shows two different magnification levels (200x and 400x) and two different staining methods (Hematoxylin and Eosin (HE) and Immunoglobulin G (IgG)). Here's a breakdown of what's visible: HE Stain (Hematoxylin and Eosin): This stain is a general stain used to visualize the tissue architecture. It highlights nuclei (blue) and cytoplasm (pink). In the HE stain, there appears to be cellular changes, possibly indicating inflammation or other pathological processes. IgG Stain: This stain is used to detect the presence of IgG antibodies. In the IgG stain, there are numerous IgG-positive cells scattered throughout the tissue... [truncated from 347 words]}\par
\vspace{0.35em}
\textbf{MedGemma-27B.} \textit{The provided image shows a set of histological slides, likely from a bone marrow biopsy. Let's break down the different sections:A \& B: These sections are stained with Hematoxylin and Eosin (H\&E). They show a hypercellular bone marrow with a high myeloid-to-erythroid (M:E) ratio. There is a marked increase in granulocytic precursors (myeloblasts, promyelocytes, myelocytes, metamyelocytes, bands, and segmented neutrophils) compared to erythroid precursors. The megakaryocytes appear relatively normal in number and morphology. The overall architecture of the marrow is somewhat disrupted due to the cellularity. C \& D: These sections are stained with Immunoglobulin (Ig) stains. They show a significant increase in plasma cells, which are positive for immunoglobulin. This indicates a clonal proliferation of plasma cells... [truncated from 366 words]}\par
\end{minipage}
\caption{VOLMO-2B matches accuracy while excelling in conciseness and readability. }
\label{fig:tie_example1}
\end{subfigure}

\caption{Representative examples of ophthalmological image descriptions generated by different models, manually evaluated by ophthalmologists (continued).}
\end{figure*}

\begin{figure*}[p!]\ContinuedFloat
\centering

\begin{subfigure}[t]{\textwidth}
\centering
\begin{minipage}[t]{0.22\linewidth}
\vspace{0pt}
\includegraphics[width=\linewidth,height=\linewidth,keepaspectratio=false]{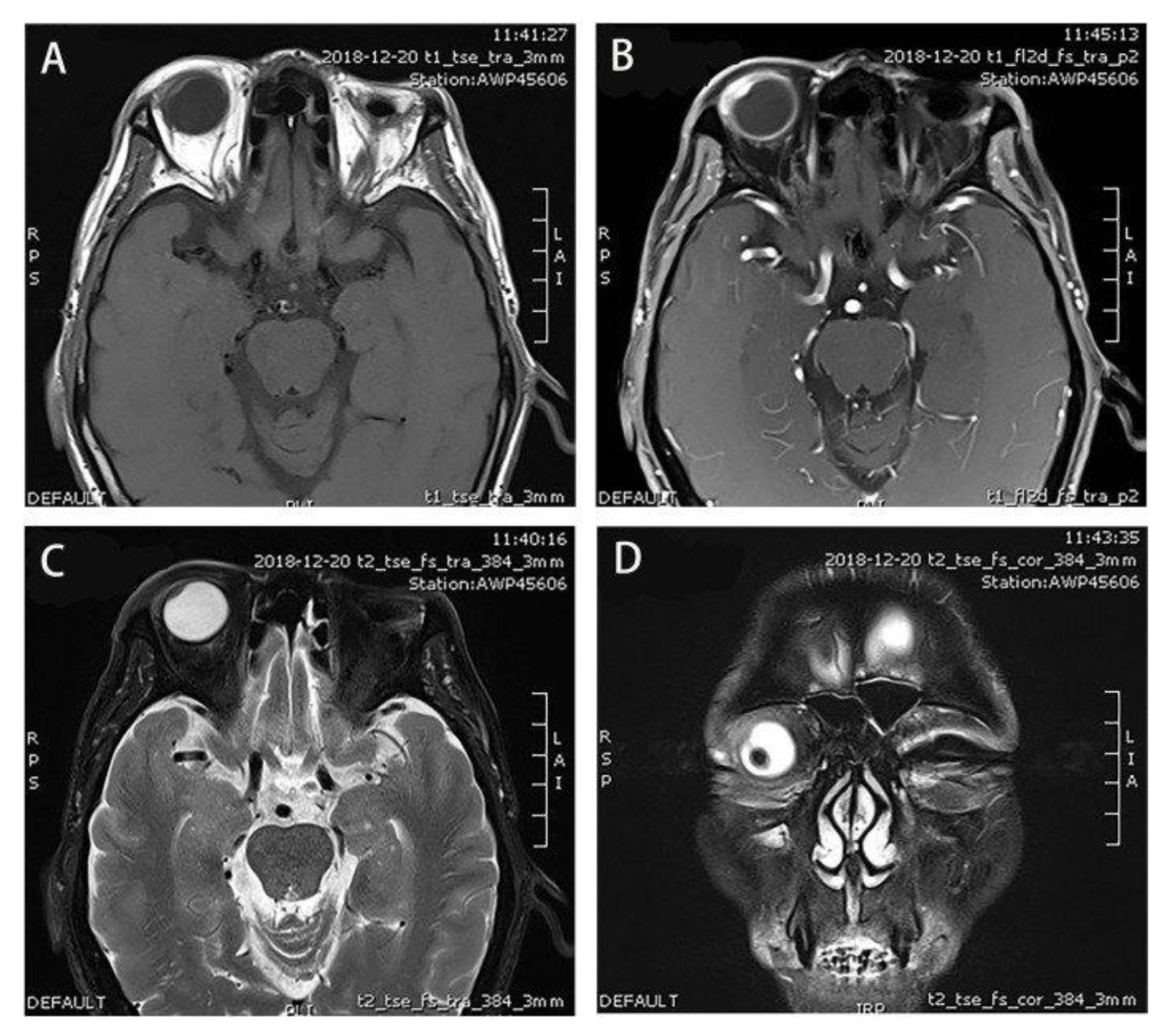}
\end{minipage}%
\hspace{0.05\linewidth}%
\begin{minipage}[t]{0.73\linewidth}
\vspace{0pt}
\footnotesize
\textbf{Gold Standard.} \RaggedRight MRI scans of the right orbit from the patient’s second visit reveal a lesion in the ciliary body region of the right eye. On the T1-weighted image (T1WI), the lesion exhibits an isointense to slightly hyperintense signal. Following gadolinium contrast enhancement, the lesion demonstrates visible enhancement in the T1-f12d fat suppression (fs) image. On T2-weighted images (T2WI), the lesion appears hypointense, as observed in both T2-tse-fs-tra and T2-tse-fs-cor scans.\par
\vspace{0.5em}
\textbf{VOLMO-2B.} \textit{The MRI images illustrate the condition of a patient with a right orbital floor fracture. Image A reveals a T2-weighted coronal section, indicating the presence of a right orbital floor fracture. Image B displays a T1-weighted axial section, highlighting the extent of the fracture. Image C presents a T2-weighted axial section, while Image D shows a T1-weighted coronal section.}\par
\vspace{0.35em}
\textbf{InternVL-2B.} \textit{This medical image shows a series of brain scans, specifically MRI scans, of a patient's head. MRI stands for Magnetic Resonance Imaging, which is a non-invasive imaging technology that produces three-dimensional detailed anatomical images. These images are often used to view the brain and spinal cord, and can be used to detect abnormalities such as tumors, stroke, and other conditions. The scans in the image are labeled A, B, C, and D, each representing a different time point or view of the brain. The scans are taken from different angles and at different times, allowing for a comprehensive analysis of the brain's structure and any potential abnormalities.}\par
\vspace{0.35em}
\textbf{MedGemma-4B.} \textit{Based on the provided images, it appears to be a series of axial T1-weighted and T2-weighted MRI scans of the brain, specifically focusing on the orbits and surrounding structures.Here's a breakdown of what's visible:T1-weighted images (A \& C): These images highlight the anatomical structures based on their proton density. You can see the brain parenchyma, the orbits, the optic nerves, and the surrounding soft tissues.T2-weighted images (B \& D): These images highlight fluid and edema. You can see the optic nerve sheath, the orbital fat, and the surrounding soft tissues. Key Observations: Optic Nerve Sheath: The optic nerve sheaths are visible in both T1 and T2 images.Orbital Structures:... [truncated from 285 words]}\par
\vspace{0.35em}
\textbf{MedGemma-27B.} \textit{The medical image shows a series of axial T2-weighted MRI scans of the brain, specifically focusing on the orbits and optic nerves. The images are labeled (A) through (D), with corresponding sagittal reconstructions (B) and (D). Here's a breakdown of what's visible:  Anatomical Structures: The images clearly depict the orbits (eye sockets), optic nerves (running from the back of the eyes towards the brain), optic chiasm (where the optic nerves cross), and surrounding brain structures like the temporal lobes and pituitary gland... [truncated from 258 words]}\par
\end{minipage}
\caption{VOLMO-2B matches accuracy while excelling in conciseness and readability. }
\label{fig:tie_example2}
\end{subfigure}

\caption{Representative examples of ophthalmological image descriptions generated by different models, manually evaluated by ophthalmologists (continued).}
\end{figure*}

\begin{table*}[t]
\centering
\caption{Consolidated results: (A) UK Biobank classification (AMD/DR), (B) Sydney Innovation (cross-domain) grading, and (C) SUSTech (cross-domain) grading. Values are mean~$\pm$~std; $p$-values from Wilcoxon signed-rank tests when reported. All numeric columns are right-aligned; model order is consistent across parts.}
\label{tab:combined_three_parts}
{\scriptsize
\setlength{\tabcolsep}{2.6pt}
\renewcommand{\arraystretch}{0.95}

\textbf{(A) UK Biobank — AMD and DR} \\[2pt]
\begin{tabular}{l|rrr|rrr}
\toprule
\textbf{Method}
& \multicolumn{3}{c|}{\textbf{AMD}} 
& \multicolumn{3}{c}{\textbf{DR}} \\
\cmidrule(lr){2-4}\cmidrule(lr){5-7}
& \textbf{F1} & \textbf{Sens} & \textbf{Spec} 
& \textbf{F1} & \textbf{Sens} & \textbf{Spec} \\
\midrule
\textbf{InternVL-2B}
& $41.47 \pm 10.10$ & $53.07 \pm 8.36$ & $52.06 \pm 7.59$
& $24.69 \pm 8.21$ & $41.37 \pm 8.28$ & $58.63 \pm 8.28$ \\
& $(p < 0.0001)$ & $(p < 0.0001)$ & $(p < 0.0001)$
& $(p < 0.0001)$ & $(p < 0.0001)$ & $(p < 0.0001)$ \\
\addlinespace[2pt]
\textbf{LLaVA-Med}
& $34.41 \pm 8.69$ & $50.63 \pm 7.90$ & $49.37 \pm 7.90$
& $43.77 \pm 9.85$ & $58.71 \pm 8.19$ & $41.29 \pm 8.19$ \\
& $(p < 0.0001)$ & $(p < 0.0001)$ & $(p < 0.0001)$
& $(p = 0.0002)$ & $(p = 0.0032)$ & $(p < 0.0001)$ \\
\addlinespace[2pt]
\textbf{MedGemma-4B}
& $32.94 \pm 8.86$ & $49.30 \pm 7.98$ & $50.70 \pm 7.98$
& $37.17 \pm 8.23$ & $47.90 \pm 7.33$ & $62.95 \pm 8.73$ \\
& $(p < 0.0001)$ & $(p < 0.0001)$ & $(p < 0.0001)$
& $(p < 0.0001)$ & $(p < 0.0001)$ & $(p = 0.0001)$ \\
\addlinespace[2pt]
\textbf{MedGemma-27B}
& $33.77 \pm 9.05$ & $49.50 \pm 8.02$ & $50.74 \pm 7.89$
& $40.16 \pm 9.37$ & $50.31 \pm 7.88$ & $62.27 \pm 8.97$ \\
& $(p < 0.0001)$ & $(p < 0.0001)$ & $(p < 0.0001)$
& $(p < 0.0001)$ & $(p < 0.0001)$ & $(p < 0.0001)$ \\
\addlinespace[2pt]
\textbf{RETFound}
& $46.39 \pm 9.95$ & $55.33 \pm 7.93$ & $56.33 \pm 7.79$
& $\mathbf{53.36 \pm 9.77}$ & $56.83 \pm 8.79$ & $67.30 \pm 7.99$ \\
& $(p < 0.0001)$ & $(p < 0.0001)$ & $(p < 0.0001)$
& $(p = 0.0004)$ & $(p = 0.0025)$ & $(p = 0.0130)$ \\
\midrule
\textbf{VOLMO-2B}
& $\mathbf{64.58 \pm 9.30}$ & $65.93 \pm 8.69$ & $66.75 \pm 8.50$
& $49.86 \pm 9.06$ & $54.20 \pm 7.82$ & $65.12 \pm 8.73$ \\
\bottomrule
\end{tabular}

\vspace{1.2ex}

\textbf{(B) Sydney Innovation — Cross-Domain Grading} \\[2pt]
\resizebox{\linewidth}{!}{%
\begin{tabular}{l|rrr|rrr|rrr|rrr|rrr|rrr}
\toprule
\textbf{Method}
& \multicolumn{3}{c|}{\textbf{Overall}}
& \multicolumn{3}{c|}{\textbf{Stage 0}}
& \multicolumn{3}{c|}{\textbf{Stage 1}}
& \multicolumn{3}{c|}{\textbf{Stage 2}}
& \multicolumn{3}{c|}{\textbf{Stage 3}}
& \multicolumn{3}{c}{\textbf{Stage 4}} \\
\cmidrule(lr){2-4}\cmidrule(lr){5-7}\cmidrule(lr){8-10}\cmidrule(lr){11-13}\cmidrule(lr){14-16}\cmidrule(lr){17-19}
& \textbf{F1} & \textbf{Sens} & \textbf{Spec}
& \textbf{F1} & \textbf{Sens} & \textbf{Spec}
& \textbf{F1} & \textbf{Sens} & \textbf{Spec}
& \textbf{F1} & \textbf{Sens} & \textbf{Spec}
& \textbf{F1} & \textbf{Sens} & \textbf{Spec}
& \textbf{F1} & \textbf{Sens} & \textbf{Spec} \\
\midrule
\textbf{InternVL-2B} &  1.23 & 20.00 & 80.00 &  0.00 &  0.00 & 100.00 &  0.00 &  0.00 & 100.00 &  0.00 &  0.00 & 100.00 &  6.17 & 100.00 &  0.00 &  0.00 &  0.00 & 100.00 \\
\textbf{LLaVA-Med}   &  0.00 &  0.00 & 100.00 &  0.00 &  0.00 & 100.00 &  0.00 &  0.00 & 100.00 &  0.00 &  0.00 & 100.00 &  0.00 &  0.00 & 100.00 &  0.00 &  0.00 & 100.00 \\
\textbf{MedGemma-4B} & 44.50 & 49.24 & 91.78 & 92.89 & 94.96 & 75.45 & 23.62 & 19.92 & 96.55 & 43.76 & 32.90 & 96.80 & 18.30 & 27.95 & 94.16 & 43.92 & 70.48 & 95.93 \\
\textbf{MedGemma-27B}& 39.96 & 46.39 & 90.49 & 92.80 & 98.00 & 65.89 & 13.91 &  9.11 & 98.46 & 35.47 & 23.69 & 98.19 & 17.00 & 21.65 & 95.63 & 40.63 & 79.52 & 94.27 \\
\textbf{RETFound}    & 40.18 & 49.02 & 87.21 & 85.65 & 88.22 & 54.08 & 10.02 &  8.16 & 96.14 & 21.25 & 14.15 & 96.52 & 34.28 & 73.62 & 91.59 & 49.71 & 60.95 & 97.72 \\
\midrule
\textbf{VOLMO-2B}    & \textbf{51.45} & 64.01 & 89.93 & 78.06 & 67.85 & 84.56 & 22.43 & 51.23 & 78.39 & 47.32 & 41.79 & 93.61 & 50.99 & 81.10 & 95.50 & 58.47 & 78.10 & 97.59 \\
\bottomrule
\end{tabular}
}

\vspace{1.2ex}

\textbf{(C) SUSTech — Cross-Domain Grading} \\[2pt]
\resizebox{\linewidth}{!}{%
\begin{tabular}{l|rrr|rrr|rrr|rrr|rrr|rrr}
\toprule
\textbf{Method}
& \multicolumn{3}{c|}{\textbf{Overall}}
& \multicolumn{3}{c|}{\textbf{Stage 0}}
& \multicolumn{3}{c|}{\textbf{Stage 1}}
& \multicolumn{3}{c|}{\textbf{Stage 2}}
& \multicolumn{3}{c|}{\textbf{Stage 3}}
& \multicolumn{3}{c}{\textbf{Stage 4}} \\
\cmidrule(lr){2-4}\cmidrule(lr){5-7}\cmidrule(lr){8-10}\cmidrule(lr){11-13}\cmidrule(lr){14-16}\cmidrule(lr){17-19}
& \textbf{F1} & \textbf{Sens} & \textbf{Spec}
& \textbf{F1} & \textbf{Sens} & \textbf{Spec}
& \textbf{F1} & \textbf{Sens} & \textbf{Spec}
& \textbf{F1} & \textbf{Sens} & \textbf{Spec}
& \textbf{F1} & \textbf{Sens} & \textbf{Spec}
& \textbf{F1} & \textbf{Sens} & \textbf{Spec} \\
\midrule
\textbf{InternVL-2B} &  4.00 & 20.00 & 80.00 &  0.00 &  0.00 & 100.00 &  0.00 &  0.00 & 100.00 &  0.00 &  0.00 & 100.00 & 20.02 & 100.00 &  0.00 &  0.00 &  0.00 & 100.00 \\
\textbf{LLaVA-Med}   &  0.00 &  0.00 & 100.00 &  0.00 &  0.00 & 100.00 &  0.00 &  0.00 & 100.00 &  0.00 &  0.00 & 100.00 &  0.00 &  0.00 & 100.00 &  0.00 &  0.00 & 100.00 \\
\textbf{MedGemma-4B} & 30.34 & 45.27 & 90.47 & 93.89 & 90.17 & 97.69 & 17.58 & 33.33 & 94.76 & 11.63 &  6.45 & 98.34 &  1.62 &  1.56 & 88.56 & 26.96 & 94.83 & 73.01 \\
\textbf{MedGemma-27B}& 33.50 & 45.95 & 91.20 & 97.57 & 98.73 & 95.58 & 35.90 & 29.17 & 99.29 &  8.56 &  4.52 & 99.64 &  1.00 &  0.78 & 92.96 & 24.45 & 96.55 & 68.53 \\
\textbf{RETFound}    & \textbf{40.52} & 53.08 & 90.74 & 87.69 & 85.26 & 88.85 &  9.38 & 25.00 & 91.30 & 16.42 &  9.03 & 99.64 & 55.45 & 87.50 & 83.97 & 33.66 & 58.62 & 89.94 \\
\midrule
\textbf{VOLMO-2B}    & 39.45 & 55.26 & 89.72 & 79.28 & 66.40 & 98.65 &  8.73 & 41.67 & 82.70 & 18.92 & 11.29 & 97.03 & 47.75 & 82.81 & 79.47 & 42.57 & 74.14 & 90.76 \\
\bottomrule
\end{tabular}
}
}
\end{table*}

\section{Results}

\subsection{Ophthalmology image-description generation}

Table~\ref{tab:volmo_comparison} summarizes the performance of VOLMO-2B and baseline models on ophthalmology image-description generation. Overall, VOLMO-2B outperformed all baselines across nearly all automatic metrics. For BLEU-1, VOLMO-2B achieved the highest score ($0.1741$), compared with LLaVA-Med ($0.1181$), MedGemma-4B ($0.1083$), MedGemma-27B ($0.0912$), and InternVL-2B ($0.0796$), with all pairwise differences statistically significant ($p < 0.0001$). Similarly, VOLMO-2B attained the highest ROUGE-L score ($0.2170$), surpassing LLaVA-Med ($0.1644$), MedGemma-4B ($0.1212$), MedGemma-27B ($0.1114$), and InternVL-2B ($0.1070$), again with $p < 0.0001$ across all comparisons.  

For semantic similarity metrics, VOLMO-2B achieved the best BERTScore ($0.7140$), outperforming LLaVA-Med ($0.6867$), InternVL-2B ($0.6464$), MedGemma-4B ($0.6428$), and MedGemma-27B ($0.6397$), all with $p < 0.0001$. The only exception was SBERT similarity, where MedGemma-27B exceeded VOLMO-2B ($0.4855$ vs.\ $0.4727$, $p < 0.0001$).

Figure~\ref{fig:medical_ai_comparison} further provides qualitative examples.  
In panel~(a), the reference annotation identified recurrent endophthalmitis; VOLMO-2B correctly recognized endogenous endophthalmitis, whereas baseline models generated lengthy, non-specific anatomical descriptions.  
In panel~(b), VOLMO-2B concisely identified the key corneal pathology, while several baselines hallucinated unrelated surgical procedures.  
In panel~(c), VOLMO-2B accurately recognized the OCTA modality and the relevant anatomical regions (optic nerve head and peripapillary areas), whereas baseline models misclassified the images as fundus photographs or provided irrelevant cross-sectional interpretations inconsistent with OCTA imaging.

\subsection{Disease screening and staging classification}

\subsubsection*{Disease screening}
\label{sec:disease_screening}

\cref{tab:volmo_classification_main} summarizes disease screening performance across the 12 primary ophthalmological conditions and signs. As described earlier, in addition to the four MLLM baselines, we also included RETFound as a domain-specific ophthalmology vision foundation model. Overall, VOLMO-2B achieved the strongest performance across metrics and conditions/signs, with a macro-F1 score of 87.41\%, outperforming RETFound (80.87\%), MedGemma-27B (61.75\%), MedGemma-4B (60.65\%), LLaVA-Med (38.53\%), and InternVL-2B (34.83\%). VOLMO-2B obtained the highest F1-score in 8 of the 12 conditions/signs and exceeded 90\% F1 in 6 conditions/signs: myopic fundus (98.97\%), macular edema (96.00\%), scar (96.07\%), diabetic retinopathy (92.31\%), glaucoma (91.89\%), and AMD (90.34\%). Sensitivity and specificity showed similar performance trends across conditions. These findings are consistent with recent results reported on related datasets using task-specific models. For instance, Nakayama et al.~\cite{nakayama2024brset} achieved an F1-score of 89\% for DR screening on the BRSET dataset.

\paragraph{Comparison with MLLM baselines.}
VOLMO-2B consistently outperformed all MLLM baselines across the 12 disease categories and all three evaluation metrics. Compared with InternVL-2B (identical backbone size), VOLMO-2B achieved absolute F1-score improvements of 40--60\%. Examples include glaucoma (91.89\% vs.\ 31.79\%), drusen (84.47\% vs.\ 34.17\%), and scar (96.07\% vs.\ 33.18\%).

Performance among existing medical \mnames{} was markedly lower and highly variable. LLaVA-Med produced F1-scores between 33.39\% and 59.67\%. VOLMO-2B matched or outperformed both MedGemma-4B and MedGemma-27B in 11 of 12 conditions and signs by up to 60\%, with the only exception being DR, where VOLMO-2B was only slightly lower (92.31\% vs.\ 94.25\% for MedGemma-4B and 92.83\% for MedGemma-27B). MedGemma’s performance variability (34.20\%–94.25\%) may reflect its training corpus composition: according to its technical report, MedGemma was pretrained heavily on EyePACS (199,258 images) and internal retinal datasets focused primarily on DR~\cite{sellergren2025medgemma}. This likely contributes to its competitiveness on DR and macular edema, but is limited to other ophthalmological conditions and signs, which may explain poorer performance elsewhere. Notably, MedGemma-27B did not consistently outperform MedGemma-4B, with the 4B model surpassing the 27B variant in 4 of 12 conditions and signs (e.g., glaucoma: 63.74\% vs.\ 32.78\%), suggesting that increasing model size alone does not guarantee improved ophthalmology performance without domain-aligned training data.

\paragraph{Comparison with ophthalmology vision foundation models.}
VOLMO-2B also outperformed RETFound in 8 of the 12 conditions and signs, with absolute F1-score improvements of up to 24\% (e.g., nevus: 84.46\% vs.\ 60.07\%). In the remaining conditions and signs where RETFound exceeded VOLMO-2B, the differences were small (e.g., hemorrhage: 76.37\% vs.\ 79.43\%).
Importantly, RETFound is a vision transformer–based encoder that requires task-specific fine-tuning. Following its original study~\cite{zhou2023foundation}, we trained 12 separate RETFound models (one per disease condition), evaluating both frozen and unfrozen strategies; the frozen strategy yielded superior average performance (macro-F1: 80.87\% vs.\ 78.17\%), and thus frozen results are reported. RETFound is therefore optimized for each individual task, whereas VOLMO-2B is a single unified multimodal model without task-specific fine-tuning. 

\subsubsection*{Disease staging classification}

Table~\ref{tab:disease_stage} presents the results for disease staging classification. Consistent with the disease screening task, VOLMO-2B achieved the strongest overall performance among all evaluated models. For macular hole, VOLMO-2B reached an overall F1-score of 92.62\%, with consistently high stage-specific performance: 91.13\% (Stage~2), 92.18\% (Stage~3), and 94.53\% (Stage~4). For DR staging, VOLMO-2B achieved the best overall F1-score of 46.80\%. 

\paragraph{Comparison with MLLMs.}

As with the screening task, VOLMO-2B substantially outperformed all MLLM baselines across both diseases. InternVL-2B and LLaVA-Med showed near-zero F1-scores for most severity levels. For example, InternVL-2B produced a nonzero score only for macular hole Stage~2 (13.81\%), while the remaining stages were close to 0\%. LLaVA-Med achieved 50.00\% for macular hole Stage~4 but 0\% across all other macular hole stages and all DR stages.

Performance for the MedGemma variants was comparatively higher yet remained suboptimal. For macular hole, MedGemma-4B achieved F1-scores ranging from 0 to 34.27, and MedGemma-27B ranged from 6.11 to 48.78—far below VOLMO-2B’s consistent 90\%+ performance.  
For DR staging, both MedGemma variants performed best on the no-DR class (Stage~0)—93.08\% (4B) and 92.87\% (27B)—surpassing VOLMO-2B (78.08\%) and RETFound (83.30\%). However, their accuracy declined sharply for patients with DR. For proliferative DR (Stage~4), MedGemma-4B reached 49.64\% and MedGemma-27B 42.56\%, in contrast to 52.17\% for VOLMO-2B and 54.42\% for RETFound.  
Taken together with the screening results, these findings suggest that although MedGemma models can detect the presence of DR, they struggle to distinguish finer severity levels—a limitation likely stemming from their DR-focused pretraining data, which lacks broader coverage of non-DR ophthalmological pathologies~\cite{sellergren2025medgemma}.

Notably, MedGemma-27B did not consistently outperform the smaller 4B variant. In four of the twelve conditions and signs, MedGemma-4B achieved higher scores—sometimes by large margins (e.g., glaucoma: 63.74\% vs.\ 32.78\%)—indicating that increasing model size alone does not guarantee improved performance without domain-aligned training data.

\paragraph{Comparison with ophthalmology vision foundation models.}

Consistently, we fine-tuned RETFound separately for macular hole and DR following its original design, whereas VOLMO-2B was evaluated as a single unified model without task-specific tuning. Overall, VOLMO-2B outperformed RETFound across most disease stages.

For macular hole, VOLMO-2B achieved a 24\% higher overall F1-score (92.62\% vs.\ 68.12\%), with stage-specific gains across Stage~2 (91.13\% vs.\ 66.25\%), Stage~3 (92.18\% vs.\ 64.14\%), and Stage~4 (94.53\% vs.\ 73.98\%).  
For DR staging on EyePACS, VOLMO-2B produced a higher overall F1-score than RETFound (46.80\% vs.\ 43.42\%). VOLMO-2B showed clear advantages in intermediate stages—Stage~1 (18.60\% vs.\ 11.85\%), Stage~2 (44.78\% vs.\ 34.75\%), and Stage~3 (40.35\% vs.\ 32.77\%)—with comparable performance to RETFound for proliferative DR (Stage~4: 52.17\% vs.\ 54.42\%).

Taken together, the results also illustrate the substantially greater difficulty of disease staging compared with screening. In DR,
for example, models achieved up to 94\% F1-scores for screening, yet staging performance dropped to about 47\% even for the best-performing model. Both VOLMO-2B and RETFound achieved moderate performance for Stage~0 (78.08\% and 83.30\%) and Stage~4 (52.17\% and 54.42\%), but all models struggled with intermediate severity levels (F1-scores 11–45\%). This pattern mirrors clinical challenges, where distinguishing early and intermediate disease stages remains difficult~\cite{gulshan2016development, chen2019multi}. Even with task-specific fine-tuning,
for example, RETFound remained suboptimal for Stage~1 (11.85\%).

Finally, it is also important to note that although VOLMO-2B was not fine-tuned separately for each condition or dataset, its test sets were derived from standard benchmark splits. To further evaluate its generalization, we performed external validation across multiple independent cohorts, as detailed in the Manual and external evaluations section.

\subsection{Assessment-and-Management Generation}

Figure~\ref{fig:medical_model_comparison} summarizes model performance on \textit{Clinical Assessment and Plan}, \textit{Treatment Recommendations} and {Follow-up Care}. Across both tasks, VOLMO-2B consistently achieved the highest overall performance.

For \textit{Clinical Assessment and Plan}, VOLMO-2B achieved the highest BERTScore (0.687), surpassing MedGemma-27B (0.654), MedGemma-4B (0.645), InternVL-2B (0.618), and LLaVA-Med (0.580). The SBERT-based evaluation showed a larger separation: VOLMO-2B reached 0.702, ahead of InternVL-2B (0.536), MedGemma-27B (0.531), MedGemma-4B (0.413), and LLaVA-Med (0.307).

For \textit{Treatment Recommendations}, VOLMO-2B again led on BERTScore (0.680), followed by InternVL-2B (0.650), MedGemma-4B (0.646), MedGemma-27B (0.643), and LLaVA-Med (0.575). SBERT results mirrored this trend with wider gaps: VOLMO-2B obtained 0.625, compared with 0.510 for InternVL-2B, 0.438 for MedGemma-4B, 0.402 for MedGemma-27B, and 0.334 for LLaVA-Med.

For \textit{Follow-up Care}, VOLMO-2B maintained the top BERTScore (0.693), outperforming InternVL-2B (0.667), MedGemma-4B (0.645), MedGemma-27B (0.599), and LLaVA-Med (0.559). SBERT again highlighted a larger margin: VOLMO-2B scored 0.615, exceeding InternVL-2B (0.549), MedGemma-4B (0.478), LLaVA-Med (0.406), and MedGemma-27B (0.265).

These findings suggest that MLLMs still face significant challenges in generating high-quality clinical assessments and treatment plans, particularly when stepwise reasoning is required. Notably, SBERT scores for baseline models were as low as 0.30 when compared to clinician-written references, indicating limited semantic alignment. This is consistent with recent findings on the limitations of language-only LLMs in reasoning over patient reports~\cite{qiu2025quantifying}. It reported sharp performance declines as tasks progressed from diagnosis to treatment and follow-up. In our evaluation, these challenges are further amplified in the multimodal setting, where models need to integrate diverse patient information and perform multi-step clinical reasoning to produce coherent and clinically grounded outputs.

\subsection*{Manual and external evaluations}

\subsubsection{Manual Evaluation of Ophthalmology Image-Description Generation}

Table~\ref{tab:manual_evaluation} summarizes the manual evaluation results for the ophthalmological image descriptions generated by the models. Two ophthalmology residents independently assessed 40 model-generated outputs from Section~\ref{image-description}, using three criteria—conciseness, accuracy, and readability—based on the annotation guidelines detailed in \cref{sec:imgdesc_annotation_guidelines}. Four models were selected for evaluation based on automatic metric performance: InternVL-2B, MedGemma-4B, MedGemma-27B, and VOLMO-2B. To minimize evaluator bias, model identities were anonymized, and the output order was randomized for each sample.

As shown in Figure~\ref{fig:manual_eval_avg}, VOLMO-2B consistently achieved the highest scores across all three evaluation dimensions, with especially notable improvements in conciseness and readability. For conciseness, VOLMO-2B averaged 4.43/5, outperforming InternVL-2B (2.68), MedGemma-4B (2.62), and MedGemma-27B (2.06). In terms of readability, it scored 4.41/5, compared to InternVL-2B (3.39), MedGemma-4B (2.67), and MedGemma-27B (2.38). VOLMO-2B also led in accuracy with an average score of 2.82/5, ahead of MedGemma-27B (2.66), MedGemma-4B (2.34), and InternVL-2B (2.28). However, accuracy scores were generally lower across all models compared to conciseness and readability.
Figure~\ref{fig:manual_eval_example} further provides illustrative examples: two cases where VOLMO-2B clearly outperformed other models, and two cases where its accuracy was comparable, but its outputs were more concise and professionally written. In example (b), the description references OCT scans obtained before and after surgery, including timing—details that cannot be inferred from the image alone. This illustrates a broader limitation and reflects open challenges noted in recent literature~\cite{lozano2025biomedica,chen2025compound}: disentangling and faithfully attributing what information is directly extractable from the image versus what requires external or structured textual context.

Notably, across all 40 samples, there was no instance in which VOLMO-2B underperformed all other models on every evaluation criterion.

\subsubsection{External Evaluation of Disease Screening and Staging Classification}
\label{subsec:external-validations}

Table~\ref{tab:combined_three_parts} presents external validation results across three independent patient cohorts. On the UKB cohort, VOLMO-2B achieved the highest F1-score for AMD screening (64.58\%~$\pm$~9.30\%), outperforming RETFound (46.39\%~$\pm$~9.95\%) and all MLLM baselines (32.94\%--41.47\%). For DR screening, VOLMO-2B achieved competitive performance (49.86\%~$\pm$~9.06\%), approaching RETFound (53.36\%~$\pm$~9.77\%) while exceeding all other MLLMs.
For DR staging classification, VOLMO-2B achieved the highest F1-score (51.45\%) on the Sydney Innovation
cohort, outperforming MedGemma-4B (44.50\%), RETFound (40.18\%), and other baselines.
On the SUSTech cohort, VOLMO-2B achieved performance comparable to RETFound (39.45\% vs.\ 40.52\%), with both models substantially outperforming the MLLM baselines.

While VOLMO-2B consistently demonstrated competitive or superior performance across the three independent cohorts, the variation in performance across external datasets underscores the well-documented challenges of generalization in ophthalmological AI. Prior studies have reported reduced performance when models are applied to external populations~\cite{chen2025ai,domalpally2021real,qiu2025quantifying}. More broadly, research in medical AI has highlighted that few studies conduct external validations or involve clinician-led manual evaluations~\cite{gangaputra2013comparison}, underscoring the need for rigorous reporting standards and domain adaptation strategies to ensure model robustness and clinical reliability.

\section{Discussion}

In this work, we present VOLMO, a model-agnostic and data-open framework for and openly releasing domain-specific MLLMs for ophthalmology. VOLMO uses a three-stage training pipeline: ophthalmology knowledge pretraining, domain task fine-tuning, and multi-step clinical reasoning and synthesis, each built entirely from publicly available datasets with permissive licenses. Together, these stages curate over 110{,}000 multimodal instances spanning image--text pairs, annotated disease labels, and full patient case reports. We systematically compared VOLMO-2B with five representative baselines, including general-domain MLLMs, medical MLLMs, and an ophthalmology-specific vision foundation model. Evaluation covered ophthalmological image description generation, disease screening, disease staging, and assessment and treatment plan generation, each aligned with the corresponding training stage. We further conducted manual evaluations and assessed cross-population generalization through external validations on three independent patient cohorts for AMD and DR screening and DR severity grading.

\subsection{VOLMO overcomes limitations of current MLLMs in ophthalmology}
Recent studies have performed extensive benchmarking of LLMs in ophthalmology. These works consistently show that LLMs perform well on language-only ophthalmological tasks, including general knowledge tests~\cite{antaki2025performance}, patient-facing question answering~\cite{srinivasan2025ophthalmological,abacha2023investigation}, and clinical information extraction~\cite{kim2024leme}. However, these same studies have also revealed significant limitations for multimodal ophthalmological tasks. Ophthalmology is inherently vision-centric: accurate diagnosis requires integrating imaging, such as fundus photography and OCT, with clinical history and examination findings~\cite{khan2021global}. Prior benchmarking efforts~\cite{qin2025lmod,qin2025lmod+} across more than 20 MLLMs demonstrate that state-of-the-art models struggle with essential aspects of ophthalmological image interpretation, including anatomical recognition, lesion identification, and disease severity classification. For primary eye diseases, diagnostic accuracy was often near random. Similar findings from Xu et al.~\cite{xu2025benchmarking} report that even the best MLLMs achieved only about 55\% accuracy on key classification tasks. Pioneering work has shown promising results~\cite{li2025visionunite}, suggesting that domain-specific MLLMs can improve the effectiveness in ophthalmologic applications. However, existing efforts have largely focused on single imaging modalities (e.g., fundus photographs) and have yet to capture key aspects of ophthalmologist workflows—such as synthesizing multimodal inputs, performing differential diagnosis, and drafting comprehensive assessments and treatment plans.


VOLMO directly addresses these gaps through a multi-stage, model-agnostic, ophthalmology-specific training pipeline. Across all evaluation settings, a compact 2B-parameter model trained with VOLMO outperformed general-domain and medical MLLM baselines, even those with substantially larger parameter counts. Beginning with direct MLLM comparisons, VOLMO-2B achieved the strongest performance for ophthalmological image description: it obtained the highest automatic metrics (e.g., an 11\% absolute improvement over MedGemma-27B on ROUGE-L and an 8\% improvement for BERTScore) and had the highest conciseness (4.43 vs. 2.06) and readability (4.41 vs. 2.38)
manually evaluated by clinicians. For disease screening, VOLMO-2B achieved an average F1 of 87.41\% across 12 conditions and signs, outperforming MedGemma-27B (61.75\%), MedGemma-4B (60.65\%), LLaVA-Med (38.53\%), and InternVL-2B (34.83\%). For disease staging classification, VOLMO-2B also achieved the best performance, including 92.62\% for macular hole severity (e.g., more than 70\% absolute improvement over MedGemma-27B) and 46.80\% for DR severity (3\% absolute improvement). For assessment and treatment plan generation, VOLMO-2B produced more clinically aligned and coherent outputs, achieving 17–44\% higher semantic similarity to clinician-written documentation compared with the baselines. Finally, in external validations, VOLMO-2B achieved the highest performance across three independent patient cohorts, e.g., 64.58\% F1 for AMD screening on UK Biobank (vs. 33.77\% for MedGemma-27B).
Notably, all these results were achieved using a single 2B-parameter model that can run inference on widely accessible hardware, including modest consumer-grade GPUs (e.g., RTX~3050--4090) or laptops equipped with 8--16\,GB RAM using quantized formats. This shows the potential for deployment in clinically resource-limited environments.

\subsection{Potentials of VOLMO over prior foundation models}
In addition, as mentioned earlier, the paradigm of AI in ophthalmology has shifted from task-specific CNNs to vision foundation models~\cite{luo2025survey,chia2024foundation}. To contextualize VOLMO's performance within this broader landscape, we further compared against RETFound, one of the most widely used ophthalmology-specific vision foundation models~\cite{zhang2024retfound,chen2025independent,hou2026can}. Vision-only models such as RETFound cannot perform generative tasks and require task- and condition-specific fine-tuning. Accordingly, we thoroughly fine-tuned RETFound for each eye condition and each classification task, following both partial and full layer-tuning strategies established in the original study~\cite{zhou2023foundation} to obtain its best performance.
For disease screening, VOLMO-2B outperformed each fine-tuned RETFound model in 8 of the 12 conditions, achieving up to 24\% absolute improvement. For disease staging, VOLMO-2B also achieved  better performance, including around 24\% higher F1 for macular hole severity classification (92\% vs.\ 68\%) and 4\% higher F1 for DR severity grading (47\% vs.\ 43\%). When evaluated on three independent patient cohorts, VOLMO-2B also achieved in general better generalization, with 19\% higher F1 for AMD screening (65\% vs.\ 46\%) and 11\% higher F1 for DR classification (51\% vs.\ 40\%), alongside competitive performance on other conditions and signs. Importantly, all of these results were achieved by a single VOLMO-2B model without any task- or condition-specific fine-tuning. These findings collectively highlight two key advantages of VOLMO over prior ophthalmology foundation models. First, VOLMO enables multimodal reasoning and generative capabilities that vision-only and contrastive vision--language models may not be feasible, as demonstrated by its performance in image description and assessment and plan generation. Second, VOLMO provides robust and consistent performance across diverse eye diseases and had stronger cross-population generalization, with a single unified model capable of handling all tasks without retraining.

\subsection{Limitations and Future Directions}

While VOLMO provides a practical and openly accessible framework, it also has several important limitations.

First, VOLMO is intended as a model-agnostic, open-data foundation that can be adapted using local clinical data. For example, VOLMO can be fine-tuned at individual institutions to generate more tailored assessments and management plans aligned with local workflows. However, using open datasets also imposes inherent constraints. This is reflected in our manual evaluations: although VOLMO-2B achieved the highest clinician-rated accuracy among all models (2.82 out of 5), absolute accuracy scores remained modest. Similarly, the disease annotations used in Stage~2 were derived from publicly available datasets with permissive licenses, which may not fully capture the diversity, severity spectrum, or imaging variability seen in real-world populations. These limitations have been widely discussed in prior reviews on ophthalmic datasets~\cite{khan2021global}.

Second, while we conducted head-to-head comparisons with several strong and widely adopted models—including general-domain (e.g., InternVL), medical-domain (e.g., MedGemma), and ophthalmology-specific vision foundation models (e.g., RETFound fine-tuned for each task)—we did not evaluate all newly emerging models due to the rapid pace of advancement in this space. To enable broader and ongoing benchmarking, we have made the VOLMO framework publicly available, allowing future models to be evaluated and improved upon by the community.

Third, although we evaluated the effectiveness across three core ophthalmology applications and validated its performance through both expert assessments and external cohort testing, further evaluation is warranted across additional tasks and settings. Our results highlight VOLMO-2B’s strong potential, but also underscore the importance of adapting the framework with local clinical data—especially to improve performance on complex, multi-step generative reasoning tasks such as assessment and treatment planning.

Finally, broader efforts are needed to evaluate the effectiveness of MLLMs like VOLMO within clinical workflows. It is important to assess their utility, safety, and impact through prospective studies that reflect how these models integrate into everyday practice. Such evaluations can better inform the development, refinement, and responsible adoption of MLLMs in ophthalmology and beyond.

\section*{Acknowledgements}

This study is supported by the National Institutes of Health National Library of Medicine under Award Number
R01LM014604 and R00LM014024. It was also supported in part by the Intramural Research Program of the National Institutes of Health. The contributions of the NIH authors were made as part of their official duties as NIH federal employees, are in compliance with agency policy requirements, and are considered Works of the United States Government. However, the findings and conclusions presented in this paper are those of the author(s) and do not necessarily reflect the views of the NIH or the U.S. Department of Health and Human Services. The UK Biobank data were accessed through application 41910. Some icons are from flaticon.com. 

\bibliography{refs}

\clearpage
\appendix

\section{Metrics in Detail}

We present more detailed definitions of metrics used to evaluate VOLMO across three ophthalmological aspects. 

\textbf{BLEU Score}: 
BLEU measures n-gram precision between generated and reference texts. It is particularly important for evaluating medical terminology accuracy from single words (BLEU-1) to 4-word phrases (BLEU-4).  
The definition is:
\begin{equation}
\text{BLEU} = \text{BP} \cdot \exp\left(\sum_{n=1}^{N} w_n \log p_n\right)
\end{equation}

\textbf{BERTScore}: 
BERTScore evaluates semantic similarity using contextual embeddings, recognizing synonyms and preserving meaning beyond lexical overlap.  
Its definition is as follows:
\begin{equation}
F_{\text{BERT}} = 2 \cdot \frac{P_{\text{BERT}} \cdot R_{\text{BERT}}}{P_{\text{BERT}} + R_{\text{BERT}}}
\end{equation}
where  
\begin{equation}
P_{\text{BERT}} = \frac{1}{|x|} \sum_{x_i \in x} \max_{y_j \in y} x_i^T y_j, \quad
R_{\text{BERT}} = \frac{1}{|y|} \sum_{y_j \in y} \max_{x_i \in x} x_i^T y_j
\end{equation}

\textbf{SBERT Similarity}: 
SBERT measures sentence-level semantic similarity between full descriptions, assessing global coherence.  
Its definition is:
\begin{equation}
\text{SBERT}_{\text{sim}} = \frac{\vec{u} \cdot \vec{v}}{||\vec{u}|| \cdot ||\vec{v}||}
\end{equation}

\textbf{Precision}: 
Precision assesses the proportion of predicted positives that are truly positive. The formula is:
\begin{equation}
\text{Precision} = \frac{\text{TP}}{\text{TP} + \text{FP}}
\end{equation}

\textbf{Recall}: 
Recall evaluates sensitivity by measuring the proportion of actual positives correctly identified.  
The definition is:
\begin{equation}
\text{Recall} = \frac{\text{TP}}{\text{TP} + \text{FN}}
\end{equation}

\textbf{F1 Score}: 
F1 balances precision and recall for robust comparison across models.  
Its definition is like:
\begin{equation}
\text{F1} = 2 \cdot \frac{\text{Precision} \cdot \text{Recall}}{\text{Precision} + \text{Recall}}
\end{equation}

\textbf{ROUGE-L}: 
ROUGE-L captures structural coherence using the longest common subsequence.  
It is defined as:
\begin{equation}
\text{ROUGE-L} = \frac{(1 + \beta^2)R_{\text{lcs}}P_{\text{lcs}}}{R_{\text{lcs}} + \beta^2P_{\text{lcs}}}
\end{equation}
where  
\begin{equation}
R_{\text{lcs}} = \frac{\text{LCS}(X,Y)}{\text{len}(Y)}, \quad P_{\text{lcs}} = \frac{\text{LCS}(X,Y)}{\text{len}(X)}
\end{equation}


\cref{tab:volmo_comparison_full} Provides more quantitative results. 
\begin{table*}[t]
\centering
\caption{Statistical comparison of baseline models against VOLMO-2B for image description tasks. Values show mean $\pm$ standard deviation. P-values from Wilcoxon signed-rank test are shown in parentheses.}
\label{tab:volmo_comparison_full}

\resizebox{\textwidth}{!}{%
{\scriptsize
\begin{tabular}{l|ccccccc}
\toprule
\textbf{Model} & \textbf{ROUGE-L F1} & \textbf{BLEU-1} & \textbf{BLEU-2} & \textbf{BLEU-3} & \textbf{BLEU-4} & \textbf{BERTScore F1} & \textbf{SBERT Similarity} \\
\midrule
InternVL-2B 
& $0.1070 \pm 0.0042$ & $0.0796 \pm 0.0044$ & $0.0273 \pm 0.0017$ & $0.0095 \pm 0.0007$ & $0.0046 \pm 0.0004$ & $0.6464 \pm 0.0032$ & $0.4409 \pm 0.0092$ \\
& ($p < 0.0001$) & ($p < 0.0001$) & ($p < 0.0001$) & ($p < 0.0001$) & ($p < 0.0001$) & ($p < 0.0001$) & ($p < 0.0001$) \\
\midrule
LLaVA-Med 
& $0.1644 \pm 0.0037$ & $0.1181 \pm 0.0068$ & $0.0410 \pm 0.0030$ & $0.0170 \pm 0.0014$ & $0.0094 \pm 0.0008$ & $0.6867 \pm 0.0028$ & $0.4468 \pm 0.0117$ \\
& ($p < 0.0001$) & ($p < 0.0001$) & ($p < 0.0001$) & ($p < 0.0001$) & ($p < 0.0001$) & ($p < 0.0001$) & ($p < 0.0001$) \\
\midrule
MedGemma-4B 
& $0.1212 \pm 0.0052$ & $0.1083 \pm 0.0068$ & $0.0440 \pm 0.0033$ & $0.0187 \pm 0.0017$ & $0.0092 \pm 0.0010$ & $0.6428 \pm 0.0034$ & $0.4669 \pm 0.0090$ \\
& ($p < 0.0001$) & ($p < 0.0001$) & ($p < 0.0001$) & ($p < 0.0001$) & ($p < 0.0001$) & ($p < 0.0001$) & ($p < 0.0001$) \\
\midrule
MedGemma-27B 
& $0.1114 \pm 0.0043$ & $0.0912 \pm 0.0057$ & $0.0342 \pm 0.0023$ & $0.0128 \pm 0.0010$ & $0.0060 \pm 0.0005$ & $0.6397 \pm 0.0028$ & $\mathbf{0.4855 \pm 0.0083}$ \\
& ($p < 0.0001$) & ($p < 0.0001$) & ($p < 0.0001$) & ($p < 0.0001$) & ($p < 0.0001$) & ($p < 0.0001$) & ($p < 0.0001$) \\
\midrule
VOLMO-2B 
& $\mathbf{0.2170 \pm 0.0074}$ & $\mathbf{0.1741 \pm 0.0080}$ & $\mathbf{0.0764 \pm 0.0057}$ & $\mathbf{0.0390 \pm 0.0039}$ & $\mathbf{0.0224 \pm 0.0027}$ & $\mathbf{0.7140 \pm 0.0042}$ & $0.4727 \pm 0.0113$ \\
\bottomrule
\end{tabular}
}
}
\end{table*}

\lstset{
    basicstyle=\ttfamily\small,
    breaklines=true,
    breakatwhitespace=false,
    columns=fullflexible,
    keepspaces=true,
    showstringspaces=false,
    frame=single,
    framesep=3pt,
    xleftmargin=3pt,
    xrightmargin=3pt
}

\section{Prompts}
\label{sec:prompts}

\subsection{Image Caption Revision}
\label{subsec:img_cap_revision}

\begin{lstlisting}
Using the information provided below, please rewrite the description 
content in an educated, professional tone. 
The answer should just be a paragraph. 
Do not add any new information or make up details, simply rephrase 
what is given in the information. 
Your answer should start directly with a description (for example, 
'This image depicts... or The image shows ..., please don't use 
these two, think about others') and avoid any introductory phrases 
like 'Based on the provided...' or similar.

Description:
{caption}

Output your answer in the following format:

Answer:
<Your detailed description>
\end{lstlisting}

\subsection{Binary Condition Screening}
\label{subsec:bin_con_screen}

\begin{lstlisting}
This is a colorful fundus image. 
Please tell me whether this image shows myopic fundus. 
Answer in format: TRUE or FALSE.
\end{lstlisting}

\subsection{Disease Stage Grading}
\begin{lstlisting}
Here is an eye image. Rate the presence of diabetic retinopathy in each image 
on a scale of 0 to 4, according to the following scale:

0 - No DR
1 - Mild
2 - ModeRate
3 - Severe
4 - Proliferative DR

Your response should only contain a single number, representing your rating.
\end{lstlisting}


\subsection{Clinical Assessment Synthesis}
\label{subsec:assess_plan_prompts}
\begin{lstlisting}
##################################################
### Task: Provide Differential Diagnoses ###
You are an ophthalmologist. 
You are given the clinical profile of a patient. 
Based on the clinical images and profile, please provide a differential diagnosis. 
For each potential diagnosis, include severity level.

### Input: Clinical Profile ###
[PATIENT CLINICAL PROFILE]

[MEDICAL HISTORY]
1. Medical History: Sequential nonarteritic anterior ischemic optic neuropathy; complete visual field defect; visual acuity of 1/50 Snellen in each eye

[FAMILY HISTORY]
No family history reported

[SYMPTOMS]
1. Symptom: Vision loss
2. Symptom: Further painless decrease in visual acuity
3. Symptom: Visual acuity of light perception
4. Symptom: Whitish, dense vitreal opacities; Duration: 2 months; Progression: Unchanged
5. Symptom: Deeply atrophic optic nerves
6. Symptom: Adherent posterior hyaloid in both eyes
7. Symptom: Localized retinal detachment (left eye)
8. Symptom: Retinal tear (right eye)
9. Symptom: Epiretinal remnants from adherent hyaloid (right eye); Duration: Several months; Progression: Unchanged

[EXAMINATION FINDINGS]
1. Examination Type: Ophthalmic examination (visual acuity); Finding: Severe visual impairment with light perception only.; Note: The noted decrease in visual acuity was painless, contradicting typical expectations in similar cases.
2. Examination Type: Fundoscopic examination; Finding: Dense vitreal opacities and atrophic optic nerves with unremarkable retinas.
3. Examination Type: Histopathological analysis (microscopic examination); Finding: Presence of vimentin-positive cells of mesenchymal origin without retinal or neuronal differentiation.; Note: No therapeutic benefit or cellular integration was observed post intravitreal injection of stem cells, contrary to potential expectations.

[DIAGNOSTIC IMAGING]
1. Imaging Type: Fundoscopic photograph - Right and Left Eyes; Finding: Dense vitreal opacities localized to the vitreous body without evidence of retinal integration or reaction. Post-vitrectomy, epiretinal remnants from the adherent hyaloid remained unchanged.; Key Results: Vitreal opacities persisted without therapeutic effects or severe complications. Vitrectomy cleared the opacities completely, with adherence of hyaloids causing retinal detachment and tear, both of which were treated.

### Expected Output Format ###
[DIFFERENTIAL DIAGNOSIS]
1. Diagnosis: [name]; Severity: [severity]
2. Diagnosis: [name]; Severity: [severity]
...
##################################################

##################################################
### Task: Identify Most Likely Diagnosis ###

Thank you for the differential diagnosis. Based on these findings, what is the most likely diagnosis? Please provide justification for your choice, including severity assessment.

### Expected Output Format ###
Diagnosis: [name]; Severity: [severity]; Justification: [explanation]
##################################################

##################################################
### Task: Provide Clinical Assessment and Plan ###

Given your diagnosis, could you provide a detailed clinical assessment and treatment plan?

### Expected Output Format ###
1. Assessment: [clinical assessment]; Plan: [treatment plan]
2. Assessment: [clinical assessment]; Plan: [treatment plan]
...
##################################################

##################################################
### Task: Recommend Treatments ###

Based on your assessment and plan, what specific treatments would you recommend? Please include immediate and long-term outcomes when available.

### Expected Output Format ###
1. Treatment: [treatment]; Immediate outcome: [outcome]; Long-term outcome: [outcome]; Justification: [justification]
2. Treatment: [treatment]; Immediate outcome: [outcome]; Long-term outcome: [outcome]; Justification: [justification]
...
##################################################

##################################################
### Task: Outline Follow-Up Care ###

Finally, please outline the recommended follow-up care for this patient. Include justification, prognosis, and any unexpected outcomes where appropriate.

### Expected Output Format ###
1. Follow-up care: [care details]; Justification: [justification]; Prognosis: [prognosis]; Unexpected outcomes: [outcomes]
2. Follow-up care: [care details]; Justification: [justification]; Prognosis: [prognosis]; Unexpected outcomes: [outcomes]
...
##################################################
\end{lstlisting}




\section{Manual Evaluation Annotation Guidelines for Image Descriptions}
\label{sec:imgdesc_annotation_guidelines}
\begin{tcolorbox}[colback=gray!5!white,colframe=gray!75!black,title=\textbf{Image Description Generation Evaluation Guidelines},breakable]
\subsection*{Task Overview}
The task is image description generation. We collected a random set of 40 images and their captions from medical literature.
\begin{itemize}
\item \textbf{Input:} an image.
\item \textbf{Output:} a generated description.
\end{itemize}
For each image, there are two reference captions:
\begin{itemize}
\item The original caption (from the source).
\item The revised caption (edited into natural sentences).
\end{itemize}
These two together should be treated as the gold standard reference.
\textbf{For evaluation, there are four generated descriptions per image. To minimize bias, these outputs are anonymized (model identity is hidden) and shuffled (the order is randomized across images).}
Your task is to evaluate each generated description on three criteria (conciseness, accuracy, readability, in a scale 1-5) and answer additional questions about clinical relevance.
\subsection*{Evaluation Criteria}
\subsubsection*{Criteria 1. Conciseness: Is the generated caption precise?}
\begin{itemize}
\item \textbf{5 (Very Concise):} Direct and focused, uses the fewest words needed to convey the essential findings, with no redundancy or irrelevant background.
\item \textbf{4 (Concise):} Mostly focused, but includes one or two extra phrases or mild repetition without significantly affecting clarity.
\item \textbf{3 (Moderately Concise):} Noticeably wordier than necessary, with some background or tangential details, though the main findings remain clear.
\item \textbf{2 (Verbose):} Overly long and distracts from describing the actual image.
\item \textbf{1 (Extremely Verbose/Unfocused):} Dominated by irrelevant details or excessive length, with little emphasis on describing the image itself.
\end{itemize}
\subsubsection*{Criteria 2. Accuracy: Does the caption correctly describe the findings?}
\begin{itemize}
\item \textbf{5 (Fully Accurate):} Correctly describes the findings with no errors or invented details.
\item \textbf{4 (Mostly Accurate):} Largely correct but contains a minor error (e.g., mislabeled sequence, small omission) while preserving the main finding.
\item \textbf{3 (Partially Accurate):} Contains both correct and incorrect elements (e.g., correct modality but missing or misdescribing key findings).
\item \textbf{2 (Poor Accuracy):} Describes the wrong modality or misrepresents findings, though it may still vaguely reference related structures.
\item \textbf{1 (Inaccurate/Hallucinated):} Fundamentally incorrect, fabricating details not present in the image.
\end{itemize}
\subsubsection*{Criteria 3. Readability: Is the caption easy to read and professional?}
\begin{itemize}
\item \textbf{5 (Very Readable):} Clear, fluent, and professional, with smooth phrasing that is easy to understand on first read.
\item \textbf{4 (Readable):} Understandable and generally smooth, though with minor awkward phrasing or slight wordiness.
\item \textbf{3 (Neutral/Clunky):} Comprehensible but clunky, with run-on sentences, inconsistent flow, or unnecessary jargon.
\item \textbf{2 (Hard to Read):} Poorly structured, fragmented, or overly jargony, requiring rereading to understand.
\item \textbf{1 (Very Poor Readability):} Confusing or incoherent, making it very difficult to extract meaning.
\end{itemize}
\subsection*{Additional Questions}
\begin{enumerate}
\item Is this image clinically relevant? (Yes / No)
\item Additional comments: (e.g., "The gold standard caption is not informative," or other notes on the quality of reference captions or model outputs.)
\end{enumerate}
\end{tcolorbox}
\section{Annotation Guidelines for Assessment and Management Generation}
\label{sec:pmc_annotation_guidelines}
\begin{tcolorbox}[colback=gray!5!white,colframe=gray!75!black,title=\textbf{PMC Case Report Annotation Guidelines},breakable]

\subsection*{Task Overview}
This task evaluates large vision-language models from an ophthalmologist's perspective on clinical case reports.

\begin{itemize}
\item \textbf{Input:} PMC case report with pre-generated responses to clinical questions.
\item \textbf{Output:} Verification and correction of machine-generated responses.
\end{itemize}

Each document includes the PMC case report URL. Review the original document and images when formulating responses. All images from the PMC report are provided at the beginning for reference.

\textbf{Note:} Machine-generated responses are organized into bullet points with corresponding "Original Text" sections. 

\subsection*{Evaluation Criteria}

For each bullet point, provide input in the "Your Response" section based on one of three categories:

\subsubsection*{1. Fully Correct}
\begin{itemize}
\item The GPT response is entirely accurate.
\item \textbf{Action:} Write "Fully Correct" in the "Your Response" section.
\end{itemize}

\subsubsection*{2. Partially Correct}
\begin{itemize}
\item The GPT response requires visual information or minor corrections.
\item \textbf{Action:} Write "Partially Correct" in the "Your Response" section, then provide your revised version.
\item \textbf{Example:} \\
\textit{Question:} Are there any relevant details in the patient's past medical history? \\
\textit{LLM output:} Down syndrome; trisomy 21; myopia in the right eye; no family history of hormonal disorders... \\
\textit{Your input:} Partially correct; Down syndrome diagnosed at birth; trisomy 21; myopia in the right eye; secondary hypothyroidism...
\end{itemize}

\subsubsection*{3. Completely Incorrect}
\begin{itemize}
\item The GPT response contains significant errors.
\item \textbf{Action:} Write "Completely Incorrect" in the "Your Response" section, briefly explain why, and provide correct information.
\end{itemize}

\subsubsection*{Additional Information}
\begin{itemize}
\item If important information is missing from GPT-generated answers, add details in the "Additional Information" section.
\item Follow the same format as GPT responses and include "Original Text" for added content.
\end{itemize}

\subsubsection*{Image References}
\begin{itemize}
\item When an answer requires image reference, begin with \textbf{[IMAGE INCLUDED]} and use specific figure IDs (e.g., "Figure 1").
\end{itemize}

\subsection*{Clinical Questions to Evaluate}

\begin{enumerate}
\item \textbf{Patient History:} Are there any relevant details in the patient's past medical or ocular history? Include demographics, chronic conditions, previous ocular conditions/surgeries, medications, and allergies.

\item \textbf{Family History:} Are there any relevant details in the patient's family history? Consider hereditary eye conditions, family history of systemic diseases, and genetic factors.

\item \textbf{Presenting Symptoms:} What are the key symptoms reported by the patient at presentation? What was the duration and progression? Focus on patient-reported symptoms only (answer "NA" if routine examination).

\item \textbf{Examination Findings:} What were the main ophthalmological examination findings at presentation? \textit{Do not include imaging or diagnostic studies here} (move to Question 5).

\item \textbf{Diagnostic Studies:} What diagnostic studies were performed? What were the key results? Include imaging (OCT, fundus photography, angiography), visual function tests, genetic/laboratory tests, and specialized tests.

\item \textbf{Assessment and Plan:} At presentation, what assessment and plan were provided? Evaluate clinical assessment, initial treatment plan, alignment with standard practice, and follow-up recommendations.

\item \textbf{Treatment:} What treatment was chosen? What was the rationale? What were immediate and long-term outcomes? Treatment may include observation/monitoring.

\item \textbf{Follow-up and Prognosis:} What follow-up care was recommended? Were there complications or unexpected developments? What is the long-term prognosis?
\end{enumerate}

\end{tcolorbox}

\subsubsection*{Diagnostic Performance}

Consider a complex presentation of orbital pathology where the ground truth included orbital metastatic breast cancer and corneal perforation with uveal prolapse. LLaVA-Med's response was clinically inadequate, producing fragmented text with repetitive medical history entries:
\begin{quote}
\raggedright
\texttt{antibiotics for 2 weeks}\\
\texttt{5. Medical History: 61-year-old African American women, blurry vision of right eye for 1 month; periorbital fullness; swelling}\\
\texttt{6. Medical History: 61-year-old African American women, blurry vision of right eye for 1 month; periorbital fullness; swelling}\\
\texttt{[...repeated 10 times...]}\\
\texttt{[DIFFERENTIAL DIAGNOSES]}\\
\texttt{1. Differential Diagnosis: Orbital cellulitis and retrobulbar abscess}\\
\texttt{[...repeated 10 times...]}
\end{quote}

In contrast, VOLMO-2B provided structured, clinically relevant differential diagnoses:
\begin{quote}
\raggedright
\texttt{[DIFFERENTIAL DIAGNOSIS]}\\
\texttt{1. Diagnosis: Pseudo-exfoliative glaucoma; Severity: Moderate}\\
\texttt{2. Diagnosis: Ex-PRESS shunt; Severity: Moderate}\\
\texttt{3. Diagnosis: Cataract surgery; Severity: N/A}
\end{quote}

While not perfectly matching the ground truth, VOLMO-2B demonstrated the ability to generate properly formatted diagnoses with severity assessments, avoiding the repetitive and fragmented output characteristic of LLaVA-Med.

In another case of vitamin A deficiency with bilateral corneal changes , the difference was even more pronounced. LLaVA-Med produced nonsensical output:
\begin{quote}
\raggedright
\texttt{-s-Pleoglottis}\\
\texttt{The image shows the results of the 2-s-Pleoglottis test for the patient. The test was performed using a 2-s-Pleoglottis probe, and the results are presented in the form of a graph...}
\end{quote}

VOLMO-2B generated structured differential diagnoses:
\begin{quote}
\raggedright
\texttt{[DIFFERENTIAL DIAGNOSIS]}\\
\texttt{1. Diagnosis: Anterior Uveitis; Severity: Moderate}\\
\texttt{2. Diagnosis: Posterior Uveitis; Severity: Moderate}\\
\texttt{3. Diagnosis: Uveal Nodules; Severity: Mild}\\
\texttt{4. Diagnosis: Ciliary body Malignant Melanoma; Severity: Not applicable (excluded)}
\end{quote}

Although not including vitamin A deficiency in this specific output, VOLMO-2B demonstrated coherent clinical reasoning and proper formatting, contrasting sharply with LLaVA-Med's complete failure to generate meaningful diagnostic content.

\subsubsection*{Clinical Assessment and Planning}

When assessing a patient with advanced rubeosis iridis , the differences in assessment quality became clinically apparent. LLaVA-Med produced disconnected content with extensive repetition:
\begin{quote}
\raggedright
\texttt{14,000 cells muL}\\
\texttt{6. Medical History: Elevated C-reactive protein levels}\\
\texttt{7. Medical History: Normal abdominal ultrasound}\\
\texttt{8. Medical History: Normal colonoscopy}\\
\texttt{[CLINICAL IMAGES]}\\
\texttt{1. Clinical Image: Retinal vasculitis with perivascular cuffing}\\
\texttt{[...repeated 49 times...]}
\end{quote}

VOLMO-2B generated a comprehensive clinical assessment:
\begin{quote}
\raggedright
\texttt{Assessment: The patient has advanced rubeosis iridis in both eyes, with neovascularization on the iris and retinal edema affecting the left eye. The condition is associated with poor visual acuity and increased risk for retinal detachment, requiring surgical intervention. The right eye has a large central corneal edema and bullous appearance, while the left eye shows signs of rubeosis iridis with a large corneal pupil and neovascularization of the iris.}
\end{quote}

This demonstrated VOLMO-2B's ability to synthesize clinical findings into coherent assessments with integrated treatment plans, while LLaVA-Med failed to move beyond repetitive listing of isolated findings.

For treatment planning, consider a case of acute angle closure glaucoma (Item 38). LLaVA-Med generated an extensive list of repetitive diagnoses:
\begin{quote}
\raggedright
\texttt{[DIAGNOSES]}\\
\texttt{1. Diagnosis: branch retinal vein occlusion (BRVO)}\\
\texttt{2. Diagnosis: macular edema (ME)}\\
\texttt{3. Diagnosis: glaucoma}\\
\texttt{[...]}\\
\texttt{11. Diagnosis: retinal vascular tortuosity (RVT)}\\
\texttt{[...repeated as items 12-56...]}
\end{quote}

In contrast, VOLMO-2B provided a structured assessment and plan:
\begin{quote}
\raggedright
\texttt{Plan: The patient was treated with oral acetazolamide, pilocarpine, eye drops (betamethasone), and topical prednisolone acetate 1\% eye drops. LPI was performed 10 days prior to initial presentation. Detailed monitoring included repeated UBM examinations to assess anterior chamber fibrin and episcleral veins. The patient was advised on avoiding eye rubbing and maintaining follow-up appointments. Follow-up at 2 and 9 days demonstrated resolution of symptoms, with best-corrected visual acuity improving to 10/10 in the right eye and 6/16 in the left eye. No recurrence of uveal effusion was noted.}
\end{quote}

VOLMO-2B demonstrated understanding of the clinical presentation and appropriate management strategies, while LLaVA-Med's output consisted primarily of repetitive, non-specific diagnoses.

\subsubsection*{Treatment and Long-term Care Recommendations}

In treatment planning for exudative retinal detachment, LLaVA-Med provided extensively repetitive, irrelevant text:
\begin{quote}
\raggedright
\texttt{sensory deficits, or other neurological symptoms; no history of trauma, surgery, or other ocular or systemic conditions; no history of any previous episodes of diplopia, transient visual loss, jaw claudication, myalgia, constitutional symptoms, motor or sensory deficits, or other neurological symptoms; no history of trauma, surgery, or other ocular or systemic conditions...}\\
\texttt{[...this exact text repeated approximately 20 times...]}
\end{quote}

VOLMO-2B generated structured treatment recommendations:
\begin{quote}
\raggedright
\texttt{1. Treatment: Photodynamic therapy (PDT) with verteporfin; Immediate outcome: Subretinal fluid was completely disappeared; vision improved to 20/200.; Long-term outcome: By 10 months, exudative retinal detachment had completely resolved, and BCVA improved to 20/20...; Justification: PDT was implemented due to the exudative retinal detachment with subretinal haemorrhage...}
\end{quote}

For long-term care planning in a complex case, LLaVA-Med produced highly repetitive content:
\begin{quote}
\raggedright
\texttt{intravitreal injection of 0.1 ml of 10\% dexamethasone}\\
\texttt{3. Examination Type: Intravitreal injection; Finding: The patient was treated with intravitreal injection of 0.1 ml of 10\% dexamethasone}\\
\texttt{[...repeated 25 times with identical text...]}
\end{quote}

VOLMO-2B provided comprehensive follow-up care recommendations:
\begin{quote}
\raggedright
\texttt{1. Follow-up care: The patient underwent OCT imaging at one-month follow-up to monitor the closure of the macular hole (MH), resolution of the ERM, and reduction in retinoschisis and macular edema.; Justification: OCT imaging was necessary to assess the structural changes in the retina...; Prognosis: The prognosis indicates partial visual recovery with VA improving to 20/40 after vitrectomy...}
\end{quote}

\begin{table*}[t]
\centering
\caption{Comparison of frozen versus unfrozen RETFound fine-tuning strategies for binary condition classification tasks. The best F1 score in each column is highlighted in bold.}
\label{tbl:retfound}

{\scriptsize
\setlength{\tabcolsep}{3.5pt}

\begin{tabular}{c|ccc||ccc||ccc}
\toprule
& \multicolumn{3}{c||}{\textbf{Glaucoma}} & \multicolumn{3}{c||}{\textbf{AMD}} & \multicolumn{3}{c}{\textbf{DR}} \\
\textbf{Model} & \textbf{F1} & \textbf{Sensitivity} & \textbf{Specificity} & \textbf{F1} & \textbf{Sensitivity} & \textbf{Specificity} & \textbf{F1} & \textbf{Sensitivity} & \textbf{Specificity} \\
\midrule
RETFound (frozen) & $\mathbf{92.33}$ & $92.33$ & $92.71$ & $87.91$ & $87.93$ & $87.87$ & $84.94$ & $84.93$ & $85.15$ \\
\midrule
RETFound (unfrozen) & $91.65$ & $86.67$ & $96.67$ & $\mathbf{90.71}$ & $85.19$ & $96.30$ & $78.73$ & $76.92$ & $80.54$ \\
\bottomrule
\end{tabular}

\vspace{0.6em}

\begin{tabular}{c|ccc||ccc||ccc}
\toprule
& \multicolumn{3}{c||}{\textbf{Drusen}} & \multicolumn{3}{c||}{\textbf{Hemorrhage}} & \multicolumn{3}{c}{\textbf{Hypertensive Retinopathy}} \\
\textbf{Model} & \textbf{F1} & \textbf{Sensitivity} & \textbf{Specificity} & \textbf{F1} & \textbf{Sensitivity} & \textbf{Specificity} & \textbf{F1} & \textbf{Sensitivity} & \textbf{Specificity} \\
\midrule
RETFound (frozen) & $68.98$ & $69.03$ & $69.22$ & $\mathbf{79.43}$ & $79.87$ & $79.89$ & $63.34$ & $64.03$ & $64.46$ \\
\midrule
RETFound (unfrozen) & $47.87$ & $18.64$ & $90.59$ & $72.94$ & $88.24$ & $58.82$ & $33.33$ & $100.00$ & $00.00$ \\
\bottomrule
\end{tabular}

\vspace{0.6em}

\begin{tabular}{c|ccc||ccc||ccc}
\toprule
& \multicolumn{3}{c||}{\textbf{Increased Cup-Disc}} & \multicolumn{3}{c||}{\textbf{Macular Edema}} & \multicolumn{3}{c}{\textbf{Myopic Fundus}} \\
\textbf{Model} & \textbf{F1} & \textbf{Sensitivity} & \textbf{Specificity} & \textbf{F1} & \textbf{Sensitivity} & \textbf{Specificity} & \textbf{F1} & \textbf{Sensitivity} & \textbf{Specificity} \\
\midrule
RETFound (frozen) & $72.58$ & $72.77$ & $73.11$ & $87.30$ & $87.30$ & $87.21$ & $98.00$ & $98.00$ & $98.05$ \\
\midrule
RETFound (unfrozen) & $\mathbf{80.93}$ & $82.90$ & $78.97$ & $90.84$ & $91.55$ & $90.14$ & $97.00$ & $100.00$ & $94.00$ \\
\bottomrule
\end{tabular}

\vspace{0.6em}

\begin{tabular}{c|ccc||ccc||ccc}
\toprule
& \multicolumn{3}{c||}{\textbf{Nevus}} & \multicolumn{3}{c||}{\textbf{Scar}} & \multicolumn{3}{c}{\textbf{Vascular Occlusion}} \\
\textbf{Model} & \textbf{F1} & \textbf{Sensitivity} & \textbf{Specificity} & \textbf{F1} & \textbf{Sensitivity} & \textbf{Specificity} & \textbf{F1} & \textbf{Sensitivity} & \textbf{Specificity} \\
\midrule
RETFound (frozen) & $60.07$ & $60.27$ & $59.70$ & $80.65$ & $80.83$ & $81.31$ & $\mathbf{94.85}$ & $94.87$ & $94.92$ \\
\midrule
RETFound (unfrozen) & $74.18$ & $92.86$ & $57.14$ & $91.82$ & $95.92$ & $87.76$ & $88.03$ & $80.95$ & $95.24$ \\
\bottomrule
\end{tabular}
}
\end{table*}

\section{RETFound Frozen vs Unfrozen Versions}
\label{sec:retfound_frozen}

\cref{tbl:retfound} provides results of frozen and unfrozen versions of RETFound. 
\section{List of Ophthalmology Journals Used for PMC Data Collection} 

The following 82 ophthalmology journals were used to collect full-text articles from PubMed Central: (1) Acta Ophthalmologica; (2) Advances in Ophthalmology Practice and Research; (3) American Journal of Ophthalmology; (4) American Journal of Ophthalmology Case Reports; (5) Archives of Clinical and Experimental Ophthalmology; (6) Asia-Pacific Journal of Ophthalmology (Philadelphia); (7) BMC Ophthalmology; (8) BMJ Open Ophthalmology; (9) Beyoglu Eye Journal; (10) British Journal of Ophthalmology; (11) Canadian Journal of Ophthalmology; (12) Case Reports in Ophthalmology; (13) Case Reports in Ophthalmological Medicine; (14) Clinical and Experimental Ophthalmology; (15) Clinical Ophthalmology; (16) Community Eye Health; (17) Contact Lens and Anterior Eye; (18) Current Eye Research; (19) Current Ophthalmology Reports; (20) Current Opinion in Ophthalmology; (21) Documenta Ophthalmologica; (22) European Journal of Ophthalmology; (23) Experimental Eye Research; (24) Expert Review of Ophthalmology; (25) Eye (London); (26) Eye and Brain; (27) Eye \& Contact Lens; (28) Eye and Vision (London); (29) Frontiers in Ophthalmology (Lausanne); (30) GMS Ophthalmology Cases; (31) Graefe's Archive for Clinical and Experimental Ophthalmology; (32) ISRN Ophthalmology; (33) Indian Journal of Ophthalmology; (34) Indian Journal of Ophthalmology Case Reports; (35) International Journal of Ophthalmic Research; (36) International Journal of Ophthalmology and Eye Science; (37) International Journal of Retina and Vitreous; (38) International Ophthalmology; (39) Investigative Ophthalmology \& Visual Science; (40) Journal of Clinical and Experimental Ophthalmology; (41) Journal of Clinical Ophthalmology and Eye Disorders; (42) Journal of Current Ophthalmology; (43) Journal of Neuro-Ophthalmology; (44) Journal of Ophthalmic Inflammation and Infection; (45) Journal of Ophthalmic and Vision Research; (46) Journal of Ophthalmology; (47) Journal of Pediatric Ophthalmology and Strabismus; (48) Journal of VitreoRetinal Diseases; (49) JAMA Ophthalmology; (50) JOJ Ophthalmology; (51) Korean Journal of Ophthalmology; (52) Medical Hypothesis, Discovery and Innovation in Ophthalmology; (53) Middle East African Journal of Ophthalmology; (54) Neuro-Ophthalmology; (55) Oman Journal of Ophthalmology; (56) Open Journal of Ophthalmology; (57) The Open Ophthalmology Journal; (58) Ophthalmic Epidemiology; (59) Ophthalmic Genetics; (60) Ophthalmic and Physiological Optics; (61) Ophthalmic Plastic and Reconstructive Surgery; (62) Ophthalmic Research; (63) Ophthalmic Surgery, Lasers and Imaging Retina; (64) Ophthalmology and Eye Diseases; (65) Ophthalmology Glaucoma; (66) Ophthalmology Retina; (67) Ophthalmology Science; (68) Ophthalmology and Therapy; (69) Ophthalmologica; (70) Ophthalmologie; (71) Ophthalmology; (72) Progress in Retinal and Eye Research; (73) Retinal Cases and Brief Reports; (74) Retina; (75) Romanian Journal of Ophthalmology; (76) Saudi Journal of Ophthalmology; (77) Survey of Ophthalmology; (78) Taiwan Journal of Ophthalmology; (79) Therapeutic Advances in Ophthalmology; (80) Turkish Journal of Ophthalmology; (81) US Ophthalmic Review; (82) Veterinary Ophthalmology.
\begin{table}[ht]
\centering
\caption{Training hyperparameters}
\begin{tabular}{ll}
\hline
\textbf{Hyperparameter} & \textbf{Value} \\
\hline
\multicolumn{2}{l}{\textit{Training Configuration}} \\
Number of GPUs & 4 \\
Total batch size & 1 \\
Per-device batch size & 1 \\
Gradient accumulation steps & 1 \\
Mixed precision & bfloat16 \\
DeepSpeed optimization & ZeRO Stage 1 \\
\hline
\multicolumn{2}{l}{\textit{Optimization}} \\
Learning rate & $4 \times 10^{-5}$ \\
LR scheduler & Cosine \\
Weight decay & 0.01 \\
Warmup ratio & 0.03 \\
\hline
\multicolumn{2}{l}{\textit{Vision Configuration}} \\
Image resolution & $448 \times 448$ \\
Max dynamic patches & 6 \\
Downsampling ratio & 0.5 \\
Drop path rate & 0.1 \\
Vision select layer & -1 \\
\hline
\multicolumn{2}{l}{\textit{Model Components}} \\
Frozen backbone & True \\
Frozen LLM & False \\
Frozen MLP & False \\
\hline
\multicolumn{2}{l}{\textit{Sequence \& Memory}} \\
Max sequence length & 9000 \\
Gradient checkpointing & Enabled \\
Group by length & Enabled \\
\hline
\end{tabular}
\label{tab:hyperparameters}
\end{table}

\section{Training Hyperparameters}
\label{sec:training_hyperparameters}

We trained our model using distributed data parallelism across 4 H100 GPUs with mixed precision training (bfloat16) and DeepSpeed ZeRO Stage 1 optimization. The total effective batch size was 4, achieved through a per-device batch size of 1 and gradient accumulation over 1 steps. We employed a cosine learning rate scheduler with an initial learning rate of $4 \times 10^{-5}$, weight decay of 0.01, and a warmup ratio of 0.03 for the first 3\% of training steps.

For the vision encoder configuration, we set the input image resolution to $448 \times 448$ pixels with dynamic image sizing enabled. The model supports up to 6 dynamic patches and uses thumbnail generation for efficient processing. We applied a drop path rate of 0.1 for regularization and selected features from the final layer of the vision encoder.

Following our frozen encoder strategy, we kept the vision backbone frozen (\texttt{freeze\_backbone = True}) while allowing the language model and MLP projector to be trainable (\texttt{freeze\_llm = False}, \texttt{freeze\_mlp = False}). The maximum sequence length was set to 9000 tokens to accommodate detailed medical descriptions and multi-turn conversations. We enabled gradient checkpointing to optimize memory usage and grouped samples by length for efficient batching.

All experiments were conducted using PyTorch's distributed training framework with tensorboard logging for monitoring training progress. Table~\ref{tab:hyperparameters} provides a comprehensive summary of all hyperparameters used in our training pipeline.

\section{License Information}
\label{sec:license_info}

All datasets used in this study were obtained from two primary sources: PubMed Central (PMC) documents and public benchmarking datasets. The PMC documents, which provided image-caption pairs for ophthalmology knowledge pretraining and case reports for reasoning and synthesis training, were distributed under Creative Commons licenses including CC BY, CC BY-NC, CC BY-NC-SA, and CC0. The public benchmarking datasets used for domain task fine-tuning were obtained under the following licenses: BRSET under the PhysioNet Credentialed Health Data License 1.5.0, OIMHS and FIVES under the Creative Commons Attribution 4.0 International License (CC BY 4.0), and EyePACS under the MIT License. Most of these licenses permit redistribution, modification, and derivative works, while BRSET requires credential-based access through the PhysioNet platform to ensure responsible data use. This licensing framework enables the research community to reproduce our work, build upon our model, and deploy it. 

\end{document}